\documentclass[nohyperref]{article}

\usepackage{microtype}
\usepackage{graphicx}
\usepackage{bbm}

\graphicspath{{images/},{prebuiltimages/}}

\usepackage{subcaption}

\usepackage{booktabs} 

\usepackage{hyperref}

\usepackage[accepted]{arxiv}

\usepackage{amsmath}
\usepackage{amssymb}
\usepackage{mathtools}
\usepackage{amsthm}

\usepackage{wrapfig}
\usepackage{shortcuts_cg}

\DeclarePairedDelimiter{\abs}{\lvert}{\rvert}

\usepackage[capitalize,noabbrev]{cleveref}

\theoremstyle{plain}
\newtheorem{theorem}{Theorem}[section]
\newtheorem{proposition}[theorem]{Proposition}

\theoremstyle{definition}
\newtheorem{definition}[theorem]{Definition}

\theoremstyle{remark}

\usepackage[textwidth=2.0cm, textsize=tiny]{todonotes}
\usepackage{adjustbox}
\usepackage{cleveref}

\icmltitlerunning{Stochastic smoothing of the top-K calibrated hinge loss for deep imbalanced classification}

\begin{document}

\twocolumn[
    \icmltitle{Stochastic smoothing of the top-K calibrated hinge loss\\ for deep imbalanced classification}



    \icmlsetsymbol{equal}{*}

    \begin{icmlauthorlist}
        \icmlauthor{Camille Garcin}{imag,inria}
        \icmlauthor{Maximilien Servajean}{lirmm,upvm}
        \icmlauthor{Alexis Joly}{inria}
        \icmlauthor{Joseph Salmon}{imag,iuf}
    \end{icmlauthorlist}

    \icmlaffiliation{imag}{IMAG, Univ Montpellier, CNRS, Montpellier, France}
    \icmlaffiliation{iuf}{Institut Universitaire de France (IUF)}
    \icmlaffiliation{lirmm}{LIRMM, Univ Montpellier, CNRS, Montpellier, France}
    \icmlaffiliation{upvm}{AMIS, Paul Valery University, Montpellier, France}
    \icmlaffiliation{inria}{Inria, LIRMM, Univ Montpellier, CNRS, Montpellier, France}

    \icmlcorrespondingauthor{Camille Garcin}{camille.garcin@umontpellier.fr}

    \icmlkeywords{Machine Learning, ICML}

    \vskip 0.3in
]



\printAffiliationsAndNotice{}  

\begin{abstract}
    In modern classification tasks, the number of labels is getting larger and larger, as is the size of the datasets encountered in practice.
    As the number of classes increases, class ambiguity and class imbalance become more and more problematic to achieve high top-$1$ accuracy.  Meanwhile, Top-$K$ metrics (metrics allowing $K$ guesses) have become popular, especially for performance reporting. Yet, proposing top-$K$ losses tailored for deep learning remains a challenge, both theoretically and practically.
    In this paper we introduce a stochastic top-$K$ hinge loss inspired by recent developments on top-$K$ calibrated losses.
    Our proposal is based on the smoothing of the top-$K$ operator building on the flexible "perturbed optimizer" framework. We show that our loss function performs very well in the case of balanced datasets, while benefiting from a significantly lower computational time than the state-of-the-art top-$K$ loss function. In addition, we propose a simple variant of our loss for the imbalanced case. Experiments on a heavy-tailed dataset show that our loss function significantly outperforms other baseline loss functions.
\end{abstract}


\section{Introduction}
\label{sec:Introduction}
Fine-grained visual categorization (FGVC) has recently attracted a lot of attention \cite{wang2022guest}, in particular in the biodiversity domain \cite{inat2017, plantnet-300k, van2015building}.
In FGVC, one aims to classify an image into subordinate categories (such as plant or bird species) that contain many visually similar instances.
The intrinsic ambiguity among the labels makes it difficult to obtain high levels of top-1 accuracy as is typically the case with standard datasets such as CIFAR10 \cite{cifar100} or MNIST \cite{mnist}.
For systems like Merlin \cite{van2015building} or Pl@ntNet \cite{plantnet}, due to the difficulty of the task, it is generally relevant to provide the user with a set of classes in the hope that the true class belongs to that set.
In practical applications, the display limit of the device only allows to give a few labels back to the user.
A straightforward strategy consists in returning a set of $K$ classes for each input, where $K$ is a small integer with respect to the total number of classes.
Such classifiers are called top-$K$ classifiers, and their performance is evaluated with the well known top-$K$ accuracy \cite{lapin2015, imagenet}.
While such a metric is very popular for evaluating applications, common learning strategies typically consist in learning a deep neural network with the cross-entropy loss, neglecting the top-$K$ constraint in the learning step.

Yet, recent works have focused on optimizing the top-$K$ accuracy directly.
\citet{lapin2015} have introduced the top-$K$ hinge loss and a convex upper-bound, following techniques introduced by \citet{Usunier_Buffoni_Gallinari09} for ranking.
A limit of this approach was raised by \citet{berrada}, as they have shown that the top-$K$ hinge loss by \citet{lapin2015} can not be directly used for training a deep neural network.
The main arguments put forward by the authors to explain this practical limitation are: (i) the non-smoothness of the top-$K$ hinge loss and (ii), the sparsity of its gradient.
Consequently, they propose a smoothed alternative adjustable with a temperature parameter.
However, their smoothing procedure is computationally costly when $K$ increases (as demonstrated in our experiments), despite the efficient algorithm they provide to cope with the combinatorial nature of the loss.
Moreover, this approach has the drawback to be specific to the top-$K$ hinge loss introduced by \citet{lapin2015}.

In contrast, we propose a new top-$K$ loss that relies on the smoothing of the top-$K$ operator (the operator returning the $K$-th largest value of a vector).
The smoothing framework we consider, the perturbed optimizers \citep{berthet2020learning}, can be used to smooth variants of the top-$K$ hinge loss but could independently be considered for other learning tasks such as $K$-nearest neighbors or top-$K$ recommendation \citep{He_2019_CVPR, covington2016deep}.
Additionally, we introduce a simple variant of our loss to deal with imbalanced datasets.
Indeed, for many real-world applications, a long-tailed phenomenon appears \cite{reed2001pareto}: a few labels enjoy a lot of items (\eg images), while the vast majority of the labels receive only a few items, see for instance a dataset like Pl@ntnet-300k \citep{plantnet-300k} for a more quantitative overview.
We find that the loss by \citet{berrada} fails to provide satisfactory results on the tail classes in our experiments.
On the contrary, our proposed loss based on uneven margins outperforms the loss from \citet{berrada} and the LDAM loss \citep{Cao_Wei_Gaidon_Arechiga_Ma19}, a loss designed for imbalance cases known for its very good performance in fine-grained visual classification challenges \citep{RIDE}.
To the best of our knowledge, our proposed loss is the first loss function tackling both the top-$K$ classification and class imbalance problems jointly.


\section{Related work}
\label{sec:related-work}

Several top-$K$ losses have been introduced and experimented with in \cite{lapin2015, lapin2016, lapin2017analysis}.
However, the authors assume that the inputs are features extracted from a deep neural network and optimize their losses with SDCA \cite{sdca}.
\citet{berrada} have shown that the top-$K$ hinge loss from \citet{lapin2015} could not be directly used in a deep learning optimization pipeline.
Instead, we are interested in end-to-end deep neural network learning.
The state-of-the art top-$K$ loss for deep learning is that of \citet{berrada}, which is a smoothing of a top-$K$ hinge loss by \citet{lapin2015}.
The principle of the top-$K$ loss of \citet{berrada} is based on the rewriting the top-$K$ hinge loss of \citet{lapin2015} as a difference of two maxes on a combinatorial number of terms,
smooth the max with the logsumexp, and use a divide-and-conquer approach to make their loss tractable.
Instead, our approach relies on smoothing the top-$K$ operator and using this smoothed top-$K$ operator on a top-$K$ calibrated loss recently proposed by \citet{topk_yang}.
Our approach could be used out-of-the box with other top-$K$ hinge losses. In contrast, the smoothing method of \citet{berrada} is tailored for the top-$K$ hinge loss of \citet{lapin2015}, which is shown to be not top-$K$ calibrated in \cite{topk_yang}.

For a general theory of smoothing in optimization, we refer the reader to \citet{beckteboule, nesterov2005} while for details on perturbed optimizers, we refer the reader to \citet{berthet2020learning} and references therein.
In the literature, other alternatives have been proposed to perform top-$K$ smoothing.
\citet{topk_ot} formulate the smooth top-$K$ operator as the solution of a regularized optimal transport problem between well-chosen discrete measures.
The authors rely on a costly optimization procedure to compute the optimal plan. \citet{subsetsampling} propose a smoothing of the top-$K$ operator through $K$ successive \textit{softmax}.
Besides the additional cost with large $K$, the computation of $K$ successive \textit{softmax} brings numerical instabilities.

Concerning imbalanced datasets, several recent contributions have focused on architecture design \citep{BBN, RIDE}.
Instead, we focus here on the design of the loss function and leverage existing popular neural networks architectures.
A popular loss for imbalanced classification is the focal loss \citep{focal} which is a modification of the cross entropy where well classified-examples induce a smaller loss, putting emphasis on difficult examples.
Instead, we use uneven margins in our formulation, requiring examples of the rarest classes to be well classified by a larger margin than examples of the most common classes.
Uneven margin losses have been studied in the binary case in \citep{scott, uneven_margin_svm, cost-sensitive-svm}.
For the multi-class setting, the LDAM loss \citep{Cao_Wei_Gaidon_Arechiga_Ma19} is a widely used uneven margin loss which can be seen as a cross entropy incorporating uneven margins in the logits.
Instead, our imbalanced top-$K$ loss relies on the smoothing of the top-$K$ operator.

\section{Proposed method}
\label{sec:Proposed method}

\captionsetup[subfigure]{justification=justified}
\begin{figure*}[t!]
    \centering
    \begin{subfigure}[b]{0.187\textwidth}
        \centering
        \includegraphics[width=\textwidth]{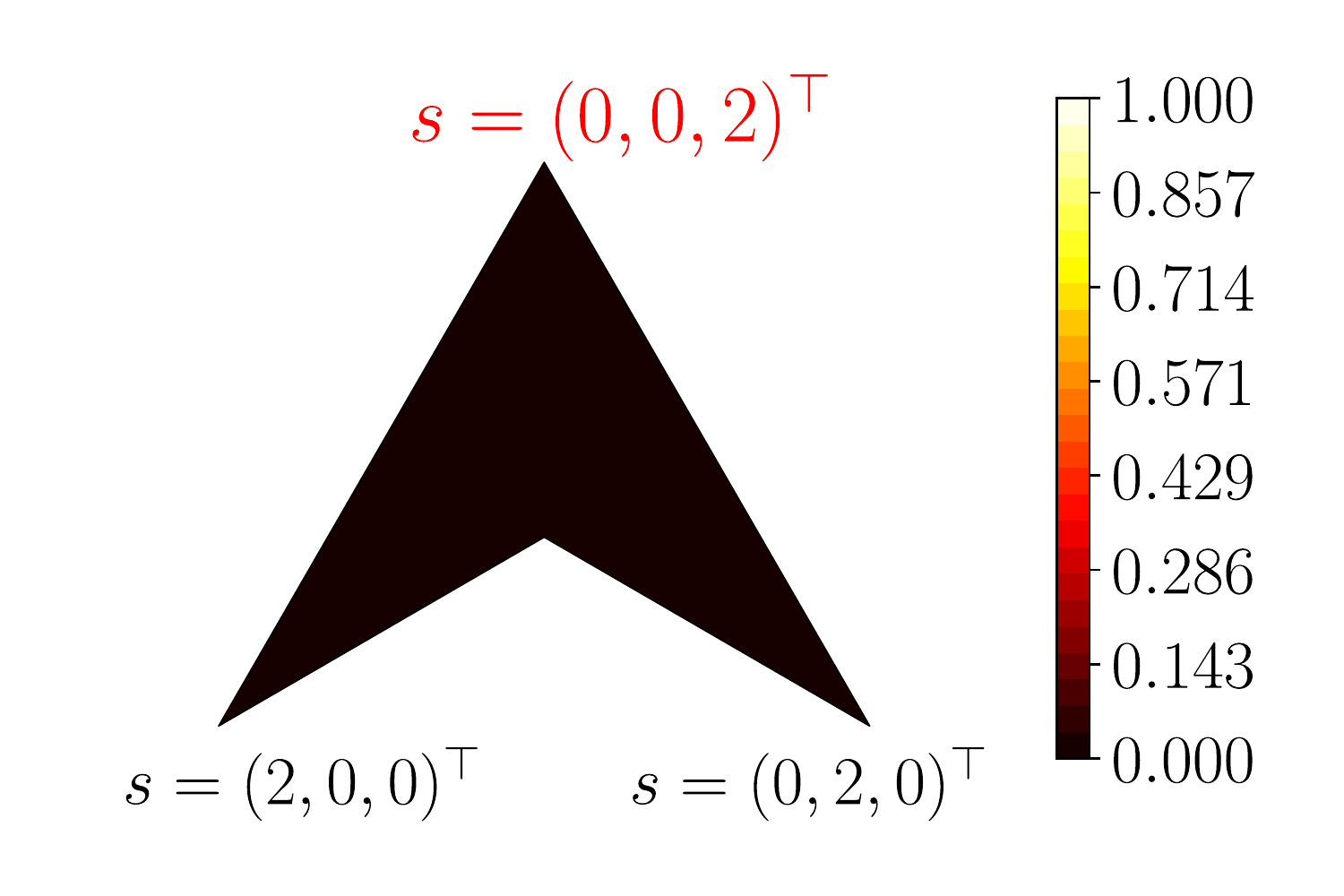}
        \caption{Top-$K$: \phantom{C}\\ $\ell=\ell^{K}$.\phantom{$\ell=\ell_{\textrm{Cal.~Hinge}}^K$}}
        \label{subfig:topk}
    \end{subfigure}
    \hspace{0.1cm}
    \begin{subfigure}[b]{0.187\textwidth}
        \centering
        \includegraphics[width=\textwidth]{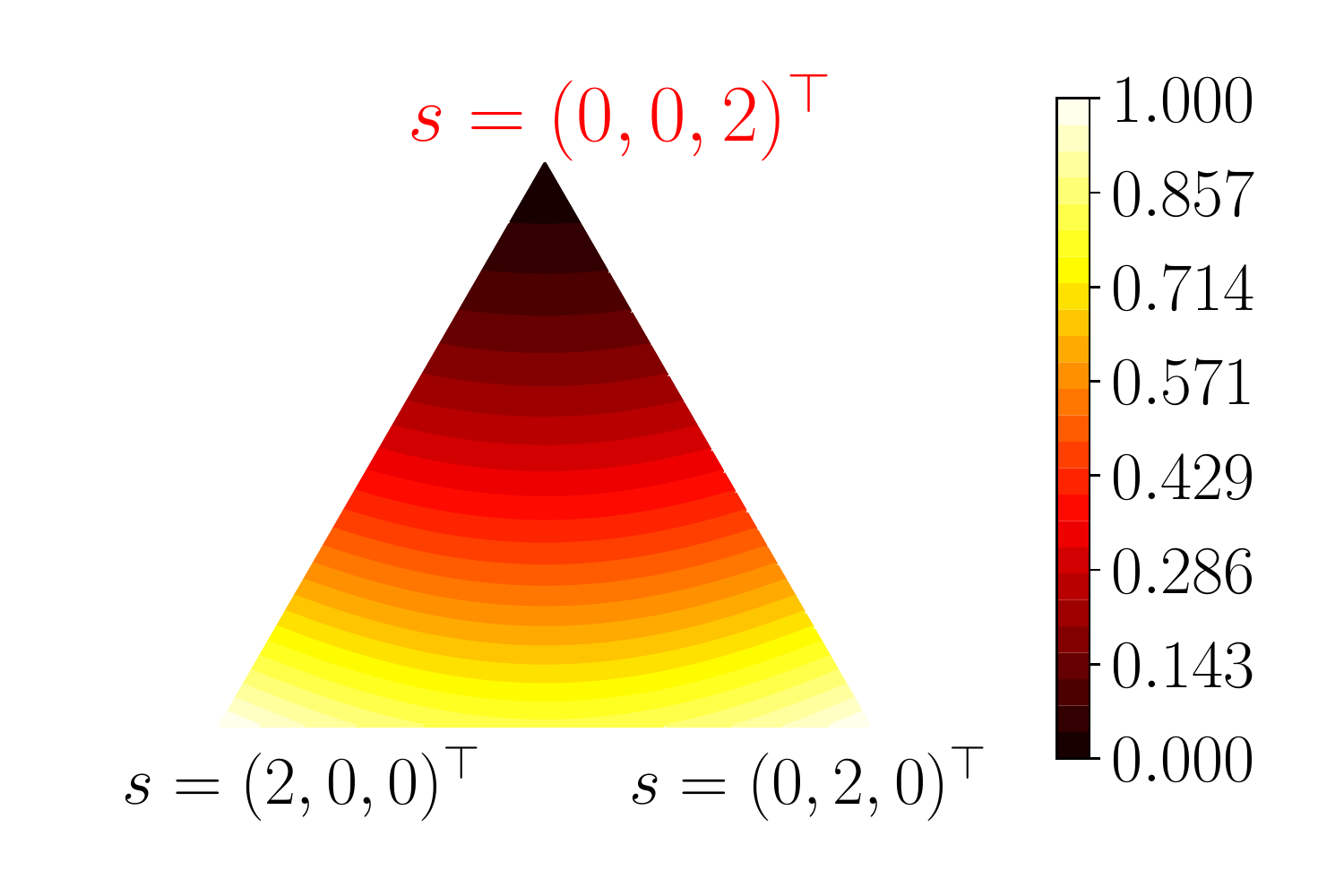}
        \caption{Cross-entropy:\\ $\ell=\ell_{\textrm{CE}}$.\phantom{$\ell=\ell_{\textrm{Cal.~Hinge}}^K$}}
        \label{subfig:ce}
    \end{subfigure}
    \hspace{0.1cm}
    \begin{subfigure}[b]{0.187\textwidth}
        \centering
        \includegraphics[width=\textwidth]{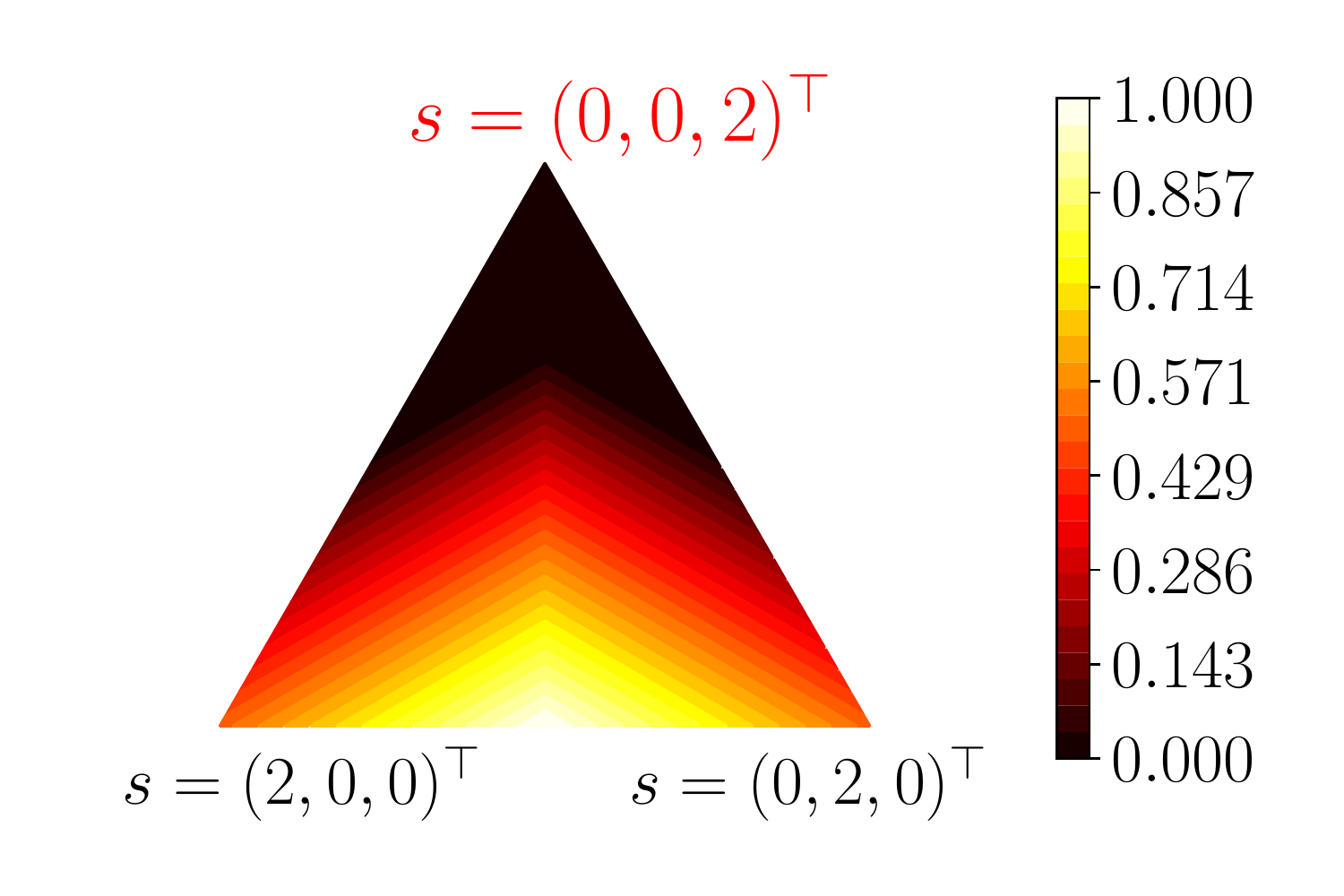}
        \caption{Multi-class hinge:\\ $\ell=\ell_{\textrm{Hinge}}^K$.}
        \label{subfig:lapin}
    \end{subfigure}
    \hspace{0.1cm}
    \begin{subfigure}[b]{0.187\textwidth}
        \centering
        \includegraphics[width=\textwidth]{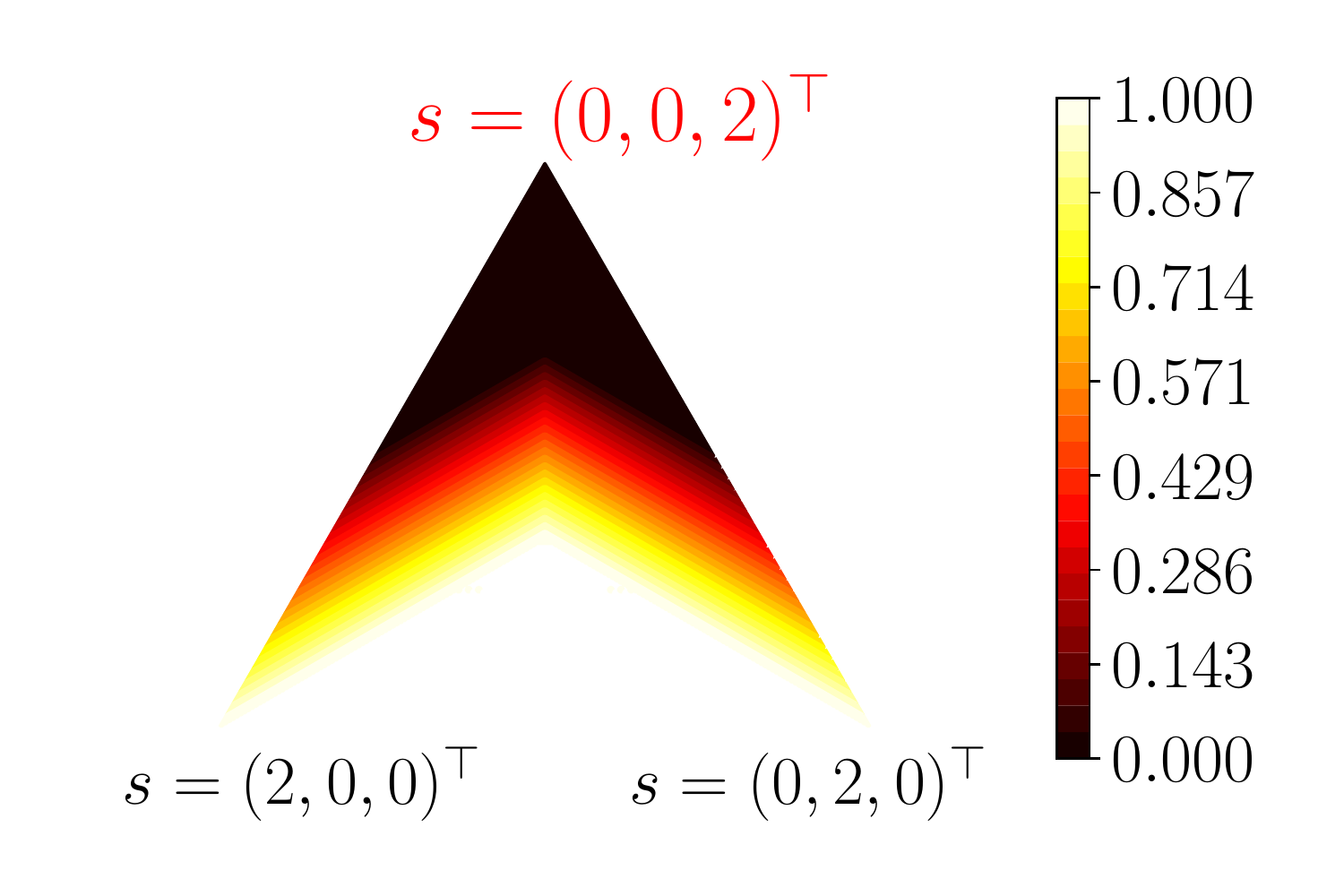}
        \caption{Calibrated hinge:\\$\ell=\ell_{\textrm{Cal.~Hinge}}^K$.}
        \label{subfig:yang}
    \end{subfigure}
    \hspace{0.1cm}
    \begin{subfigure}[b]{0.187\textwidth}
        \centering
        \includegraphics[width=\textwidth]{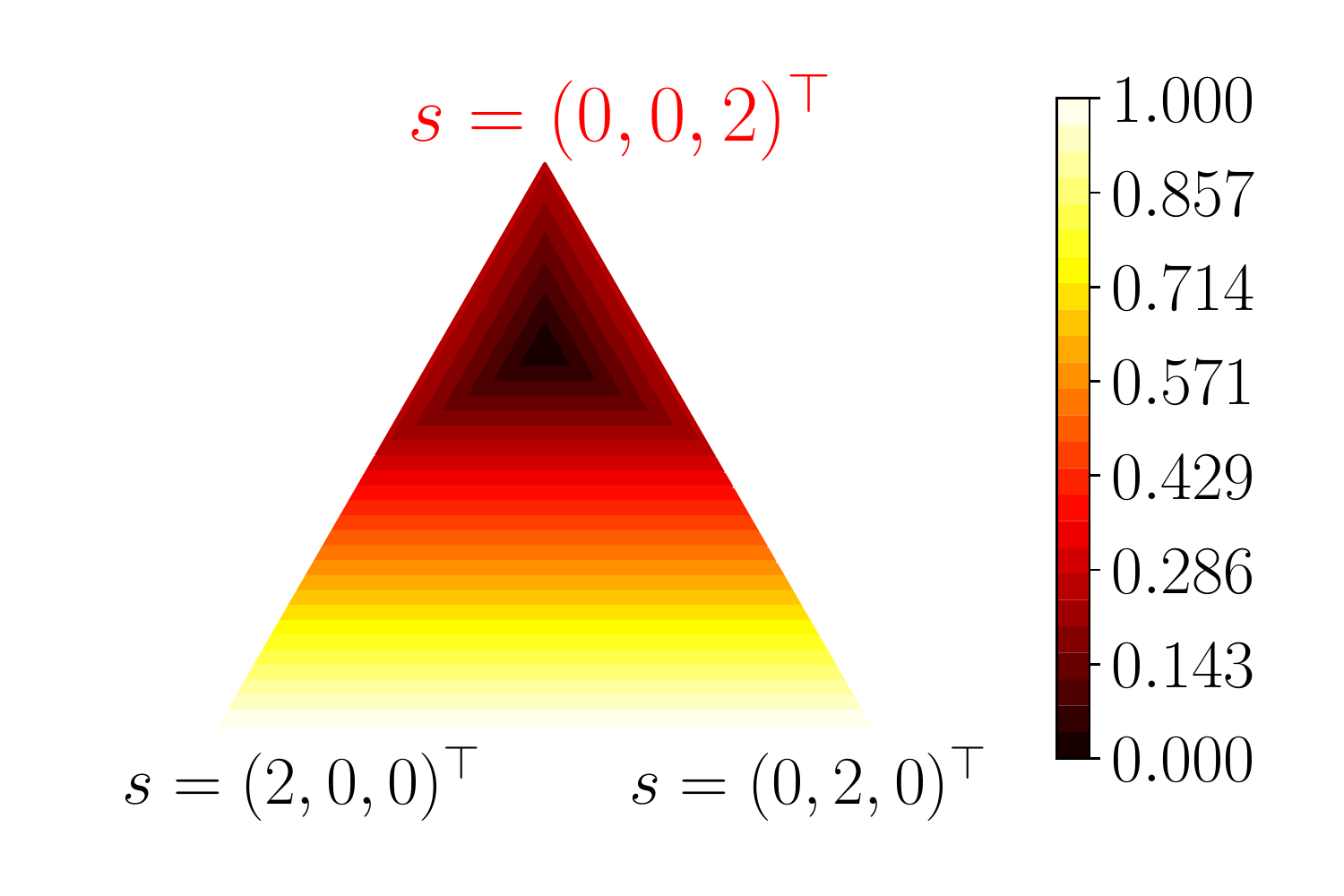}
        \caption{Convexified hinge:\\$\ell=\ell_{\textrm{CVXHinge}}^K$.}
        \label{subfig:lapinconvex}
    \end{subfigure}\\
    \begin{subfigure}[b]{0.187\textwidth}
        \centering
        \includegraphics[width=\textwidth]{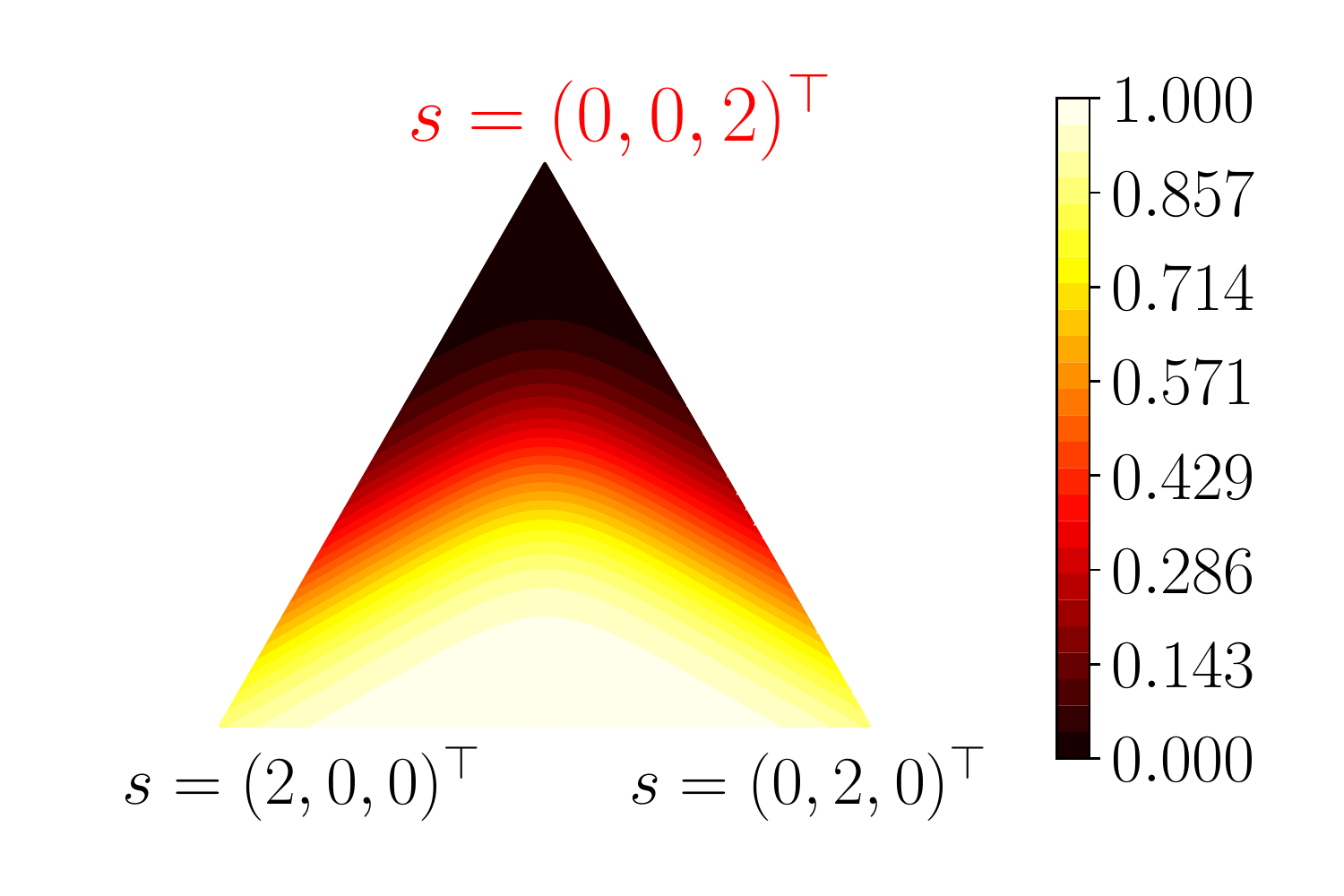}
        \caption{Smoothed hinge\\$\ell_{\text{Smoothed Hinge}}^{K, 0.1}$.}
        \label{subfig:berrada0.01}
    \end{subfigure}
    \hspace{0.1cm}
    \begin{subfigure}[b]{0.187\textwidth}
        \centering
        \includegraphics[width=\textwidth]{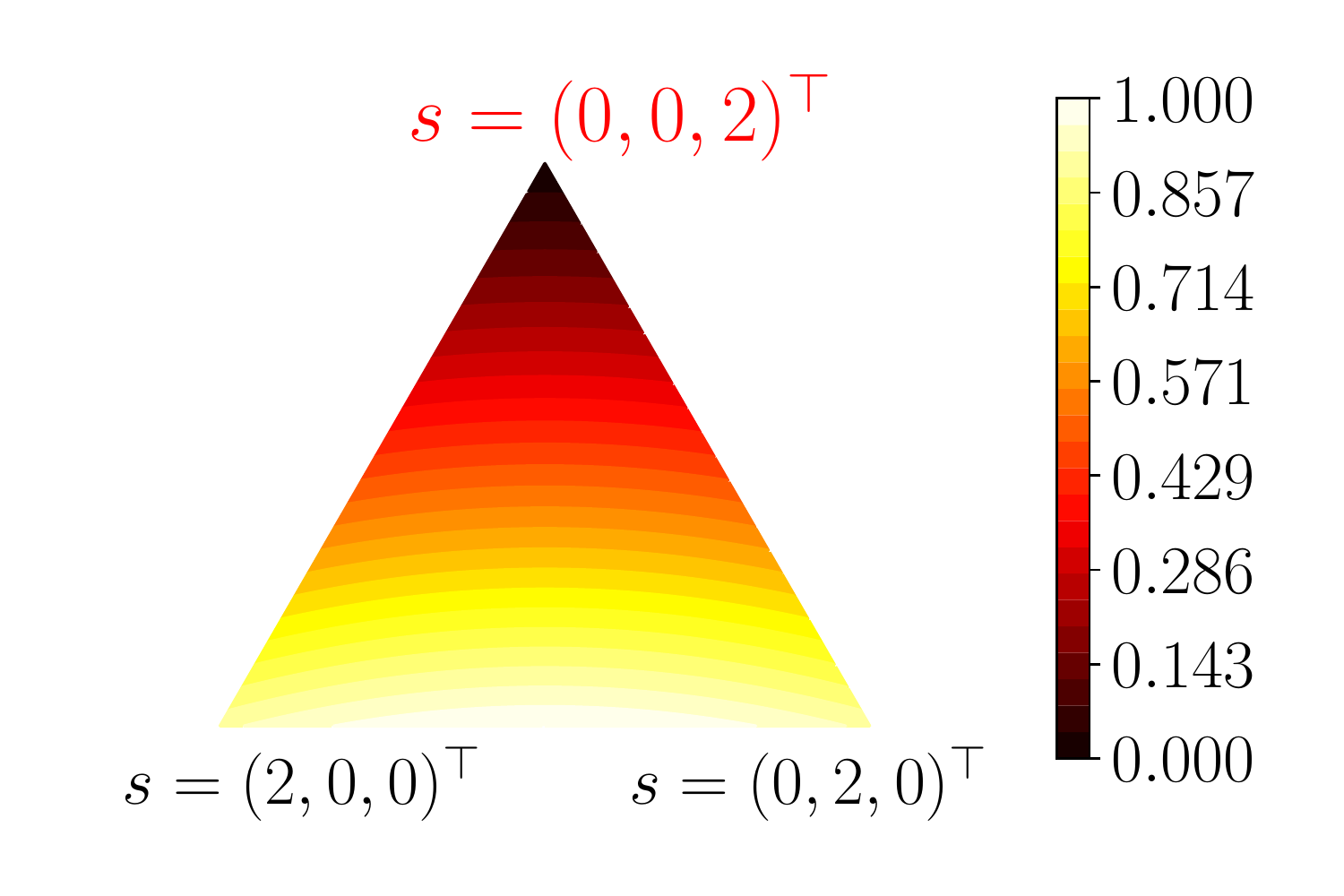}
        \caption{Smoothed hinge\\$\ell_{\text{Smoothed Hinge}}^{K, 1}$.}
        \label{subfig:berrada0.4}
    \end{subfigure}
    \hspace{0.1cm}
    \begin{subfigure}[b]{0.187\textwidth}
        \centering
        \includegraphics[width=\textwidth]{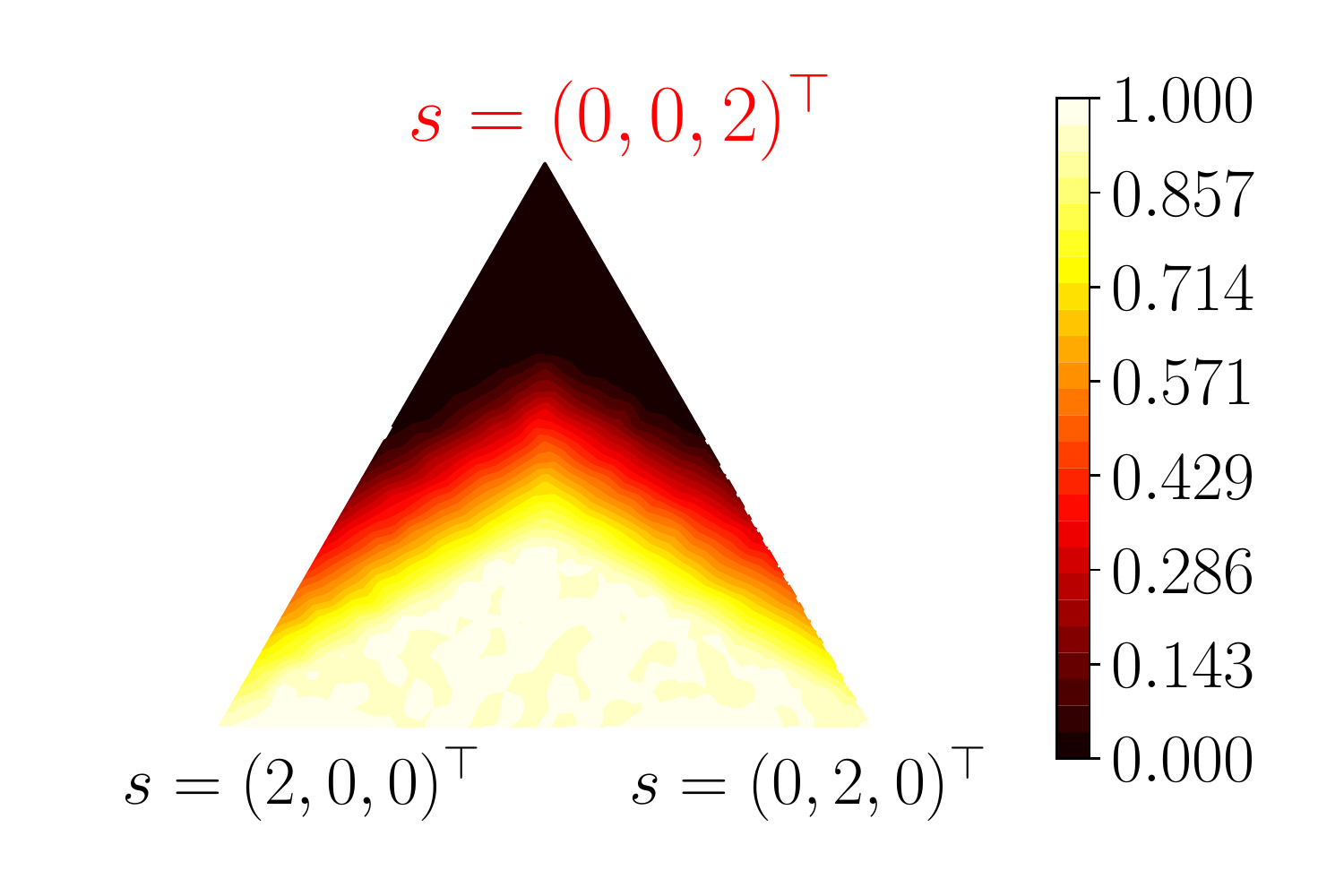}
        \caption{Noised balanced: \phantom{g}\\ $\ell_{\text{Noised bal.}}^{K, 0.3, 30}$. \phantom{$\ell_{\text{Smoothed Hinge}}^{K, 0.4}$}  }
        \label{subfig:epsilon_K2_03}
    \end{subfigure}
    \hspace{0.1cm}
    \begin{subfigure}[b]{0.187\textwidth}
        \centering
        \includegraphics[width=\textwidth]{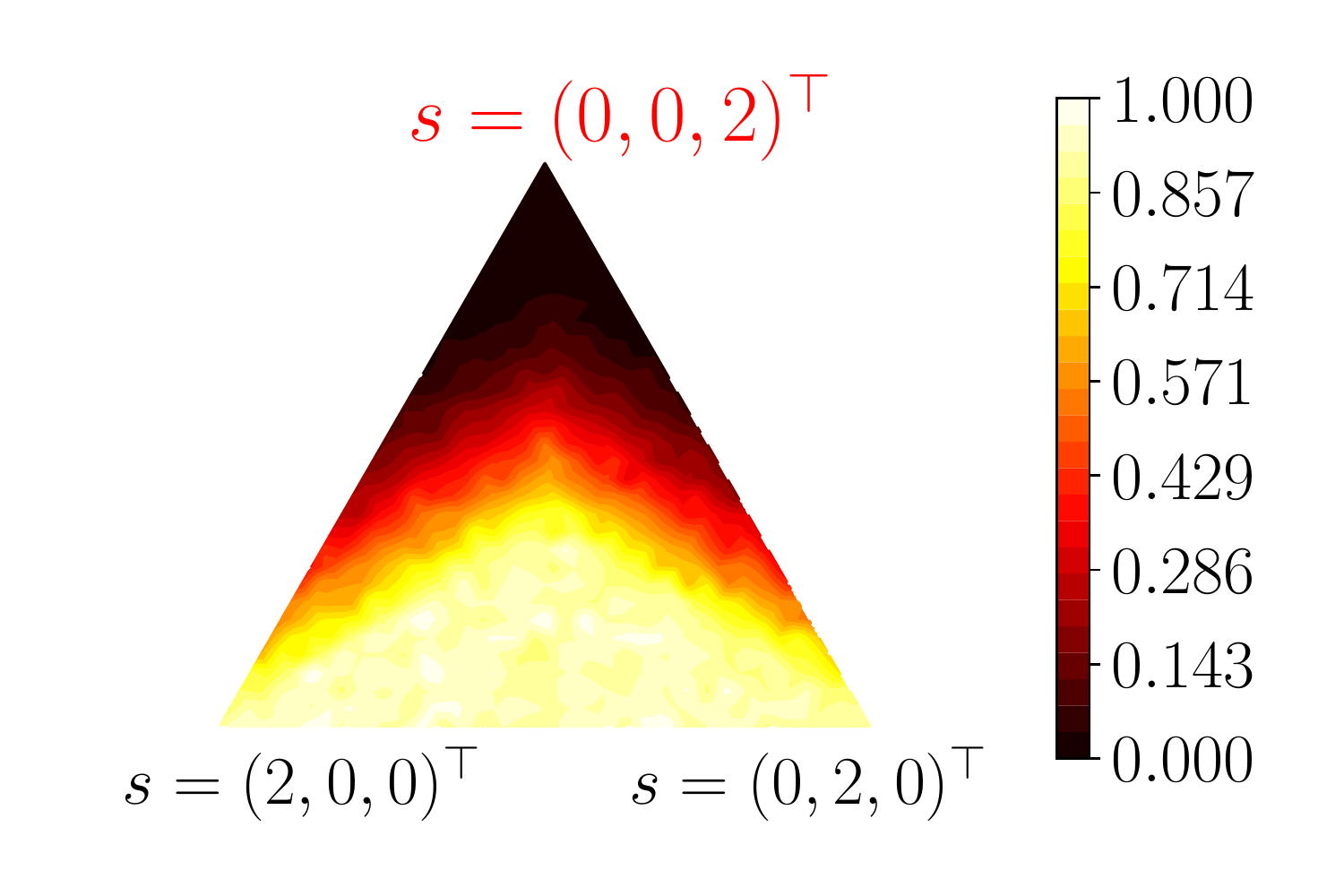}
        \caption{Noised balanced: \phantom{g}\\ $\ell_{\text{Noised bal.}}^{K, 1, 30}$. \phantom{$\ell_{\text{Smoothed Hinge}}^{K, 0.4}$}}
        \label{subfig:epsilon_K2_1}
    \end{subfigure}
    \hspace{0.1cm}
    \begin{subfigure}[b]{0.187\textwidth}
        \centering
        \includegraphics[width=\textwidth]{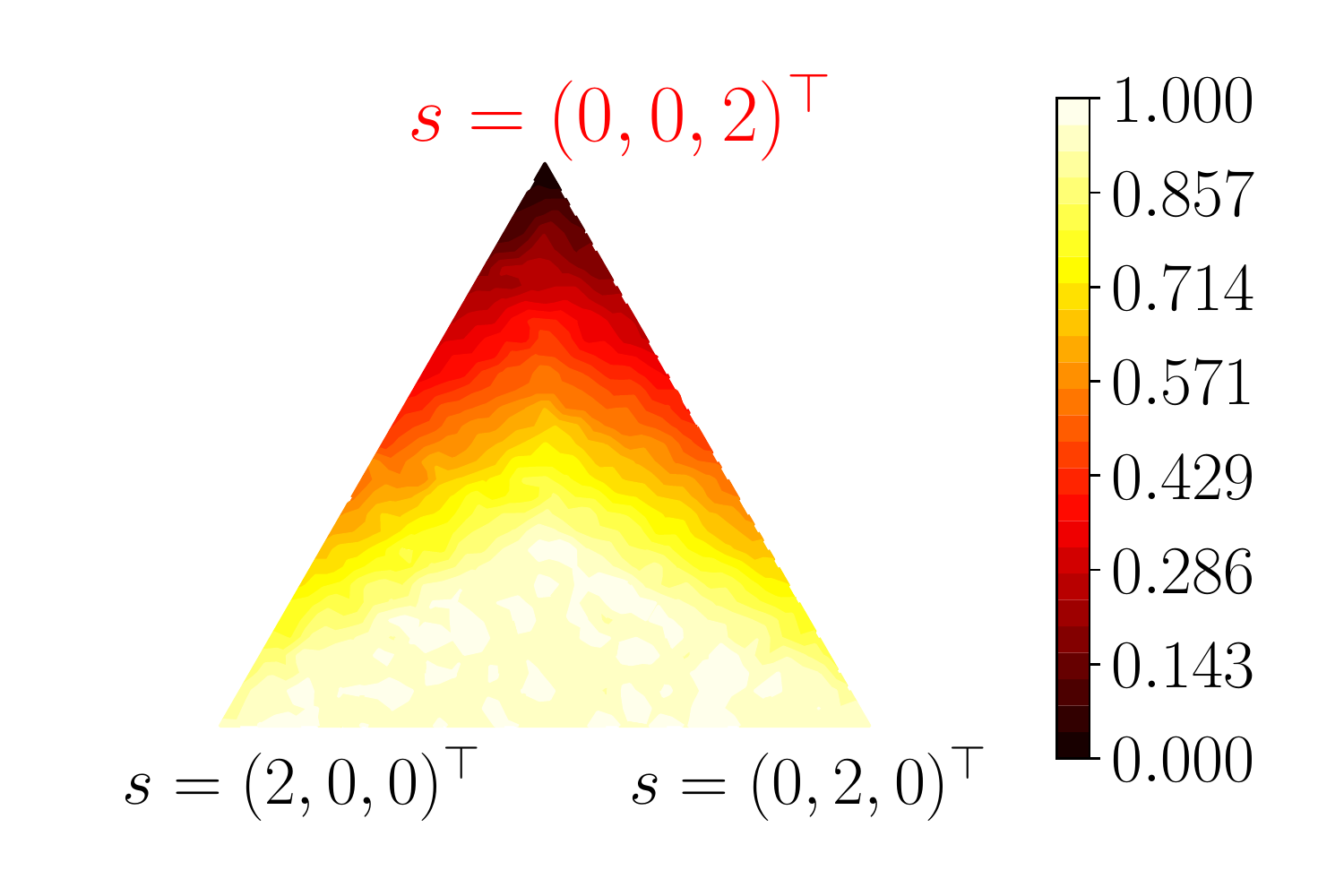}
        \caption{Noised imbalanced: \phantom{g}\\ $\ell_{\text{Noised Imbal.}}^{K, 1, 30,5}$. \phantom{$\ell_{\text{Smoothed Hinge}}^{K, 0.4}$}}
        \label{subfig:epsilon_K2_Imbal_1}
    \end{subfigure}
    \caption{Level sets of the function $\bfs\mapsto \ell(\bfs,y)$ for different losses described in \Cref{tab:losses}, for $L=3$ classes, $K=2$ and a true label $y=3$ (corresponding to the upper corner of the triangles).
        For visualization the loss are rescaled between 0 and 1, and the level sets are restricted to vector $\bfs \in 2 \cdot \Delta_3$.
        The losses have been harmonized to display a margin equal to 1.
        For our proposed loss, we have averaged the level sets over 100 replications to avoid meshing artifacts.}
    \label{fig:simplices_k=2}
\end{figure*}

\renewcommand{\arraystretch}{1.8}
\begin{table*}[t]
    \caption{Summary of standard top-$K$ losses: vanilla top-$K$ $\ell^K$; Cross Entropy $\ell_{\textrm{CE}}$; hinge top-$K$ $\ell_{\textrm{Hinge}}^K$; Convexified hinge top-$K$ $\ell_{\textrm{CVXHinge}}^K$; Calibrated hinge top-$K$ $\ell_{\textrm{Cal.~Hinge}}^K$; Log-sum Smoothed hinge top-$K$ $\ell_{\text{Smoothed Hinge}}^{K, \tau}$; Noised balanced hinge top-$K$ $\ell_{\text{Noised bal.}}^{K, \epsilon, B}$ (\textbf{proposed}); Noised imbalanced hinge top-$K$ $\ell_{\text{Noised Imbal.}}^{K, \epsilon, B,m_y}$ (\textbf{proposed}).}
    \vskip 0.1in
    \centering
    \begin{scriptsize}
        \begin{tabular}{lccr}
            \toprule
            \textbf{Loss} : $\ell(\bfs,y)$                              & \textbf{Expression}                                                                                                                      & \textbf{Param.}             & \textbf{Reference}                          \\ \midrule
            $\ell^K(\bfs, y)$                                           & $\ind_{\{ \topp_{K}(\bfs)>s_y\}}$                                                                                                        & $K$                         & \Cref{eq:topkloss}                          \\
            $\ell_{\textrm{CE}}(\bfs, y)$                               & $-\ln\Big(e^{s_y}/\sum_{k\in[L]} e^{s_k}\Big)$                                                                                           & ---                         &                                             \\
            $\ell_{\textrm{LDAM}}^{m_y}(\bfs, y)$                       & $-\ln\Big(e^{s_y-m_y}/\big[e^{s_y-m_y}+ \sum_{k\in[L], k\neq y} e^{s_k}\big]\Big)$                                                       & $m_y$                       & \citep{focal}                               \\
            $\ell_{\textrm{focal}}^{\gamma}(\bfs, y)$                   & $(1-\log\left[\ell_{\textrm{CE}}(\bfs, y)\right]\Big)^\gamma \ell_{\textrm{CE}}(\bfs, y)$                                                & $\gamma$                    & \citep{Cao_Wei_Gaidon_Arechiga_Ma19}        \\
            $\ell_{\textrm{Hinge}}^K(\bfs, y)$                          & $\left( 1 + \topp_{K}(\bfs_{\setminus y}) - s_{y}\right)_{+}$                                                                            & $K$                         & \Cref{eq:hinge_lapin}, \cite{lapin2015}     \\
            $\ell_{\textrm{CVXHinge}}^K(\bfs, y)$                       & $\left( \frac{1}{K}\sum_{k\in [K]} \topp_{k}(\1_L -\delta_y + \bfs) - s_{y}\right)_{+}$                                                  & $K$                         & \cite{lapin2015}                            \\
            $\ell_{\textrm{Cal.~Hinge}}^K(\bfs, y) $                    & $(1+ \topp_{K+1}(\bfs) - s_{y})_{+}$                                                                                                     & $K$                         & \Cref{eq:hinge yang}, \cite{topk_yang}      \\
            $\ell_{\text{Smoothed Hinge}}^{K, \tau}(\bfs, y) $          & $\begin{aligned} \displaystyle\tau \ln \Big[\sum_{\substack{A \subset [L], |A|=K}} e^{\tfrac{\ind_{\{y\notin A\}}}{\tau}+ \sum\limits_{j \in A} \tfrac{s_{j}}{K\tau}}\Big]-\tau \ln \Big[\sum_{\substack{A \subset [L], |A|=K}} e^{ \sum\limits_{j \in A} \tfrac{s_{j}}{K \tau}} \Big]\end{aligned}$                                                                                                             & $K$, $\tau$                 & \cite{berrada}                              \\

            $\ell_{\text{Noised bal.}}^{K, \epsilon, B}(\bfs, y)$       & $(1 + \widehat{\topp}_{K + 1 , \epsilon, B}(\bfs) - s_{y})_{+}$, where $\widehat{\topp}_{K+1, \epsilon,B}(\bfs)$ is the noisy top-$K+1$  & $K$, $\epsilon$, $B$        & \Cref{eq:bal-loss},   (\textbf{proposed})   \\
            $\ell_{\text{Noised Imbal.}}^{K, \epsilon, B,m_y}(\bfs, y)$ & $(m_y + \widehat{\topp}_{K + 1 , \epsilon,B}(\bfs) - s_{y})_{+}$, where $\widehat{\topp}_{K+1, \epsilon,B}(\bfs)$ is the noisy top-$K+1$ & $K$, $\epsilon$, $B$, $m_y$ & \Cref{eq:imbalanced},   (\textbf{proposed}) \\ \bottomrule
        \end{tabular}
    \end{scriptsize}
    \label{tab:losses}
    \vskip -0.1in
\end{table*}

\subsection{Preliminaries}
Following classical notation, 
we deal with multi-class classification that considers the problem of learning a classifier from $\cX$ to $[L]\eqdef\{1,\dots,L\}$ based on $n$ pairs of \textit{(input, label)} \iid sampled from a joint distribution $\bbP$: $(x_1, y_1), \dots, (x_n, y_n) \in \cX \times [L]$, where $\cX$ is the input data space ($\cX$ is the space of RGB images of a given size in our vision applications) and the $y$'s are the associated labels among $L$ possible ones.

For a training pair of observed features and label $(x, y)$, $\bfs \in \bbR^{L}$ refers to the associated score vector (often referred to as \emph{logits}).
From now on, we use bold font to represent vectors.
For $k \in [L]$, $s_{k}$ refers to the score attributed to the $k$-th class while $s_{y}$ refers to the score of the true label and $s_{(k)}$ refers to the $k$-th largest score\footnote{Ties are broken arbitrarily.}, so that $s_{(1)}\geq \dots \geq s_{(k)} \geq \dots \geq s_{(L)}$.
For $K\in [L]$, we define $\topp_{K}$ and $\topsum_{K}$, functions from $\bbR^L$ to $\bbR$ as:
\begin{align}\label{eq:topks}
    \topp_{K}   & : \bfs \mapsto s_{(K)}                          \\
    \topsum_{K} & : \bfs \mapsto \sum_{k\in [K]} s_{(k)}\enspace.
\end{align}
We write $\1_L=(1,\dots,1)^\top\in \bbR^L$.
For $\bfs \in \bbR^L$, the gradient $\nabla\topp_{K}(\bfs)$ is a vector with a single one at the $K$-th largest coordinate of $\bfs$ and 0 elsewhere (denoted as $\argtopp_K(\bfs)$). Similarly, $\nabla\topsum_{K}(\bfs)$ is a vector with $K$ ones at the $K$ largest coordinates of $\bfs$ and 0 elsewhere (denoted as $\argtops_K(\bfs)$).


The top-$K$ loss (a 0/1 loss) can now be written
\begin{align}\label{eq:topkloss}
    \ell^K(\bfs, y)= \ind_{\{ \topp_{K}(\bfs)>s_y\}}\enspace.
\end{align}
This loss reports an error when the score of the true label $y$ is not among the $K$-th largest scores.
One would typically seek to minimize this loss.
Yet, being a piece-wise constant function \wrt to its first argument\footnote{See for instance \Cref{subfig:topk} for a visualization.}, numerical difficulties make solving this problem particularly hard in practice.
In what follows we recall some popular surrogate top-$K$ losses from the literature before providing new alternatives.
We summarize such variants in \Cref{tab:losses} and illustrate their differences in \Cref{fig:simplices_k=2} for $L=3, K=2$ (see also \Cref{fig:simplices_k=1} in Appendix, for $L=3, K=1$).

A first alternative introduced by \citet{lapin2015} is a relaxation
generalizing the multi-class hinge loss introduced by \citet{crammer} to the top-$K$ case:
\begin{align}
    \label{eq:hinge_lapin}
    \ell_{\textrm{Hinge}}^K(\bfs, y) = \left( 1 + \topp_{K}(\bfs_{\setminus y}) - s_{y}\right)_{+}\enspace,
\end{align}
where $\bfs_{\setminus y}$ is the vector in $\bbR^{d-1}$ obtained by removing the $y$-th coordinate of $\bfs$, and $(\cdot)_+ \eqdef \max(0,\cdot)$.
The authors propose a convex loss function $\ell_{\textrm{CVXHinge}}^K$ (see \Cref{tab:losses}) which upper bounds the loss function $\ell_{\textrm{Hinge}}^K$.

\citet{berrada} have proposed a smoothed counterpart of $\ell_{\textrm{Hinge}}^K$, relying on a recursive algorithm tailored for their combinatorics smoothed formulation.
Yet, a theoretical limitation of $\ell_{\textrm{Hinge}}^K$ and $\ell_{\textrm{CVXHinge}}^K$ was raised by \citet{topk_yang} showing that they are not top-$K$ calibrated. Top-$K$ calibration is a property defined by \citet{topk_yang}.
We recall some technical details in \Cref{sec:Reminder on Top-$K$ calibration} and the precise definition of top-$K$ calibration is given in \Cref{def:topk-calibrated}.

We let $\Delta_L\eqdef \{\bfpi \in \bbR^L: \sum_{k\in [L]}\pi_k=1, \pi_k \geq 0\}$ denote the probability simplex of size $L$.
For a score $\bfs\in\bbR^L$ and $\bfpi \in \Delta_L $ representing the conditional distribution of $y$ given $x$, we write the conditional risk at $x\in\cX$ as $\cR_{\ell|x}(\bfs, \bfpi)=\bbE_{y|x\sim\pi}(\ell(\bfs,y))$ and the (integrated) risk as $\cR_{\ell}(f)\eqdef\bbE_{(x,y)\sim \bbP}[\ell(f(x),y)]$ for a scoring function $f : \cX \to \bbR^{L}$.
The associated Bayes risks are defined respectively by $\cR_{\ell|x}^*(\bfpi)\eqdef  \inf _{\bfs \in \mathbb{R}^{L}} \cR_{\ell|x}(\bfs, \bfpi)$ and $\cR_{\ell}^*\eqdef \inf_{f: \cX \to \bbR^L}\cR_{\ell}(f)$.
The following result by \citep{topk_yang} shows that a top-$K$ calibrated loss is top-$K$ consistent, meaning that a minimizer of such a loss would also lead to Bayes optimal classifiers:
\begin{theorem}
    \citep[Theorem 2.2]{topk_yang}\textbf{.}
    Suppose $\ell$ is a nonnegative top-$K$ calibrated loss function. Then, $\ell$ is top-$K$ consistent, \ie for any sequence of measurable
    functions $f^{(n)} : \cX \rightarrow \bbR^{L}$, we have:
    \begin{align*}
        \cR_{\ell}\left(f^{(n)}\right) \rightarrow \cR_{\ell}^{*} \Longrightarrow \cR_{\ell^K}\left(f^{(n)}\right) \rightarrow \cR_{\ell^K}^{*}\enspace.
    \end{align*}
\end{theorem}
In their paper, \citet{topk_yang} propose a slight modification of the multi-class hinge loss $    \ell_{\textrm{Hinge}}^K$ and show that it is top-$K$ calibrated:
\begin{align}
    \label{eq:hinge yang}
    \ell_{\textrm{Cal.~Hinge}}^K(\bfs, y) = (1 + \topp_{K+1}(\bfs) - s_{y})_{+}\enspace.
\end{align}


The loss $\ell_{\textrm{Cal.~Hinge}}^K$ thus has an appealing theoretical guarantee that  $\ell_{\textrm{Hinge}}^K$ does not have.
Therefore we will use $\ell_{\textrm{Cal.~Hinge}}^K$ as the starting point of our smoothing proposal.

\subsection{New loss for balanced top-$K$ classification}
\label{subsec:balanced}

\citet{berrada} have shown experimentally that a deep learning model trained with $\ell_{\textrm{Hinge}}^K$ does not learn.
The authors claim that the reason for this is the non smoothness of the loss and the sparsity of its gradient.

We also show in \Cref{tab:tab-epsilon} that a deep learning model trained with $\ell_{\textrm{Cal.~Hinge}}^K$ yields poor results.
The problematic part stems from the top-$K$ function which is non-smooth and whose gradient has only one non-zero element (that is equal to one).
In this paper we propose to smooth the top-$K$ function with the \emph{perturbed optimizers} method developed by \citet{berthet2020learning}.
We follow this strategy due to its flexibility and to the ease of evaluating associated first order information (a crucial point for deep neural network frameworks).


\begin{definition}
    For a smoothing parameter $\epsilon>0$, we define for any $\bfs \in \bbR^{L}$ the  $\epsilon$-smoothed version of $\topsum_{K}$ as:
    \begin{align}\label{eq:topsum_epsilon}
        \topsum_{K, \epsilon}(\bfs) \eqdef \bbE_Z[\topsum_{K}(\bfs + \epsilon Z)]\enspace,
    \end{align}
    where $Z$ is a standard normal random vector, \ie $Z\sim\mathcal{N}(0,\Id_L)$.
\end{definition}

\begin{proposition}
    \label{prop:smoothness}
    For a smoothing parameter $\epsilon>0$,
    \begin{itemize}
        \item The function $\topsum_{K, \epsilon}: \bbR^{L} \to \bbR$ is strictly convex, twice differentiable and $\sqrt{K}$-Lipschitz continuous.
        \item The gradient of $\topsum_{K, \epsilon}$ reads:
              \begin{align}\label{eq:gradients}
                  \nabla_{\bfs} \topsum_{K, \epsilon}(\bfs) = \bbE[\argtops_{K}(\bfs + \epsilon Z)] \enspace.
              \end{align}
        \item $\nabla_{\bfs}\topsum_{K, \epsilon}$ is $\tfrac{\sqrt{KL}}{\epsilon}$-Lipschitz. 
        \item When $\epsilon \rightarrow 0$, $\topsum_{K, \epsilon}(\bfs) \rightarrow \topsum_{K}(\bfs)$.
    \end{itemize}
\end{proposition}
All proofs are given in the appendix.

The smoothing strategy introduced leads to a natural smoothed approximation of the top-$K$ operator, leveraging the link $\topp_{K}(\bfs) = \topsum_{K}(\bfs) - \topsum_{K-1}(\bfs)$ for any score $\bfs\in\bbR^{L}$ (where we use the convention $\topsum_{0}=\mathbf{0}_L\in\bbR^L)$:
\begin{definition}
    \label{def:top-k}
    For any $s\in\bbR^{L}$ and $k\in [L]$, we define
    \begin{align*}
        \topp_{K, \epsilon}(\bfs) \eqdef \topsum_{K, \epsilon}(\bfs) - \topsum_{K-1, \epsilon}(\bfs)\enspace.
    \end{align*}
\end{definition}
This definition leads to a smooth approximation of the $\topp_{K}$ function, in the following sense:
\begin{proposition}\label{prop:topkapprox}
    For a smoothing parameter $\epsilon>0$,
    \begin{itemize}
        \item $\topp_{K, \epsilon}$ is $\frac{4\sqrt{KL}}{\epsilon}$-smooth.
        \item For any $\bfs \in \bbR^{L}$, $\abs{\topp_{K, \epsilon}(\bfs) - \topp_{K}(\bfs)} \leq \epsilon \cdot C_{K, L}$,
              where $C_{K, L}=K\sqrt{2\log L} $.
    \end{itemize}
\end{proposition}
Observe that the last point implies that for any $\bfs \in \bbR^{L}$, $\topp_{K, \epsilon}(\bfs) \rightarrow \topp_{K}(\bfs)$ when $\epsilon \rightarrow 0$.

We can now define an approximation of the calibrated top-$K$ hinge loss $\ell_{\textrm{Cal.~Hinge}}^K$ using $\topp_{K, \epsilon}$ in place of $\topp_{K}$ (see \Cref{subfig:epsilon_K2_03,subfig:epsilon_K2_1} for level sets with $K=2$)\footnote{For illustrations with $K=1$, see \Cref{subfig:epsilon_K1_03,subfig:epsilon_K1_1}}.
\begin{definition}\label{def:epsilon_topk}
    We define $\ell_{\text{Noised bal.}}^{K, \epsilon}$ the noised balanced top-$K$ hinge loss as:
    \begin{align}
        \label{eq:bal-loss}
        \ell_{\text{Noised bal.}}^{K, \epsilon}(\bfs, y) = (1 + \topp_{K +1 , \epsilon}(\bfs) - s_{y})_{+} \enspace.
    \end{align}
\end{definition}
We call the former balanced as the margin (equal to 1) is the same for all $L$ classes. The parameter $\epsilon$ controls the variance of the noise added to the score vectors. When $\epsilon = 0$, we recover the top-$K$ calibrated loss of \citet{topk_yang}, $\ell_{\textrm{Cal.~Hinge}}^K$.
\begin{proposition}
    \label{prop:balanced}
    For a smoothing parameter $\epsilon>0$ and a label $y \in [L]$,
    \begin{tiny}$\bullet$\end{tiny} $\ell_{\text{Noised bal.}}^{K, \epsilon}(\cdot, y)$ is continuous, differentiable almost everywhere, with continuous derivative.
    \begin{tiny}$\bullet$\end{tiny} The gradient of $\ell (\cdot,y)\eqdef\ell_{\text{Noised bal.}}^{K, \epsilon}(\cdot, y)$ is given by:
    \begin{small}\begin{align}\label{eq:grad_balanced}
            \nabla \ell(\bfs, y) \! = \! \ind_{\{1 + \topp_{K +1 , \epsilon}(\bfs)  \geq s_{y}\}} \! \cdot \! ( \nabla\topp_{K+1, \epsilon}(\bfs) - \delta_{y}),
        \end{align}\end{small}
    where $\delta_{y}\in\bbR^L$ is the vector with 1 at coordinate $y$ and 0 elsewhere.
\end{proposition}


\textbf{Practical implementation}: As is, the proposed loss can not be used directly to train modern neural network architectures due to the expectation and remains a theoretical tool.
Following \cite{berthet2020learning}, we simply rely on a Monte Carlo method to estimate the expectation for both the loss and its gradient: we draw $B$ noise vectors $Z_{1}, \dots, Z_{B}$, with $Z_{b}\stackrel{\iid}{\sim} \cN(0,\Id_L)$ for $b \in [B]$.
The loss $\ell_{\text{Noised bal.}}^{K, \epsilon}$ is then estimated by:
\begin{align}
    \label{eq:estimated_balanced}
    \ell_{\text{Noised bal.}}^{K, \epsilon, B}(\bfs, y) & = (1 + \widehat{\topp}_{K +1 , \epsilon,B}(\bfs) - s_{y})_{+}\enspace,
\end{align}
where $\widehat{\topp}_{K + 1 , \epsilon,B}(\bfs) \eqdef \widehat\topsum_{K+1, \epsilon, B}(\bfs) - \widehat\topsum_{K, \epsilon,B}(\bfs)$
is a Monte Carlo estimate with $B$ samples:
\begin{align}
    \widehat\topsum_{K, \epsilon, B}(\bfs) = \frac{1}{B}\sum_{b=1}^{B} \topsum_{K}(\bfs + \epsilon Z_{b})\enspace.
    \label{eq:montecarlo}
\end{align}
We approximate $\nabla_\bfs\ell_{\text{Noised bal.}}^{K, \epsilon}(\bfs, y)$ by $G$, with:
\begin{align}\label{eq:gradient_topk}
    G \!=\! \mathbbm{1}_{\{1 + \widehat{\topp}_{K + 1 , \epsilon,B}(\bfs) \geq s_{y}\}} \!\! \cdot \! (\widehat{\nabla \topp}_{K+1 , \epsilon, B}(\bfs)\!-\! \delta_{y}),
\end{align}
where the Monte Carlo estimate
\begin{align*}
    \widehat{\nabla \topp}_{K+1 , \epsilon, B}(\bfs) \eqdef \widehat{\argtopp}_{K+1, \epsilon, B}(\bfs)
\end{align*}
is given by:
$$\widehat{\argtopp}_{K+1, \epsilon, B}(\bfs) =  \frac{1}{B}\sum_{b=1}^{B} \argtopp_{K+1}(\bfs + \epsilon Z_{b})\enspace.$$
We train our loss with SGD. Hence, $B$ repetitions are drawn each time the loss is evaluated.
The iterative nature of this process helps amplify the smoothing power of the approach, explaining why even small values of $B$ can lead to good performance (see \Cref{sec:experiments}).

\textbf{Illustration}.
Consider the case $L=4$, $K=2$, $B=3$, $\epsilon = 1.0$ with a score vector
$\bfs = \left[\begin{smallmatrix} 2.4 \\ 2.6 \\ 2.3 \\ 0.5 \end{smallmatrix}\right].$
We have $\topp_{K}(\bfs) = 2.4$ and $\argtopp_K(\bfs) = \left[\begin{smallmatrix} 1 \\ 0 \\ 0 \\ 0 \end{smallmatrix}\right]$ (the top-2 value of $\bfs$ corresponds to the first coordinate).
Assume the three noise vectors sampled are:
\begin{align*}
    Z_{1} =
    \left[
        \begin{smallmatrix*}[r] 0.2 \\ -0.1 \\ 0.1 \\ 0.3 \end{smallmatrix*}
        \right],
    \ Z_{2} =
    \left[
        \begin{smallmatrix*}[r] 0.1 \\ 0.1 \\ -0.1 \\ 0.1 \end{smallmatrix*}
        \right],
    \ Z_{3} =
    \left[
        \begin{smallmatrix*}[r] -0.1 \\ -0.1 \\ 0.1 \\ -0.1 \end{smallmatrix*}
        \right].
\end{align*}
The perturbed vectors are now:
\begin{align*}
    \bfs + \epsilon Z_{1}
    =\left[
        \begin{smallmatrix*}[r] 2.6 \\ 2.5 \\ 2.4 \\ 0.8 \end{smallmatrix*}
        \right],\
    \bfs + \epsilon Z_{2}
    = \left[
        \begin{smallmatrix*}[r] 2.5 \\ 2.7 \\ 2.2 \\ 0.6 \end{smallmatrix*}\right],\
    \bfs + \epsilon Z_{3}
    = \left[
        \begin{smallmatrix*}[r] 2.3 \\ 2.5 \\ 2.4 \\ 0.4 \end{smallmatrix*}\right].
\end{align*}
The induced perturbation may provoke a change in both $\topp_{K}$ and $\argtopp_K$.
For the perturbed vector $\bfs + \epsilon Z_{2}$, the added noise changes the top-$2$ value but it is still achieved at coordinate 1: $\topp_{K}(\bfs + \epsilon Z_{2}) = 2.5$ and $\argtopp_K(\bfs + \epsilon Z_{2}) = \left[\begin{smallmatrix*}[r] 1 \\ 0 \\ 0 \\ 0 \end{smallmatrix*}\right]$.
However, for $\bfs + \epsilon Z_{1}$ and $\bfs + \epsilon Z_{3}$, the added noise changes the coordinate at which the top-2 is achieved: $\argtopp_K(\bfs + \epsilon Z_{1}) = \left[\begin{smallmatrix*}[r] 0 \\ 1 \\ 0 \\ 0 \end{smallmatrix*}\right]$ and $\argtopp_K(\bfs + \epsilon Z_{3}) = \left[\begin{smallmatrix*}[r] 0 \\ 0 \\ 1 \\ 0 \end{smallmatrix*}\right]$, with $\topp_K(\bfs + \epsilon Z_{1}) = 2.5$ and $\topp_K(\bfs + \epsilon Z_{3}) = 2.4$, giving:
\begin{align*}
    \widehat{\topp}_{K , \epsilon, B}(s) =(2.5 + 2.5 + 2.4) / 3  = 2.47 \enspace,
\end{align*}
\begin{align*}
    \widehat{\nabla \topp}_{K , \epsilon, B}(s)
    =
    \frac{1}{3}
    \bigg(
    \left[
        \begin{smallmatrix*}[r] 0 \\ 1 \\ 0 \\ 0 \end{smallmatrix*}
        \right]
    +
    \left[
        \begin{smallmatrix*}[r] 1 \\ 0 \\ 0 \\ 0 \end{smallmatrix*}
        \right]
    +
    \left[
        \begin{smallmatrix*}[r] 0 \\ 0 \\ 1 \\ 0 \end{smallmatrix*}
        \right]
    \bigg)
    =
    \left[
        \begin{smallmatrix*}[r]
            \frac{1}{3} \\ \frac{1}{3} \\ \frac{1}{3} \\ 0
        \end{smallmatrix*}
        \right].
\end{align*}
We see the added noise results in giving weight to the gradient coordinates $k$ whose associated score $s_k$ is close to $\topp_{K}(\bfs)$ (in this example the first and third coordinates).
Note that if we set $\epsilon$ to a smaller value, \eg $\epsilon=0.1$, the added perturbation is not large enough to change the $\argtopp_{K}$ in the perturbed vectors,
leading to the same gradient as the non-smoothed top-$K$ operator:
$\widehat{\nabla \topp}_{K , 0.1}(s) =
    \left[\begin{smallmatrix*}[r] 1 \\ 0 \\ 0 \\ 0 \end{smallmatrix*}\right]$.
Hence, $\epsilon$ acts as a parameter which allows exploring coordinates $k$ whose score values $s_k$ are close to the top-$K$ score (provided that $\epsilon$ and/or $B$ are large enough).

\subsection{New loss for imbalanced top-$K$ classification}
\label{subsec:imbalanced}

In real world applications, a long-tailed distribution between the classes is often present, \ie few classes receive most of the annotated labels.
This occurs for instance in datasets such as Pl@ntNet-300K \cite{plantnet-300k} and Inaturalist \cite{inat2017}, where a few classes represent the vast majority of images.
For these cases, the performance of deep neural networks trained with the cross entropy loss is much lower for classes with a small numbers of images, see \cite{plantnet-300k}.


We present an extension of the loss presented in \Cref{subsec:balanced} to the imbalanced case.
This imbalanced loss is based on uneven margins \cite{scott, uneven_margin_svm, cost-sensitive-svm, Cao_Wei_Gaidon_Arechiga_Ma19}.
The underlying idea is to require larger margins for classes with few examples, which leads to a higher incurred loss for mistakes made on examples of the least common classes.

Imposing a margin $m_y$ parameter per class in \Cref{eq:estimated_balanced} leads to the following formulation:
\begin{align}\label{eq:imbalanced}
    \ell_{\text{Noised Imbal.}}^{K, \epsilon, B, m_y}(\bfs, y)
    =
    (m_y + \widehat{\topp}_{K +1 , \epsilon,B}(\bfs) - s_{y})_{+}\enspace.
\end{align}
Here, we follow \citet{Cao_Wei_Gaidon_Arechiga_Ma19} and set $m_{y} = C/n_{y}^{1/4}$, with $n_{y}$ the number of samples in the training set with class $y$, and $C$ a hyperparameter to be tuned on a validation set.

\renewcommand{\arraystretch}{1.2}
\begin{table}[t]
    \caption{Influence of $\epsilon$ on CIFAR-100 best validation top-5 accuracy obtained by training a DenseNet 40-40 with loss $\ell_{\text{Noised bal.}}^{K=5, \epsilon, B=10}$. The training procedure is the same as in \Cref{subsubsec:balanced_datasets}.}
    \vskip 0.1in
    \begin{tiny}
        \begin{tabular}{lcccccccc}
            \toprule
            $\epsilon$ & 0.0   & 1e-4  & 1e-3 & 1e-2  & 1e-1  & 1.0   & 10.0  & 100.0 \\ \midrule
            Top-5 acc. & 19.38 & 14.84 & 11.4 & 93.36 & 94.46 & 94.24 & 93.78 & 93.12 \\ \bottomrule
        \end{tabular}
    \end{tiny}
    \label{tab:tab-epsilon}
    \vskip -0.1in
\end{table}

\subsection{Comparisons of various top-$K$ losses}
\label{subsec:Comparisons of the different top-$K$ losses}
In \Cref{tab:losses}, we synthesize the various top-$K$ loss functions evoked above. To better understand their differences, \Cref{fig:simplices_k=2} provides a plot of the losses for $\bfs$ in the 2-simplex, for $K=2$ and $L=3$.
The correct label is set to be $y=3$ and corresponds to the vertex on top and in red.
\Cref{subfig:topk} shows the classical top-$K$ that we would ideally want to optimize. It has 0 error when $s_3$ is larger than the smallest coordinate of $\bfs$ (\ie is in the top-2) and 1 otherwise.
\Cref{subfig:ce} shows the cross-entropy, by far the most popular (convex) loss used in deep learning.
As mentioned by \citet{topk_yang}, the cross-entropy happens to be top-$K$ calibrated for all $K$.
\Cref{subfig:lapin} shows the top-$K$ hinge loss proposed by \citet{lapin2015} and \Cref{subfig:lapinconvex} is a convex upper relaxation.
Unfortunately, \citet{topk_yang} have shown that such losses are not top-$K$ calibrated and propose an alternative, illustrated in \Cref{subfig:yang}.
We show that the loss of \citet{topk_yang} performs poorly when used for optimizing a deep neural network.
\Cref{subfig:berrada0.01} and \Cref{subfig:berrada0.4} show the smoothing proposed by \citet{berrada} of the loss in \Cref{subfig:lapin}, while \Cref{subfig:epsilon_K2_03} and \Cref{subfig:epsilon_K2_1} show our proposed noised smoothing of the loss in \Cref{subfig:yang}.
The difference with \citet{berrada} is that we start with a top-$K$ calibrated hinge loss, and our smoothing consists in smoothing only the top-$K$ operator, which mostly affects classes whose scores are close to the top-$K$ score, while the method from \citet{berrada} results in a gradient where all coordinates are non-zero.

Finally, ~\Cref{subfig:epsilon_K2_Imbal_1} shows our noised imbalanced top-$K$ loss.
Additional visualizations of our noised top-$K$ loss illustrating the effect of $B$ and $\epsilon$ can be found in the appendix, see \Cref{fig:impact_B,fig:impact_epsilon}.



\section{Experiments}
\label{sec:experiments}
The Pytorch \citep{pytorch} code for our top-$K$ loss and experiments can be found at: \url{https://github.com/garcinc/noised-topk}.
\subsection{Influence of $\epsilon$ and gradient sparsity}
\label{subsec:noise-parameter}

\begin{figure}[t]
  \centering
  \includegraphics[width=0.48\textwidth]{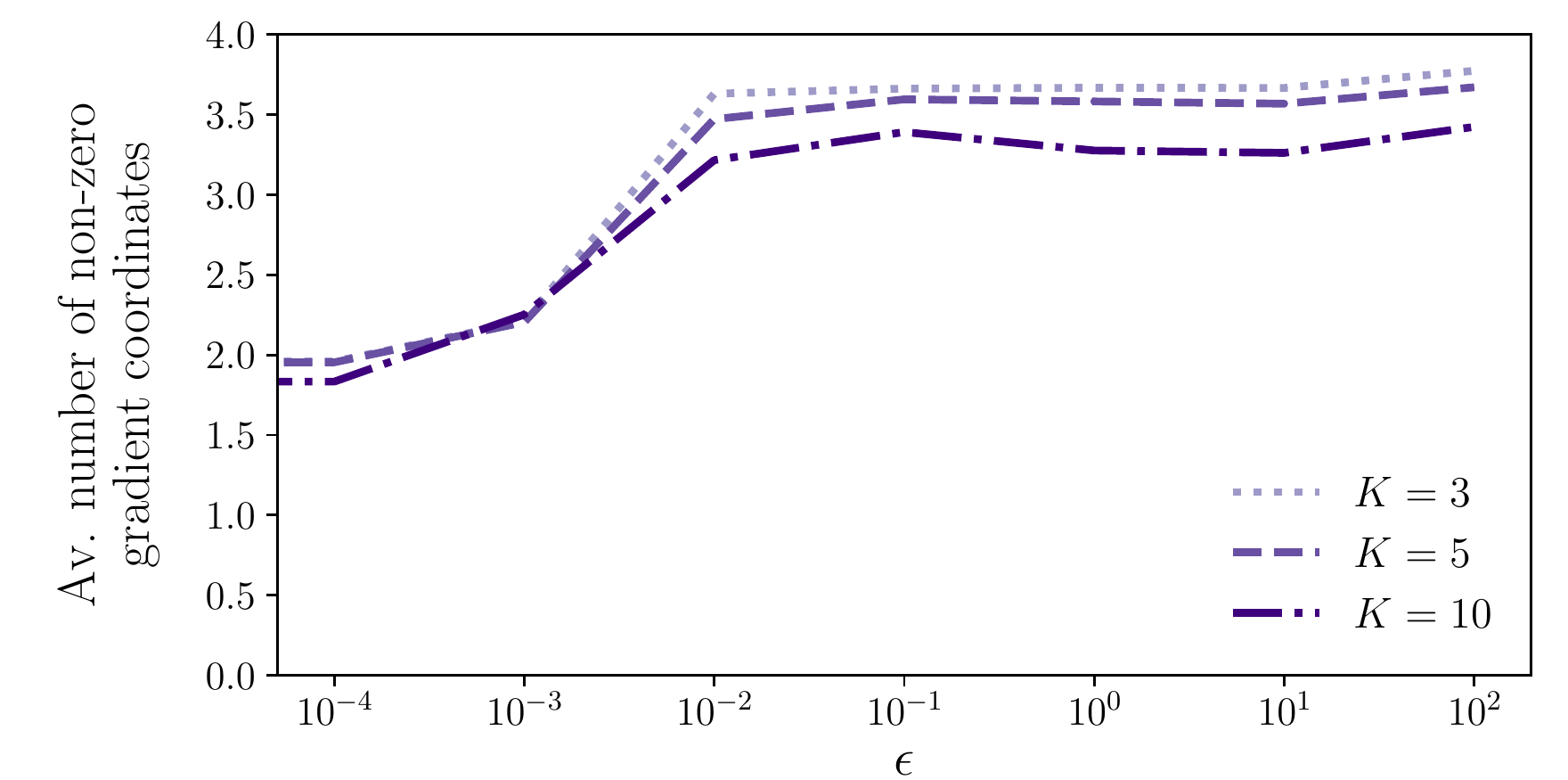}
  \caption{Average number of non-zero gradient coordinates as a function of $\epsilon$ (loss $\ell_{\text{Noised bal.}}^{K, \epsilon, 3}$, CIFAR-100 dataset, DenseNet 40-40 model, 1st epoch).
    The gradient dimension is 100. We see that the gradient remains sparse even for large values of $\epsilon$. Together with \Cref{tab:tab-epsilon}, this shows that having a non-sparse gradient is not a necessary condition for successful learning, contrary to what is suggested in \cite{berrada}}
  \label{fig:gradient}
\end{figure}

\Cref{tab:tab-epsilon} shows the influence of $\epsilon$ on CIFAR-100\footnote{For experiments with CIFAR-100, we consider a DenseNet 40-40 model \citep{densenet}, similarly as \citet{berrada}.} top-5 accuracy.
When $\epsilon = 0$, $\ell_{\text{Noised bal.}}^{K, \epsilon, B}$ coincides with  $\ell_{\textrm{Cal. Hinge}}^K$.
We see that a model trained with this loss fails to learn properly.
This resonates with the observation made by \citet{berrada} that  a model trained with $\ell_{\textrm{Hinge}}^K$, which is close to $\ell_{\textrm{Cal. Hinge}}^K$, also fails to learn.
\Cref{tab:tab-epsilon} also shows that when $\epsilon$ is too small ($\epsilon= 10^{-4}$ or $\epsilon= 10^{-3}$), the optimization remains difficult and the learned models have very low performance.
For sufficiently high values of $\epsilon$, in the order of $10^{-2}$ or greater, the smoothing is effective and the learned models achieve a very high top-5 accuracy. \Cref{tab:tab-epsilon} also shows that although the optimal value of $\epsilon$ appears to be around $10^{-1}$, the optmization is robust to high values of $\epsilon$.

\citet{berrada} argue that a reason a model trained with $\ell_{\textrm{Hinge}}^K$ fails to learn is because of the sparsity of the gradient of $\ell_{\textrm{Hinge}}^K$.
Indeed, for any $\bfs \in \bbR^L$, $y \in [L]$, $\nabla_{s} \ell_{\textrm{Hinge}}^K(\bfs, y)$ has at most two non-zero coordinates.
This is one of the main reasons put forward by the authors to motivate the smoothing of $\ell_{\textrm{Hinge}}^K$ into $\ell_{\text{Smoothed Hinge}}^{K, \tau}$, whose gradient coordinates are all non-zero.

We investigate the behaviour of the gradient of our loss by computing $\nabla_{s} \ell_{\text{Noised bal.}}^{K, \epsilon, B}(\bfs, y)$ for each training example during the first epoch.
We then compute the average number of non-zero coordinates in the gradient. We repeat this process for several values of $\epsilon$ and report the results in \Cref{fig:gradient}.
There are two points to highlight:

\begin{tiny}$\bullet$\end{tiny}
The number of non-zero gradients coordinates increases with $\epsilon$. This is consistent with our illustration example in \Cref{subsec:balanced}: high values of $\epsilon$ allow putting \textit{weights} on gradient coordinates whose score is close to the $\topp_{K}$ score.

\begin{tiny}$\bullet$\end{tiny}
Even when $\epsilon$ is large, the number of non-zero gradient coordinates is small: on average, 4 out of 100.
In comparison, for $\ell^K_{\textrm{CE}}$ and $\ell_{\text{Smoothed Hinge}}^{K, \tau}$, all gradient coordinates are non-zero.
Yet, even with such sparse gradient vectors, we manage to reach better top-5 accuracies than $\ell^K_{\textrm{CE}}$ and $\ell_{\text{Smoothed Hinge}}^{K, \tau}$ (see \Cref{table:cifar100-top5-acc}).
Therefore, one of the main takeaway is that a non-sparse gradient does not appear to be a necessary condition for successful learning contrary to what is suggested in \cite{berrada}.
A sufficiently high probability (controlled by $\epsilon$) that each coordinate is updated at training is enough to achieve good performance.

\subsection{Influence of $B$}
\label{subsec:B}
\begin{table}[]
  \caption{Influence of $B$ hyper-parameter on the best validation top-5 accuracy (loss $\ell_{\text{Noised bal.}}^{5, 0.2, B}$, CIFAR-100 dataset, DenseNet 40-40 model. The training procedure is the same as in \Cref{subsubsec:balanced_datasets}.)}
  \vskip 0.1in
  \scalebox{0.8}{
    \begin{tabular}{lccccccc}
      \toprule
      $B$       & 1     & 2    & 3     & 5     & 10    & 50    & 100   \\ \midrule
      Top-5 acc & 94.28 & 94.2 & 94.46 & 94.52 & 94.24 & 94.64 & 94.52 \\ \bottomrule
    \end{tabular}}
  \vskip -0.1in
  \label{tab:tab-B}
\end{table}

\Cref{tab:tab-B} shows the influence of the number of sampled standard normal random vectors $B$ on CIFAR-100 top-5 accuracy for a model  trained with our balanced loss with $K=5$.
$B$ appears to have little impact on top-5 accuracy, indicating that there is no need to precisely estimate the expectation in \Cref{eq:montecarlo}.
As increasing $B$ comes with computation overhead (see next section) and does not yield an increase of top-$K$ accuracy, we advise setting it to a small value (\eg $3 \leq B \leq 10)$.

\subsection{Computation time}
\label{subsec:time}

\begin{figure}[t]
  \centering
  \includegraphics[width=0.99\linewidth]{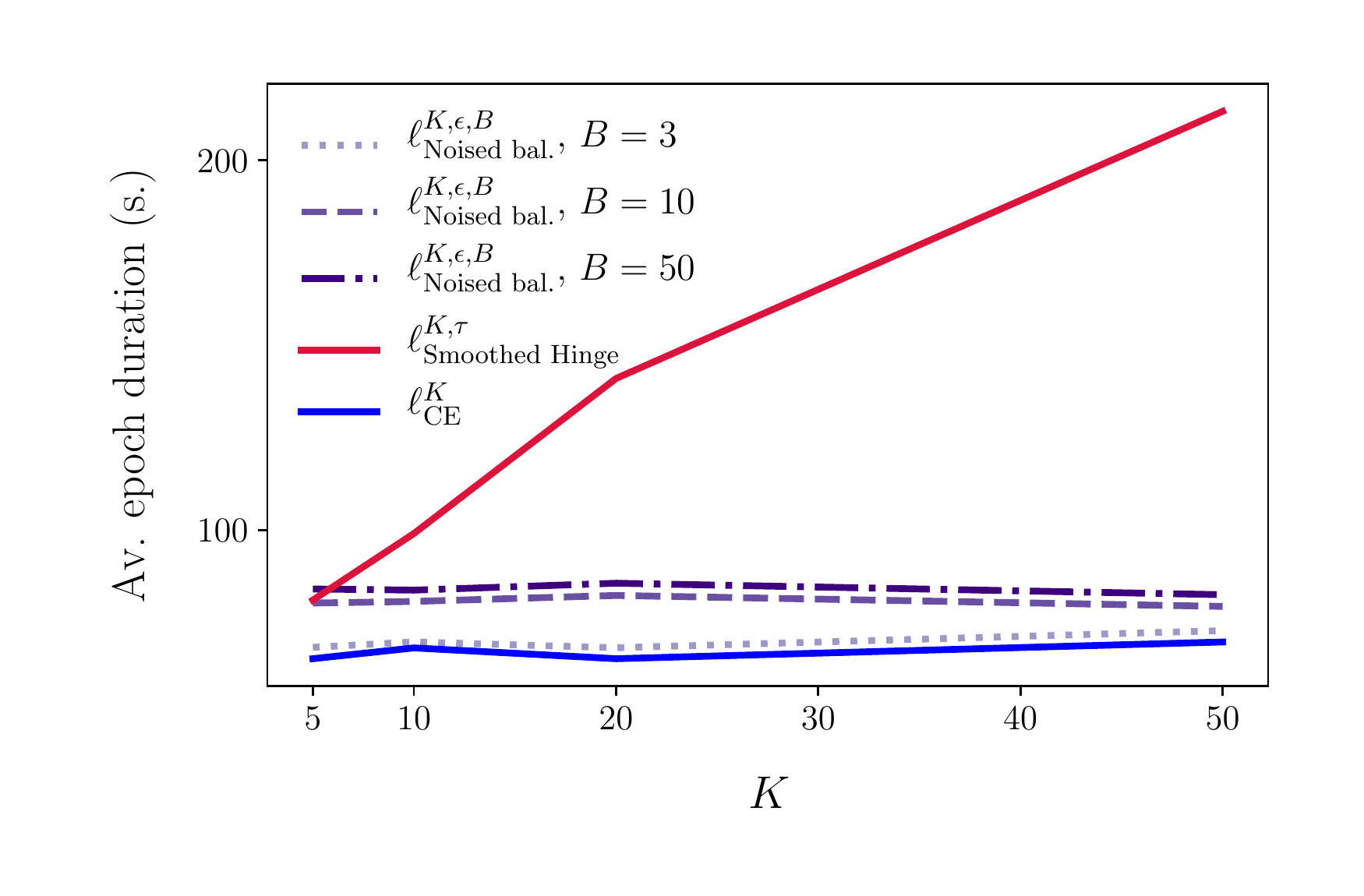}
  \caption{Average epoch time as a function of $K$ for different losses (CIFAR-100 dataset, DenseNet 40-40 model). The proposed loss $\ell_{\text{Noised bal.}}^{K, \epsilon, B}$ is not sensitive to the parameter $K$ contrarily to $\ell_{\text{Smoothed Hinge}}^{K, \tau}$ introduced by \citet{berrada}.}
  \label{fig:epoch-time}
\end{figure}

In \Cref{fig:epoch-time}, we plot the average epoch duration of a model trained with the cross entropy $\ell_{\textrm{CE}}$, the loss from \citet{berrada} $\ell_{\text{Smoothed Hinge}}^{K, \tau}$, and our balanced loss $\ell_{\text{Noised bal.}}^{K, \epsilon, B}$ (for several values of $B$) as a function of $K$.
For standard training values, \ie $K=5$, $B=3$, the average epoch time is 65s for $\ell_{\textrm{CE}}$, 68s for $\ell_{\text{Noised bal.}}^{K, \epsilon, B}$ (+4.6\% \wrt $\ell_{\textrm{CE}}$) and 81s for $\ell_{\text{Smoothed Hinge}}^{K, \tau}$ (+24.6\% \wrt $\ell_{\textrm{CE}}$).
\Cref{fig:epoch-time} further shows that while the average epoch duration of $\ell_{\text{Smoothed Hinge}}^{K, \tau}$ seems to scale linearly in $K$, our loss does not incur an increased epoch duration when $K$ increases.
Thus, for $K=10$, the average epoch time for $\ell_{\text{Smoothed Hinge}}^{K, \tau}$ is 90s versus 60s for $\ell_{\text{Noised bal.}}^{K, \epsilon, B}$ with $B =3$.
While for most classical datasets \cite{imagenet, cifar100} small values of $K$ are enough to achieve high top-$K$ accuracy,
for other applications high values of $K$ may be used \citep{covington2016deep, cole2020geolifeclef}, making our balanced loss computationally attractive.

\subsection{Comparisons for balanced classification}
\label{subsubsec:balanced_datasets}
The CIFAR-100 \cite{cifar100} dataset contains 60,000 images (50,000 images in the training set and 10,000 images in the test set) categorized in 100 classes.
The classes are grouped into 20 superclasses (\eg fish, flowers, people), each regrouping 5 classes.
Here we compare the top-$K$ accuracy of a deep learning model trained on CIFAR-100 with either our balanced loss $\ell_{\text{Noised bal.}}^{K, \epsilon, B}$, the cross entropy $\ell_{\textrm{CE}}$, or the loss from \citet{berrada}, $\ell_{\text{Smoothed Hinge}}^{K, \tau}$, which is the current state-of-the art top-$K$ loss for deep learning.
We repeat the experiment from Section 5.1 of \citet{berrada} to study how $\ell_{\text{Noised bal.}}^{K, \epsilon, B}$ reacts when noise is introduced in the labels. More precisely, for each training image, its label is sampled randomly within the same super-class with probability $p$.
Thus, $p=0$ corresponds to the original dataset and $p=0.5$ corresponds to a dataset where all the training examples have a label corresponding to the right superclass, but half of them (on average) have a different label than the original one. With such a dataset, a perfect top-5 classifier is expected to have a $100\%$ top-5 accuracy.
As in \citet{berrada} we extract 5000 images from the training set to build a validation set.
We use the same hyperparameters as \citet{berrada}: we train a DenseNet 40-40 \cite{densenet} for 300 epochs with SGD and a Nesterov momentum of 0.9. For the learning rate, our policy consists in starting with  value of 0.1 and dividing it by ten at epoch 150 and 225.
The batch size and weight decay are set to 64 and $1.10^{-4}$ respectively.
Following \citet{berrada}, the smoothing parameter $\tau$ of $\smash{\ell_{\text{Smoothed Hinge}}^{K, \tau}}$ is set to 1.0.
For $\ell_{\text{Noised bal.}}^{K, \epsilon, B}$, we set the noise parameter $\epsilon$ to 0.2 and the number of noise samples $B$ at 10.
We keep the models with the best top-5 accuracies on the validation set and report the top-5 accuracy on the test set in \Cref{table:cifar100-top5-acc}.
The results are averaged over four runs with four different random seeds (we give the 95\% confidence inverval). They show that the models trained with $\ell_{\text{Noised bal.}}^{5, 0.2, 10}$
give the best top-5 accuracies except when $p=0.1$ where it is slightly below $\ell_{\text{Smoothed Hinge}}^{5, 1.0}$.
We also observe that the performance gain over the cross entropy is significant in the presence of label noise (\ie for $p > 0)$.

We provide additional experiments on ImageNet \citep{imagenet} in \Cref{subsec:Additional experiments}.
\begin{table}[]
  \caption{Top-5 accuracy for different losses as a function of the label noise probability $p$ within the superclasses of CIFAR-100 (DenseNet 40-40 model).}
  \vskip 0.1in
  \centering
  \begin{scriptsize}
    \begin{tabular}{lccr}
      \toprule
      Label noise $p$ & $\ell^K_{\textrm{CE}}$ & $\ell_{\text{Smoothed Hinge}}^{5, 1.0} $ & $\ell_{\text{Noised bal.}}^{5, 0.2, 10}$ \\
      \midrule
      0.0             & $94.2 {\pm} 0.1$       & $93.3 {\pm} 0.0$                         & $\mathbf{94.4} {\pm} 0.1$                \\
      0.1             & $90.3 {\pm} 0.2$       & $\mathbf{92.2} {\pm} 0.3$                & $91.9 {\pm} 0.1$                         \\
      0.2             & $87.6 {\pm} 0.1$       & $90.4 {\pm} 0.2$                         & $\mathbf{90.7} {\pm} 0.5$                \\
      0.3             & $85.7 {\pm} 0.4$       & $88.8 {\pm} 0.1$                         & $\mathbf{89.7} {\pm} 0.1$                \\
      0.4             & $83.6 {\pm} 0.2$       & $87.4 {\pm} 0.1$                         & $\mathbf{87.8} {\pm} 0.6$                \\
      \bottomrule
    \end{tabular}
  \end{scriptsize}
  \label{table:cifar100-top5-acc}
  \vskip -0.1in
\end{table}
\subsection{Comparison for imbalanced classification}
\label{subsubsec:Imbalanced_datasets}

\subsubsection{Pl@ntNet-300K}
\label{subsubsubsec:p-300k}

\begin{table*}[]
  \caption{Macro-average top-$K$ accuracy (on test set) for different losses measured on Pl@ntNet-300K, a heavy-tailed dataset with high ambiguity (ResNet-50 model).
    The three numbers in parentheses represent respectively the mean top-$K$ accuracies of 1) few shot classes ($<20$ training images) 2) medium shot classes ($20 \leq . \leq 100$ training images) 3) many shot classes ($> 100$ training images).}
  \label{table:plantnet-300k}

  \vskip 0.1in
  \centering
  \begin{adjustbox}{width=0.99\textwidth}

    \begin{normalsize}
      \begin{tabular}{lcccccc}
        \toprule
        K & $\ell_{\textrm{CE}}$                             & $\ell_{\text{Smoothed Hinge}}^{K, 0.1}$          & $\ell_{\text{Noised bal.}}^{K, 1.0, 5}$          & focal ($\gamma =2.0$)                            & $\ell_{\textrm{LDAM}}^{\max m_y =0.2}$  & $\ell_{\text{Noised imbal.}}^{K, 0.01, 5,\max m_y=0.2}$            \\ \midrule
        1 & $36.3 {\pm} 0.3$ ($12.6$/$42.9$/$71.7$)          & $35.7 {\pm} 0.2$ ($13.1$/$41.5$/$71.1$)          & $35.8 {\pm} 0.3$ ($12.4$/$42.1$/$\mathbf{72.1}$) & $37.6 {\pm} 0.3$ ($15.5$/$43.4$/$71.4$)          & $40.6 {\pm} 0.1$ ($20.9$/$45.8$/$71.2$) & $\mathbf{42.4} {\pm} 0.3$ ($\mathbf{23.9}$/$\mathbf{46.3}$/$72.1$) \\
        3 & $58.8 {\pm} 0.4$ ($32.4$/$\mathbf{75.3}$/$92.0$) & $50.3 {\pm} 0.2$ ($16.7$/$69.8$/$\mathbf{92.7}$) & $58.7 {\pm} 0.4$ ($32.2$/$73.8$/$88.8$)          & $60.4 {\pm} 0.3$ ($35.9$/$74.8$/$92.0$)          & $63.3 {\pm} 0.3$ ($43.0$/$74.1$/$90.0$) & $\mathbf{64.9} {\pm} 0.4$ ($\mathbf{44.8}$/$74.5$/$92.1$)          \\
        5 & $68.7 {\pm} 0.2$ ($45.1$/$\mathbf{86.3}$/$95.4$) & $50.9 {\pm} 0.3$ ($12.1$/$78.1$/$95.7$)          & $66.4 {\pm} 0.5$ ($42.0$/$82.5$/$95.5$)          & $69.7 {\pm} 0.2$ ($47.5$/$84.8$/$\mathbf{95.8}$) & $71.9 {\pm} 0.3$ ($54.0$/$83.0$/$94.0$) & $\mathbf{73.2} {\pm} 0.5$ ($\mathbf{55.3}$/$84.2$/$95.3$)          \\
      \end{tabular}
    \end{normalsize}
  \end{adjustbox}
  \vskip -0.1in
\end{table*}

\begin{table*}[]

  \caption{Top-$K$ accuracy (on test set) for different losses measured on ImageNet-LT (ResNet-34 model).
    The three numbers in parentheses represent respectively the mean top-$K$ accuracies of 1) few shot classes ($<20$ training images) 2) medium shot classes ($20 \leq . \leq 100$ training images) 3) many shot classes ($> 100$ training images).}
  \label{table:imagenet-lt}

  \vskip 0.1in
  \centering
  \begin{adjustbox}{width=0.99\textwidth}

    \begin{normalsize}
      \begin{tabular}{lcccccc}
        \toprule
        K & $\ell_{\textrm{CE}}$                            & $\ell_{\text{Smoothed Hinge}}^{K, 0.1}$         & focal ($\gamma = 1.0$)                                             & $\ell_{\textrm{LDAM}}^{\max m_y =0.4}$                             & $\ell_{\text{Noised imbal.}}^{K, 0.1, 5, \max m_y =0.4}$  \\ \midrule
        1 & $37.0 {\pm} 0.1$ ($1.5$/$28.2$/$60.6$)          & $37.3 {\pm} 0.1$ ($1.3$/$28.6$/$\mathbf{60.7}$) & $37.7 {\pm} 0.0$ ($2.4$/$29.8$/$59.9$)                             & $\mathbf{39.3} {\pm} 0.2$ ($\mathbf{10.5}$/$\mathbf{33.1}$/$57.1$) & $38.7 {\pm} 0.0$ ($7.6$/$32.3$/$57.6$)                    \\
        3 & $55.5 {\pm} 0.1$ ($8.2$/$53.0$/$\mathbf{75.3}$) & $42.0 {\pm} 0.1$ ($0.0$/$29.1$/$72.9$)          & $56.2 {\pm} 0.0$ ($10.2$/$\mathbf{54.0}$/$75.1$)                   & $56.0 {\pm} 0.2$ ($24.1$/$52.0$/$72.3$)                            & $\mathbf{56.5} {\pm} 0.1$ ($\mathbf{27.0}$/$52.6$/$71.7$) \\
        5 & $63.2 {\pm} 0.1$ ($15.8$/$63.1$/$80.1$)         & $39.0 {\pm} 0.1$ ($0.0$/$20.7$/$75.5$)          & $\mathbf{63.8} {\pm} 0.1$ ($17.8$/$\mathbf{63.6}$/$\mathbf{80.3}$) & $63.1 {\pm} 0.2$ ($32.9$/$60.1$/$77.7$)                            & $63.5 {\pm} 0.1$ ($\mathbf{37.0}$/$60.2$/$77.0$)          \\
      \end{tabular}
    \end{normalsize}
  \end{adjustbox}
  \vskip -0.1in

\end{table*}
We consider Pl@ntNet-300K\footnote{For the experiments with Pl@ntNet-300K, we consider a ResNet-50 model \citep{resnet}.}, a dataset of plant images recently introduced in \cite{plantnet-300k}.
It consists of 306,146 plant images distributed in 1,081 species (the classes).
The particularities of the dataset are its long-tailed distribution (80\% of the species with the least number of images account for only 11\% of the total number of images)
and the class ambiguity: many species are visually similar.
For such an imbalanced dataset, accuracy and top-$K$ accuracy mainly reflect the performance of the model on the few classes representing the vast majority of images.
Often times, we also want the model to yield satisfactory results on the classes with few images.
Therefore, for this dataset we report macro-average top-$K$ accuracy,
which is obtained by computing top-$K$ accuracy for each class separately and then taking the average over classes.
Thus, the class with only a few images contributes the same as the class with thousands of images to the overall result.

In this section we compare the macro-average top-$K$ accuracy of a deep neural network trained with either $\smash{\ell_{\textrm{CE}}, \ell_{\text{Smoothed Hinge}}^{K, \tau} , \ell_{\text{Noised bal.}}^{K, \epsilon, B}, \ell_{\text{Noised imbal.}}^{K, \epsilon, B,m_y}}$ and $\smash{\ell_{\textrm{LDAM}}^{m_y}}$, the loss from \citet{Cao_Wei_Gaidon_Arechiga_Ma19} based on uneven margins providing state-of-the-art performance in Fine-Grained Visual Categorization tasks.

\textbf{Setup}: We train a ResNet-50 \cite{resnet} pre-trained on ImageNet \cite{imagenet} for 30 epochs with SGD with a momentum of 0.9 with the Nesterov acceleration.
We use a learning rate of $2.10^{-3}$ divided by ten at epoch 20 and epoch 25.
The batch size and weight decay are set to 32 and $1.10^{-4}$ respectively.
The smoothing parameter $\tau$ for $\smash{\ell_{\text{Smoothed Hinge}}^{K, \tau}}$ is set to 0.1.

To tune the margins of $\smash{\ell_{\textrm{LDAM}}^{m_y}}$ and $\ell_{\text{Noised imbal.}}^{K, \epsilon, B,m_y}$ more easily, we follow \cite{softmax_margin, Cao_Wei_Gaidon_Arechiga_Ma19}: we normalize the last hidden activation and the weight vectors of the last fully-connected layer to both have unit $\ell_{2}$-norm, and we  multiply the scores by a scaling constant, tuned for both losses on the validation set, leading to 40 for $\smash{\ell_{\textrm{LDAM}}^{m_y}}$ and 60 for $\smash{\ell_{\text{Noised imbal.}}^{K, \epsilon, B,m_y}}$.

Finally, we tune the constant $C$ by tuning the largest margin $\max_{y\in[L]} m_y$ for both $\smash{\ell_{\textrm{LDAM}}^{m_y}}$ and $\ell_{\text{Noised imbal.}}^{K, \epsilon, B,m_y}$.
We find that for both losses, the optimal largest margin is 0.2.
For $\smash{\ell_{\text{Noised imbal.}}^{K, \epsilon, B,m_y}}$, we set  $\epsilon = 10^{-2}$ and $B=5$. We further discuss hyperparameter tuning in \Cref{subsec:hyperparams}.

We train the network with the top-$K$ losses $\smash{\ell_{\text{Smoothed Hinge}}^{K, \tau}, \ell_{\text{Noised bal.}}^{K, \epsilon, B}}$ and $\smash{\ell_{\text{Noised imbal.}}^{K, \epsilon, B,m_y}}$ for $K \in \{1, 3, 5\}$.
For all losses, we perform early stopping based on the best macro-average top-$K$ accuracy on the validation set (for $K \in \{1, 3, 5\}$).
We report the results on the test set in \Cref{table:plantnet-300k} (three seeds, 95\% confidence interval).
We find that the loss from \citet{berrada} fails to generalize to the tail classes in such an imbalanced setting.
In contrast, $\ell_{\text{Noised bal.}}^{K, \epsilon, B}$ gives results similar to the cross-entropy while $\ell_{\text{Noised imbal.}}^{K, \epsilon, B,m_y}$ provides the best results (regardless of the value of $K$).
Noticeably, it outperforms $\smash{\ell_{\textrm{LDAM}}^{m_y}}$ \citep{Cao_Wei_Gaidon_Arechiga_Ma19} for all cases.

\subsubsection{ImageNet-LT}
\label{subsubsubsec:imagenet-lt}

We test $\ell_{\text{Noised imbal.}}^{K, \epsilon, B,m_y}$ on ImageNet-LT \citep{open-world}, a dataset of 115,836 training images obtained by subsampling images from ImageNet with a pareto distribution.
The resulting class imbalance is much less pronounced than for Pl@ntNet-300K.

We train a ResNet34 with several losses for $K \in \{1, 3, 5\}$ for 100 epochs with a learning rate of $1.10^{-2}$ divided by 10 at epoch 60 and 80.
We use a batch size of 128 and set the weight decay to $2.10^{-3}$. All hyperparameters are tuned on the 20,000 images validation set from \citet{open-world}.
The results on the test set (four seeds, 95\% confidence interval) are reported in \Cref{table:imagenet-lt}.
They show that $\ell_{\text{Noised Imbal.}}^{K, \epsilon ,B, m_y}$ performs very well on few shot classes compared to the other losses. Since ImageNet-LT is much less imbalanced than Pl@ntNet-300K, there are fewer such classes, hence the overall gain is less salient than for Pl@ntNet-300K.


\section{Conclusion and perspectives}
\label{sec:conclusion_and_perspectives}
We propose a novel top-$K$ loss as a smoothed version of the top-$K$ calibrated hinge loss of \citet{topk_yang}.
Our loss function is well suited for training deep neural networks, contrarily to the original top-$K$ calibrated hinge loss (\eg the poor performance of the  case $\epsilon=0$ in \Cref{tab:tab-epsilon}, that reduces to their loss).
The smoothing procedure we propose applies the perturbed optimizers framework to smooth the top-$K$ operator.
We show that our loss performs well compared to the current state-of-the-art top-$K$ losses for deep learning while being significantly faster to train when $K$ increases.
At training, the gradient of our loss \wrt the score is sparse,
showing that non-sparse gradients are not necessary for successful learning.
Finally, a slight adaptation of our loss for imbalanced datasets (leveraging uneven margins) outperforms other baseline losses.
Studying deep learning optimization methods for other set-valued classification tasks, such as \textit{average size control} or \textit{point-wise error control} \cite{chzhen2021set} are left for future work.


\section*{Acknowledgements}
The work by CG and JS was supported in part by the French National Research Agency (ANR) through the grant ANR-20-CHIA-0001-01 (Chaire IA CaMeLOt).


\bibliography{arxiv}
\bibliographystyle{arxiv}

\newpage
\appendix
\onecolumn

\section{Reminder on Top-$K$ calibration}
\label{sec:Reminder on Top-$K$ calibration}
Here we provide some elements introduced by \citet{topk_yang} on top-$K$ calibration.

\begin{definition}
    \citep[Definition 2.3]{topk_yang}\textbf{.}
    For a fixed $K\in[L]$, and given $\bfy \in \bbR^{L}$ and $\tilde{\bfy} \in \bbR^{L}$, we say that $\bfy$ is top-$K$ preserving \wrt $\tilde{\bfy}$, denoted $P_{K}(\bfy, \tilde{\bfy})$,
    if for all $k \in [L]$,
    \begin{align}
        \tilde{y}_{k} > \topp_{K+1}(\tilde{\bfy}) & \implies {y}_{k} > \topp_{K+1}({\bfy})       \\
        \tilde{y}_{k} < \topp_{K}(\tilde{\bfy})   & \implies y_{k} < \topp_{K}({\bfy}) \enspace.
    \end{align}
    The negation of this statement is $\neg P_{k}(\bfy, \tilde{\bfy})$.
\end{definition}

We let $\Delta_L\eqdef \{\bfpi \in \bbR^L: \sum_{k\in [L]}\pi_k=1, \pi_k \geq 0\}$ denote the probability simplex of size $L$.
For a score $\bfs\in\bbR^L$ and $\bfpi \in \Delta_L $ representing the conditional distribution of $y$ given $x$, we write the conditional risk at $x\in\cX$ as $\cR_{\ell|x}(\bfs, \bfpi)=\bbE_{y|x\sim\pi}(\ell(\bfs,y))$ and the (integrated) risk as $\cR_{\ell}(f)\eqdef\bbE_{(x,y)\sim \bbP}[\ell(f(x),y)]$ for a scoring function $f : \cX \to \bbR^{L}$.
The associated Bayes risks are defined respectively by $\cR_{\ell|x}^*(\bfpi)\eqdef  \inf _{\bfs \in \mathbb{R}^{L}} \cR_{\ell|x}(\bfs, \bfpi)$ (respectively by $\cR_{\ell}^*\eqdef \inf_{f: \cX \to \bbR^L}\cR_{\ell}(f)$) .
\begin{definition}\label{def:topk-calibrated}
    \citep[Definition 2.4]{topk_yang}\textbf{.}
    A loss function $\ell: \bbR^{L} \times \cY \rightarrow \bbR$ is top-$K$ calibrated if for all $\bfpi \in \Delta_{L}$ and all $x \in \cX$:
    \begin{align}
        \inf _{\bfs \in \mathbb{R}^{L}: \neg P_{k}(\bfs, \bfpi)}  \cR_{\ell|x}(\bfs, \bfpi)\! >\cR_{\ell|x}^{*}(\bfpi)  \enspace.
    \end{align}
\end{definition}
In other words, a loss is calibrated if the infimum can only be attained among top-$K$ preserving vectors w.r.t. the conditional probability distribution.
\begin{theorem}
    Suppose $\ell$ is a nonnegative top-$K$ calibrated loss function. Then $\ell$ is top-$K$ consistent, \ie for any sequence of measurable
    functions $f^{(n)} : \cX \rightarrow \bbR^{L}$, we have:
    \begin{align*}
        \cR_{\ell}\left(f^{(n)}\right) \rightarrow \cR_{\ell}^{*} \Longrightarrow \cR_{\ell^K}\left(f^{(n)}\right) \rightarrow \cR_{\ell^K}^{*}\enspace.
    \end{align*}
\end{theorem}
In their paper, \citet{topk_yang} propose a slight modification of the multi-class hinge loss $    \ell_{\textrm{Hinge}}^K$ and show that it is top-$K$ calibrated:
\begin{align}
    \label{eq:hinge yang}
    \ell_{\textrm{Cal.~Hinge}}^K(\bfs, y) = (1 + \topp_{K+1}(\bfs) - s_{y})_{+}\enspace.
\end{align}


\section{Proofs and technical lemmas}
\label{subsec:Technical lemmas}

\subsection{Proof of \Cref{prop:smoothness}}
\label{app:prop_smoothness}

\begin{proof}
    We define $\smash{\cC_{K} \eqdef \left\{\bfz \in \bbR^L, \sum_{k\in[L]} z_{k} = K, 0 \leq z_{k} \leq 1, \: \forall k \in [L]\right\}}$.
    For $\bfs\in\bbR^L$, one can check that $\argtops_{K}(\bfs) = \argmax_{\bfz \in \cC_{K}} \langle \bfz, \bfs \rangle$. Recall that $Z$ is a standard normal random vector, \ie $Z\sim\mathcal{N}(0,\Id_L)$.

    \begin{itemize}
        \item $\cC_{K}$ is a convex polytope and the multivariate normal has positive differentiable density.
              So, we can apply \citep[Proposition 2.2]{berthet2020learning}.
              For that, it remains to determine the constant $R_{\cC_{K}}$ and $M_{\mu}$.
              First,  $R_{\cC_{K}} \eqdef \max_{\bfz \in \cC_{K}} \norm{\bfz}$.
              For simplicity, let us compute $\max_{\bfz \in \cC_{K}} \norm{\bfz}^{2}$, \ie
              \begin{align}
                   & \max  \norm{\bfz}^{2} \\ &\st \sum_{k \in [L]} z_{k} = K, \, \forall k \in [L], z_k \in [0,1].
              \end{align}
              Note that this corresponds to the well known quadratic knapsack problem.
              A numerical solution can be obtained, see for instance \cite{quad_knapsack}.
              To obtain our bound,  note that for $z\in\cC_K$ one can check that
              \begin{align*}
                  \norm{\bfz}^2= & \sum_{k\in[L]} z_k^2 \leq  \sum_{k \in [L]} z_k \quad ( \text{since } \, \forall k \in [L], z_k \in [0,1]).
              \end{align*}
              Hence, we have  $\forall \bfz \in\cC_K, \norm{\bfz}^2 \leq K$.
              Now, one can check that this equality is achieved when choosing $\bfz = (1,\dots,1,0,\dots,0)^\top \in \bbR^{L}$ with $K$ non-zeros values, yielding $\max_{\bfz \in \cC_{K}} \norm{\bfz} = \sqrt{K}$.
              Following \citep[Proposition 2.2]{berthet2020learning} this guarantees that $\topsum_{K, \epsilon}$ is $\sqrt{K}$-Lipschitz.

              Let us show now that $M_{\mu} \eqdef \sqrt{\bbE_{Z} [\norm{\nabla_{Z} \nu(Z)}^{2}]} = \sqrt{L}$ with $\nu(Z) = \frac{1}{2} \norm{Z}^{2}$ and $Z \sim \cN(0, \Id_{L})$. Hence, $M_{\mu}$ can be computed as:
              \begin{align*}
                  M_{\mu} & = \sqrt{\bbE_{Z} \left[\norm{\nabla_{Z} \nu(Z)}^{2}\right]} \\
                          & = \sqrt{\bbE_{Z} \left[\norm{Z}^{2}\right]}                 \\
                          & = \sqrt{L} \enspace,
              \end{align*}
              where the last equality comes from $Z \sim \cN(0, \Id_{L})$.
              Following \citep[Proposition 2.2]{berthet2020learning} this guarantees that $\nabla \topsum_{K, \epsilon}$ is $\tfrac{\sqrt{KL}}{\epsilon}$-Lipschitz.
        \item The last bullet of our proposition comes derives now directly from of \citep[Proposition 2.3]{berthet2020learning}.
    \end{itemize}
\end{proof}

\captionsetup[subfigure]{justification=justified,singlelinecheck=false}
\begin{figure*}[t]
    \centering
    \begin{subfigure}[b]{0.187\textwidth}
        \centering
        \includegraphics[width=\textwidth]{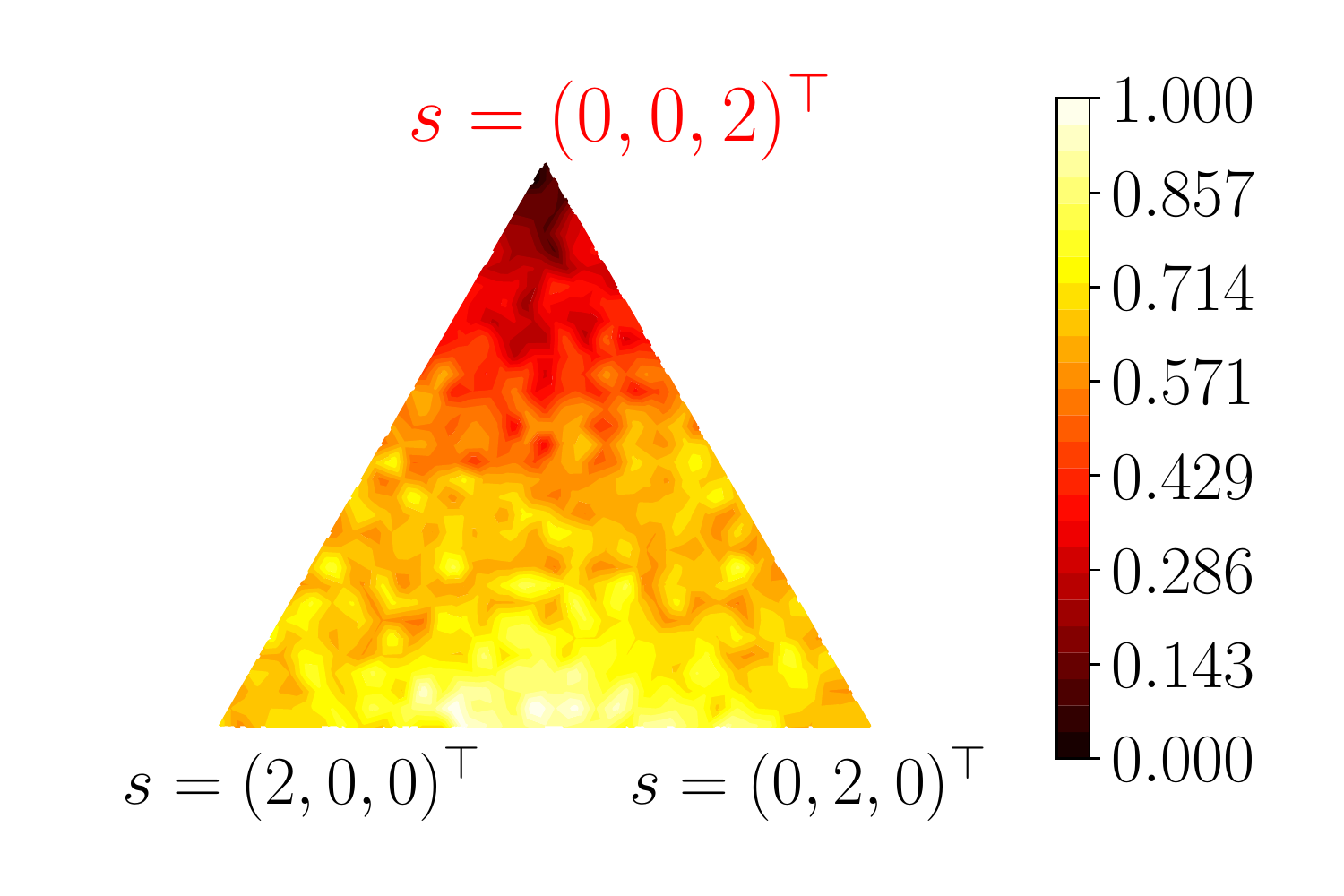}
        \caption{$B=1$}
        \label{subfig:k1B1}
    \end{subfigure}
    \hspace{0.1cm}
    \begin{subfigure}[b]{0.187\textwidth}
        \centering
        \includegraphics[width=\textwidth]{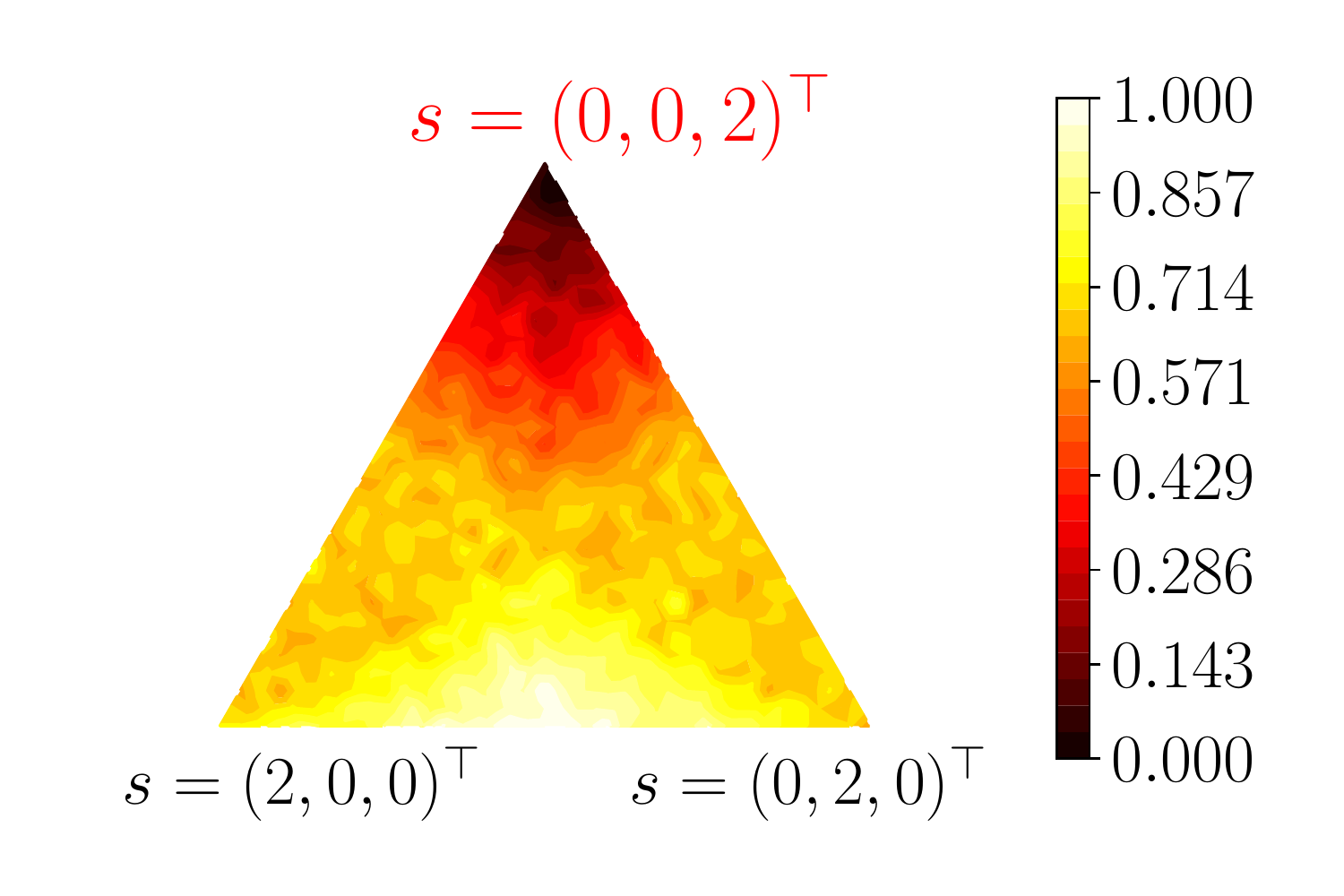}
        \caption{$B=5$}
        \label{subfig:k1B5}
    \end{subfigure}
    \hspace{0.1cm}
    \begin{subfigure}[b]{0.187\textwidth}
        \centering
        \includegraphics[width=\textwidth]{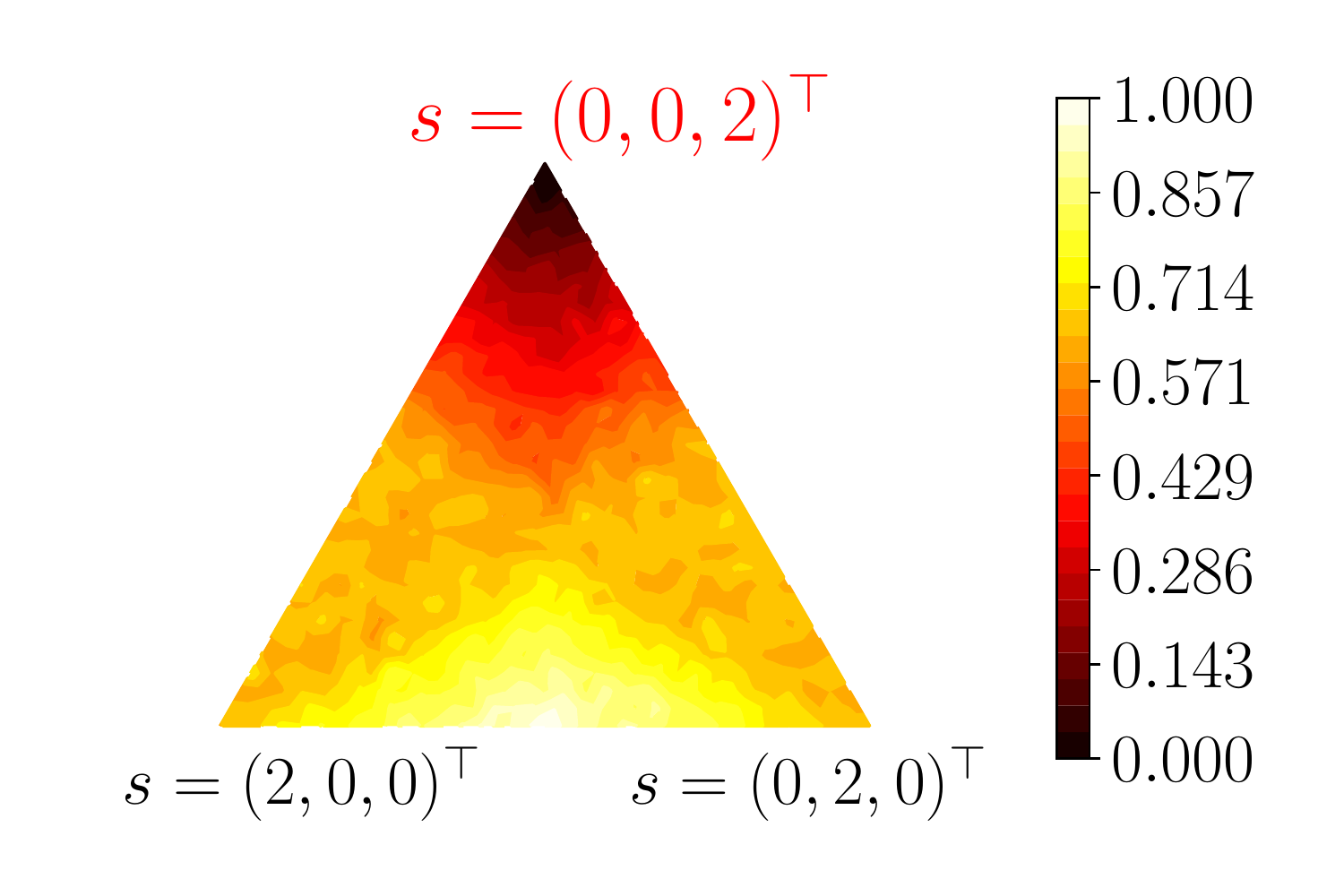}
        \caption{$B=10$}
        \label{subfig:k110}
    \end{subfigure}
    \hspace{0.1cm}
    \begin{subfigure}[b]{0.187\textwidth}
        \centering
        \includegraphics[width=\textwidth]{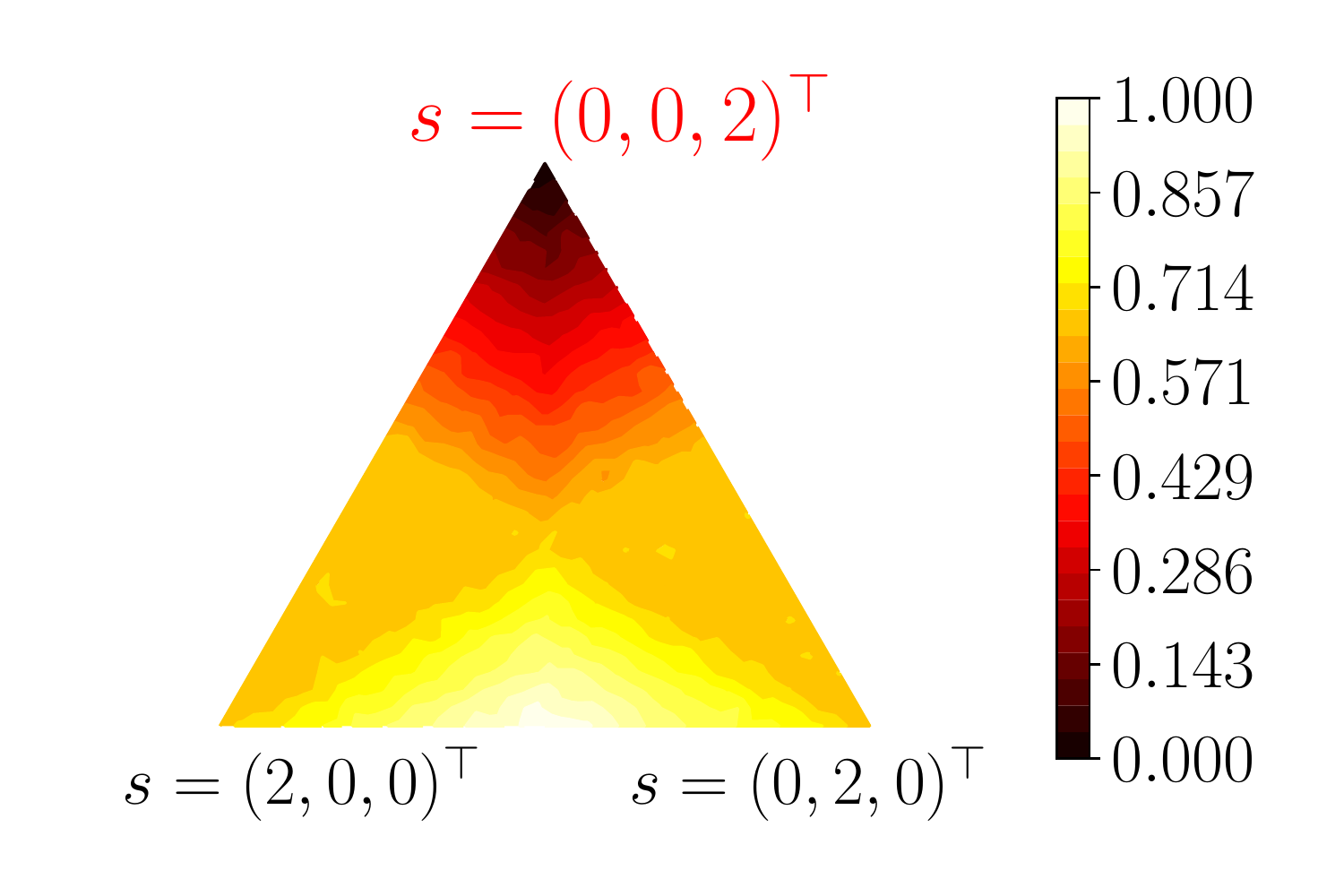}
        \caption{$B=50$}
        \label{subfig:k150}
    \end{subfigure}
    \hspace{0.1cm}
    \begin{subfigure}[b]{0.187\textwidth}
        \centering
        \includegraphics[width=\textwidth]{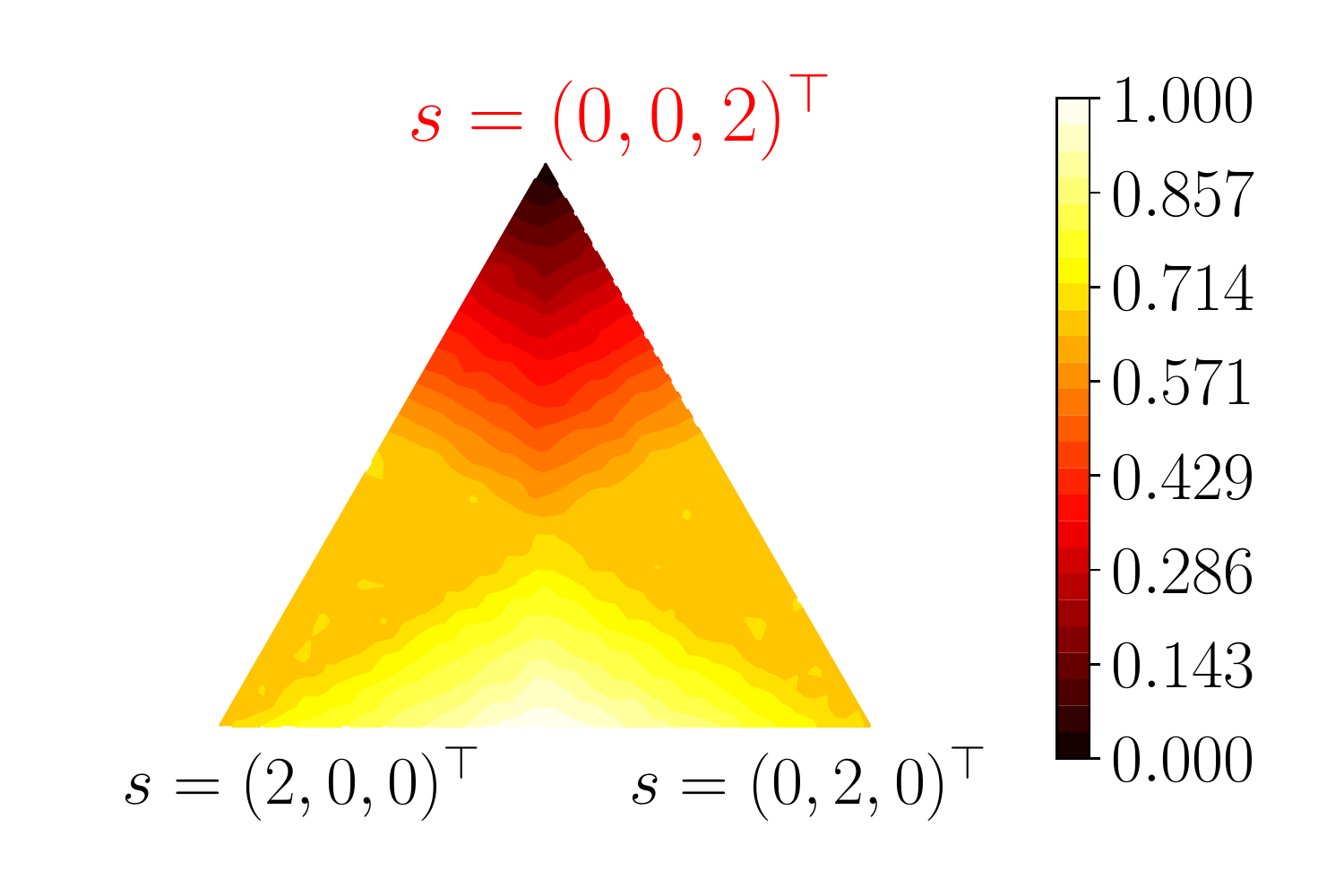}
        \caption{$B=100$}
        \label{subfig:k1B100}
    \end{subfigure}\\
    \begin{subfigure}[b]{0.187\textwidth}
        \centering
        \includegraphics[width=\textwidth]{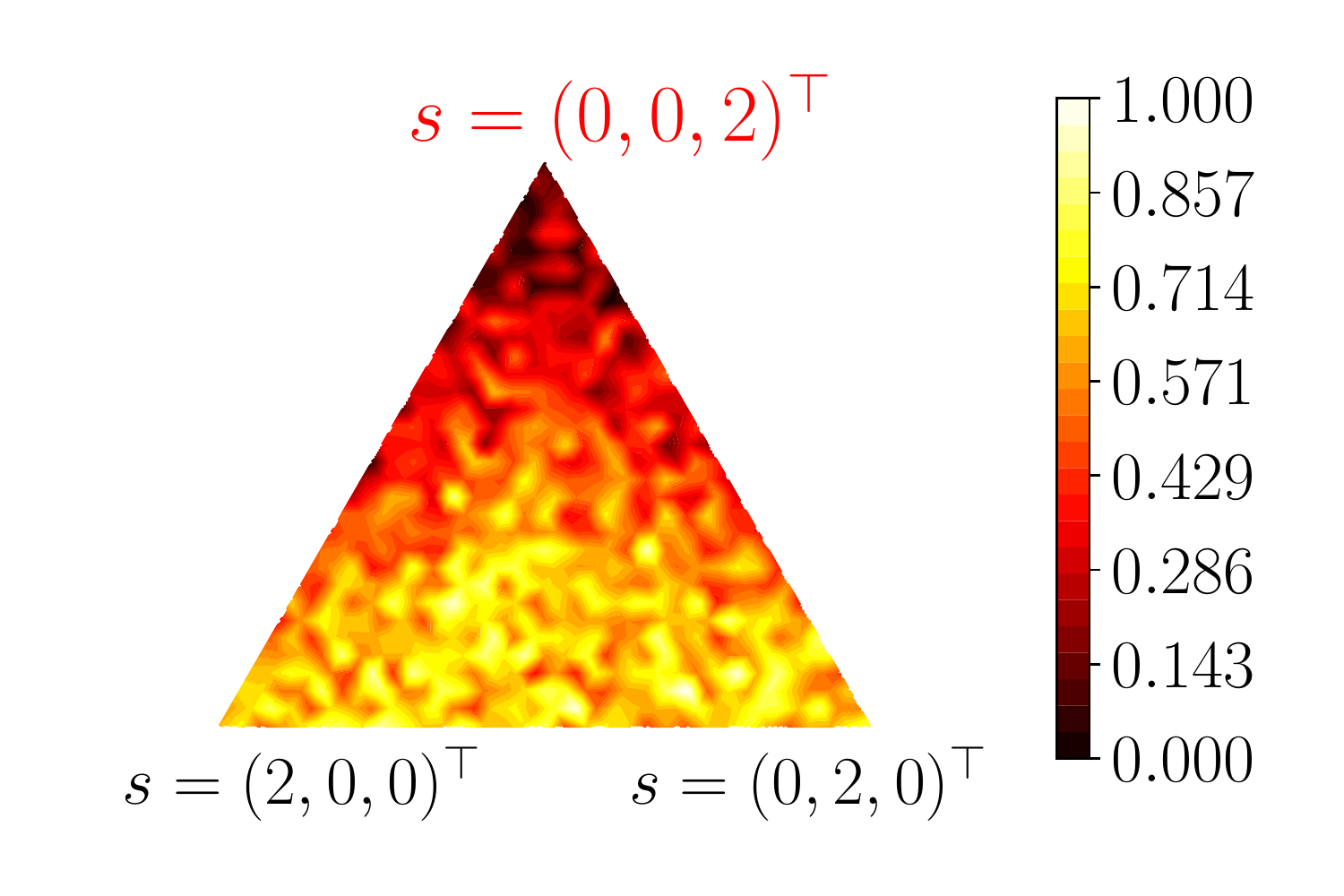}
        \caption{$B=1$}
        \label{subfig:k1B1}
    \end{subfigure}
    \hspace{0.1cm}
    \begin{subfigure}[b]{0.187\textwidth}
        \centering
        \includegraphics[width=\textwidth]{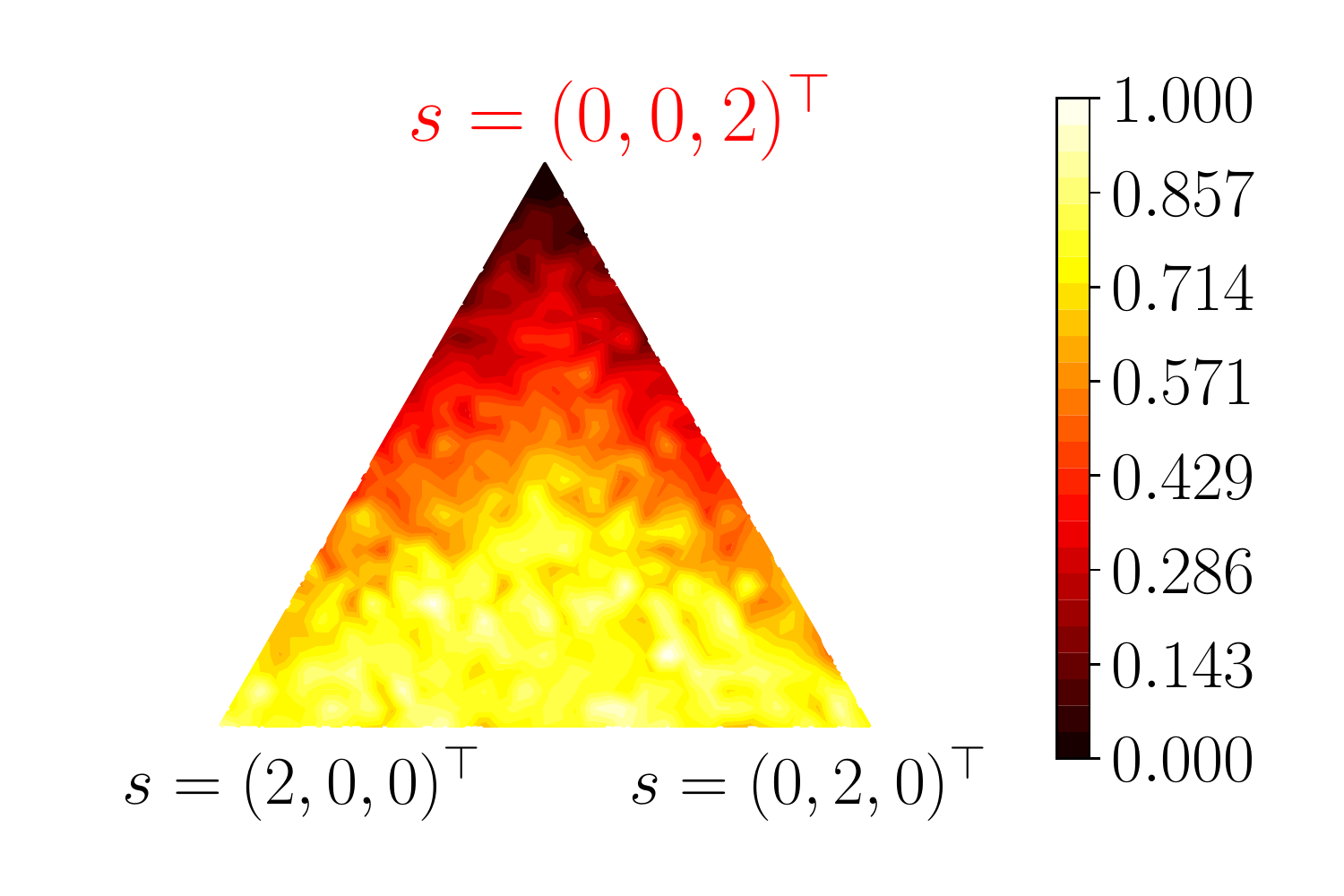}
        \caption{$B=5$}
        \label{subfig:k1B5}
    \end{subfigure}
    \hspace{0.1cm}
    \begin{subfigure}[b]{0.187\textwidth}
        \centering
        \includegraphics[width=\textwidth]{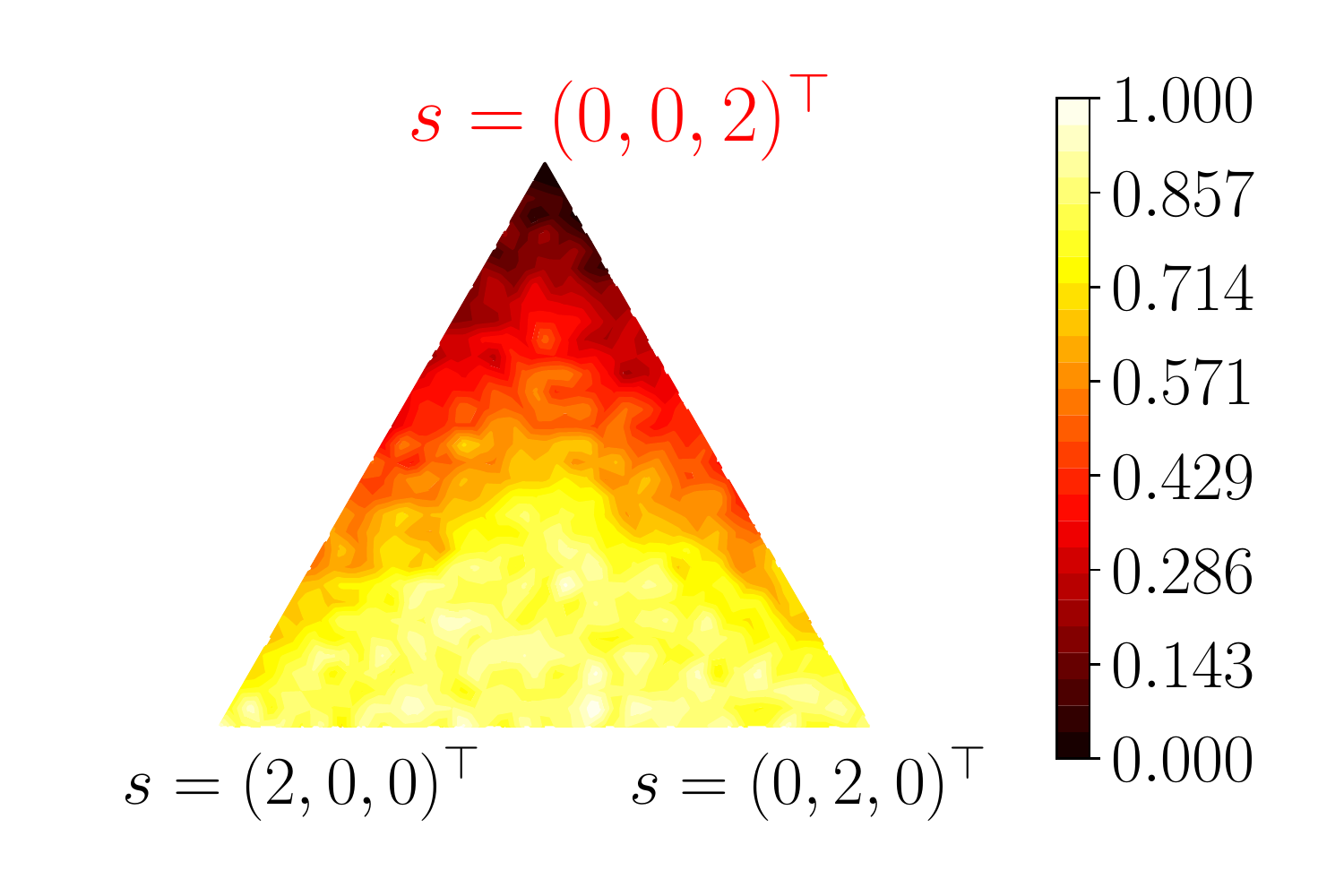}
        \caption{$B=10$}
        \label{subfig:k110}
    \end{subfigure}
    \hspace{0.1cm}
    \begin{subfigure}[b]{0.187\textwidth}
        \centering
        \includegraphics[width=\textwidth]{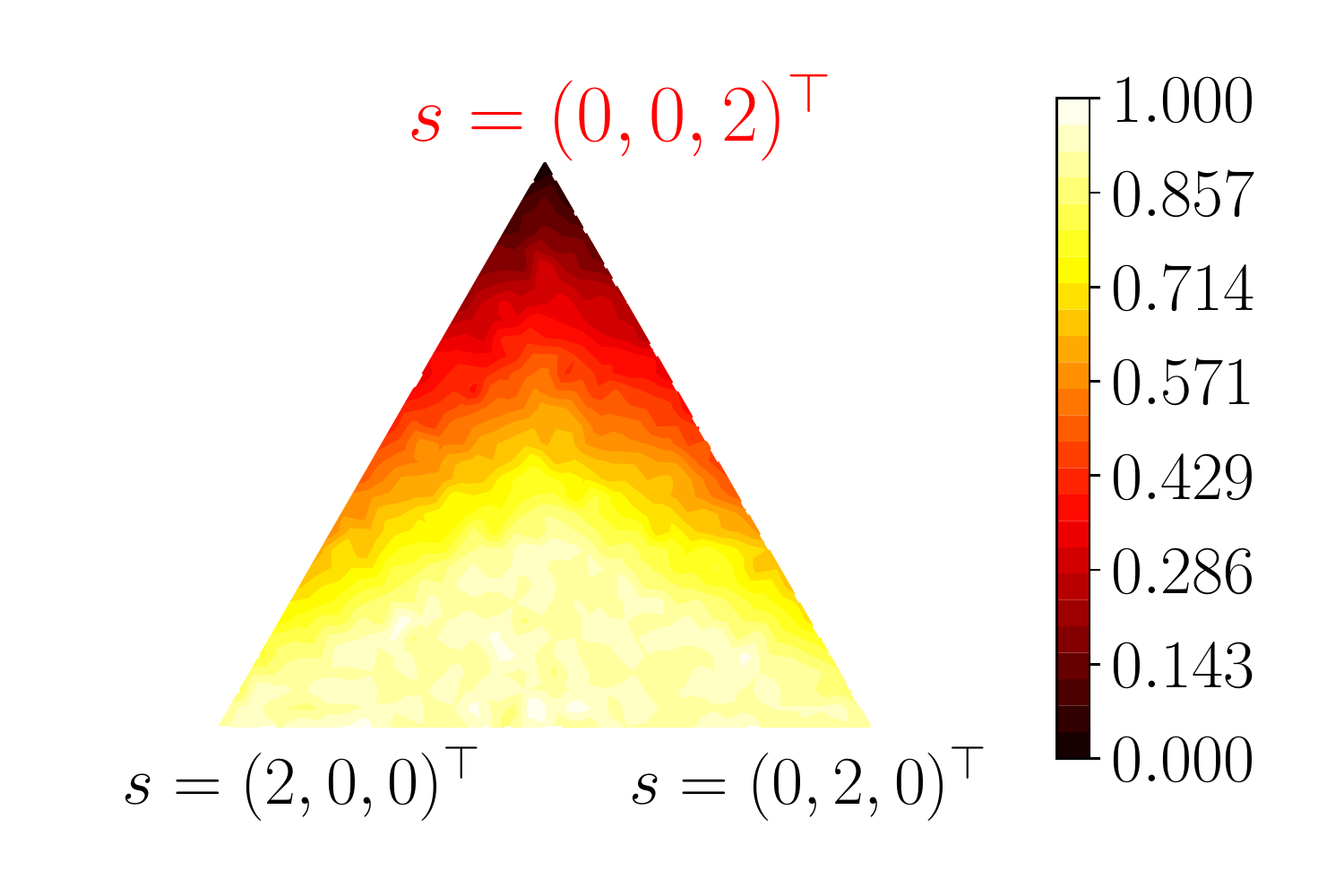}
        \caption{$B=50$}
        \label{subfig:k150}
    \end{subfigure}
    \hspace{0.1cm}
    \begin{subfigure}[b]{0.187\textwidth}
        \centering
        \includegraphics[width=\textwidth]{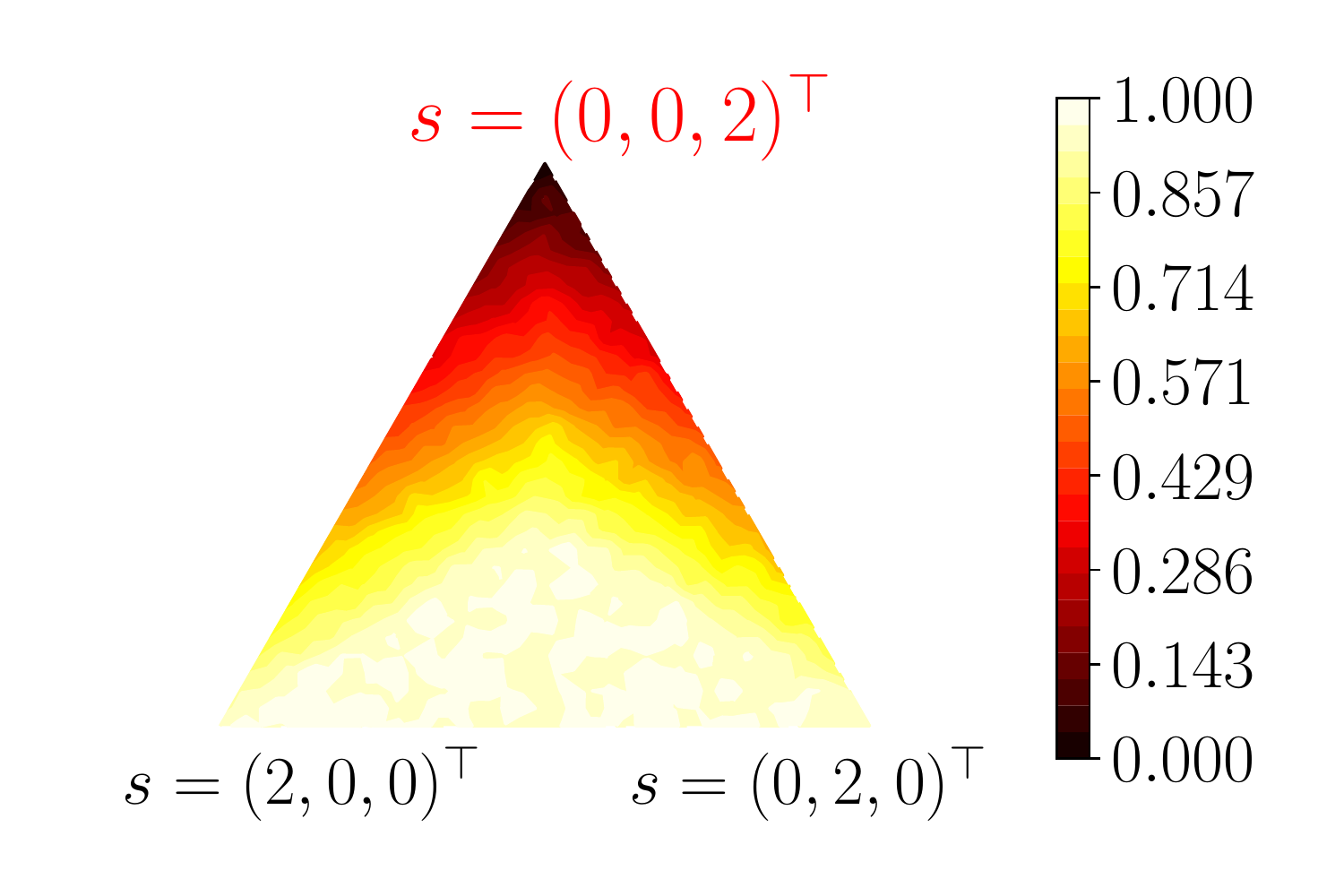}
        \caption{$B=100$}
        \label{subfig:k1B100}
    \end{subfigure}
    \caption{Impact of the sampling parameter $B$ on the loss $\ell_{\text{Noised bal.}}^{K, 1, B}$ (top part: $K=1$, bottom part: $K=2$)}
    \label{fig:impact_B}
\end{figure*}

\subsection{Proof of \Cref{prop:topkapprox}}
\label{app:prop_topkapprox}

\begin{proof}
    \begin{itemize}
        \item From the triangle inequality and \cref{prop:smoothness} we get for any $\bfs, \bfs' \in \bbR^{L}$:
              \begin{align*}
                  \norm{\nabla \topp_{K, \epsilon}(\bfs) -  \nabla \topp_{K, \epsilon}(\bfs')}
                   & =  \lVert \left[\nabla \topsum_{K, \epsilon}(\bfs) - \nabla \topsum_{K, \epsilon}(\bfs')\right] - \left[\nabla \topsum_{K-1, \epsilon}(\bfs) - \nabla \topsum_{K-1, \epsilon}(\bfs')\right] \rVert \\
                   & \leq \lVert \nabla \topsum_{K, \epsilon}(\bfs) - \nabla \topsum_{K, \epsilon}(\bfs') \rVert
                  +
                  \lVert \nabla \topsum_{K-1, \epsilon}(\bfs) - \nabla \topsum_{K-1, \epsilon}(\bfs') \rVert                                                                                                            \\
                   & \leq \left(2\frac{\sqrt{KL}}{\epsilon} + 2\frac{\sqrt{(K-1)L}}{\epsilon}\right)\norm{\bfs-\bfs'} \leq 4\frac{\sqrt{KL}}{\epsilon}\norm{\bfs-\bfs'}
              \end{align*}
        \item Using the notation from \citep[Appendix A]{berthet2020learning}, with $F(\bfs) = \topsum_{K}(\bfs)$ and
              $F_{\epsilon}(\bfs) = \topsum_{K, \epsilon}(\bfs)$, we get the following bounds:
              \begin{align}
                   & 0 \leq \topsum_{K, \epsilon}(\bfs) - \topsum_{K}(\bfs) \leq\epsilon\cdot \topsum_{K, 1}(\mathbf{0})                  \label{ineq:1} \\
                   & 0 \leq \topsum_{K-1, \epsilon}(\bfs) - \topsum_{K-1}(\bfs) \leq \epsilon\cdot \topsum_{K-1, 1}(\mathbf{0}) \enspace. \label{ineq:2}
              \end{align}
              Additionally, using the maximal inequality for \iid Gaussian variables, see for instance \citep[Section 2.5]{Boucheron_Lugosi_Massart13}, leads to:
              \begin{align*}
                  \topsum_{K, 1}(\mathbf{0})=\mathbb{E}\left[\sum_{k\in[K]} Z_{(k)}\right]\leq K\mathbb{E}\big[Z_{(1)}\big]\leq K\sqrt{2\log L}
              \end{align*}
              Subtracting \eqref{ineq:2} to \eqref{ineq:1}, and reminding that $\topp_{K}(\bfs)=\topsum_{K}(\bfs)-\topsum_{K-1}(\bfs)$ (and similarly $\topp_{K,\epsilon}(\bfs)=\topsum_{K,\epsilon}(\bfs)-\topsum_{K-1,\epsilon}(\bfs)$) gives:
              \begin{align*}
                  -\epsilon\cdot (K-1) \sqrt{2\log L}\leq \topp_{K, \epsilon}(\bfs) - \topp_{K}(\bfs) & \leq \epsilon\cdot K \sqrt{2\log L} \enspace,
              \end{align*}
              thus leading to:
              \begin{align}
                  \abs{\topp_{K, \epsilon}(\bfs) - \topp_{K}(\bfs)} \leq \epsilon\cdot C_{K, L} \enspace,
              \end{align}
              with $C_{K, L} =K\sqrt{2\log L} $.
    \end{itemize}
\end{proof}

\subsection{Proof of \Cref{prop:balanced}}
\label{app:prop_balanced}
\begin{proof}
    \begin{itemize}
        \item First, note that  $\bfs \mapsto \ell_{\text{Noised bal.}}^{K, \epsilon}(\bfs, y)$ is continuous as a composition and sum of continuous functions.
              It is differentiable wherever $\psi : \bfs \mapsto 1 + \topp_{K +1 , \epsilon}(\bfs) - s_{y}$ is non-zero.
              From \Cref{def:top-k} and \Cref{prop:smoothness} we get $\nabla_{s} \psi(\bfs) = \bbE[\argtops_{K+1}(\bfs + \epsilon Z)] - \delta_{y}$.
              The formula of the gradient follows from the chain rule.
    \end{itemize}
\end{proof}

\section{Illustrations of the various losses encountered}
\label{sec:illustrations_of_the_various_loss_encountered}

\begin{figure*}[t]
    \centering
    \begin{subfigure}[b]{0.187\textwidth}
        \centering
        \includegraphics[width=\textwidth]{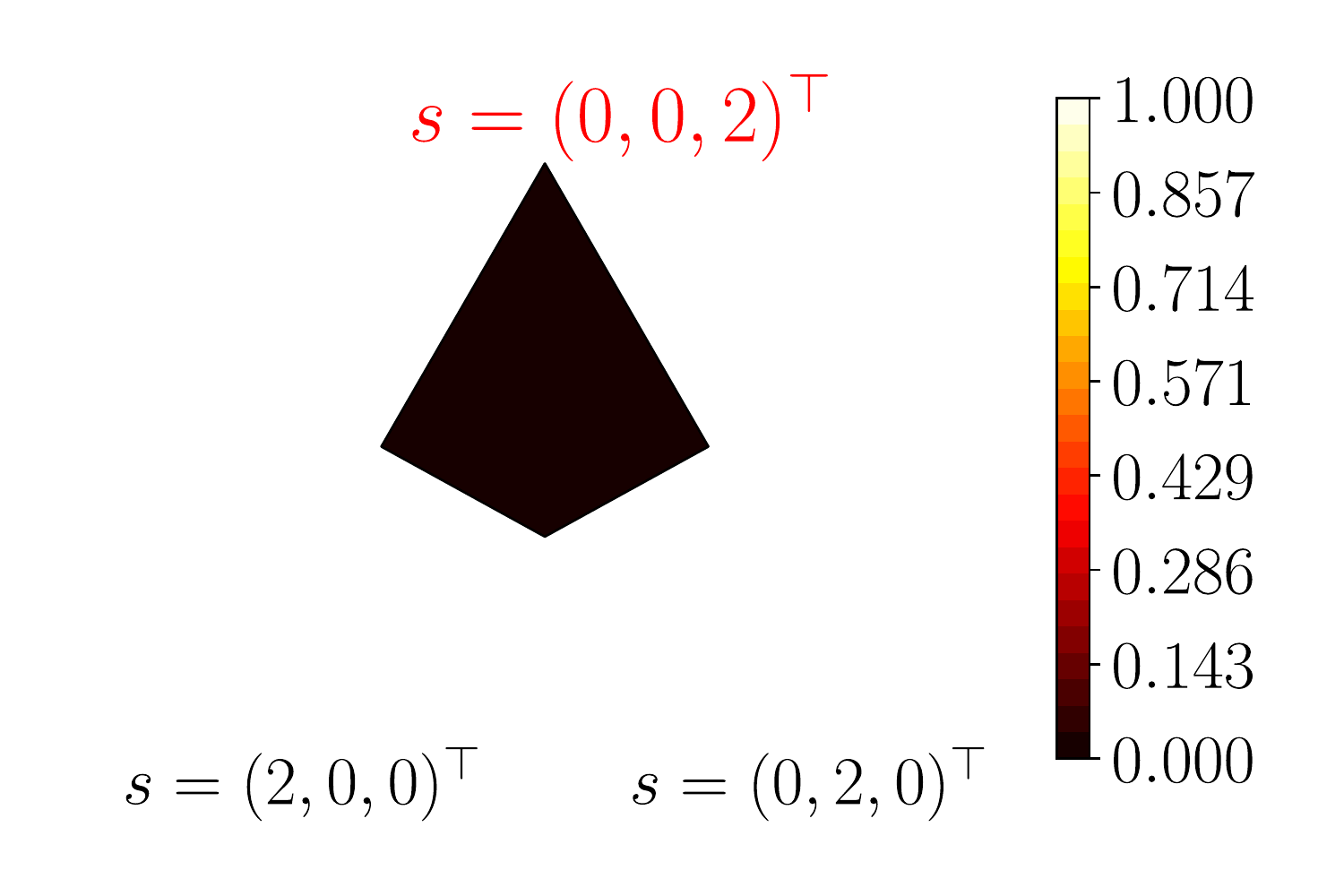}
        \caption{Top-$K$:\\ $\ell=\ell^{K}$.\phantom{$\ell_{\textrm{Cal.~Hinge}}^K$}}
        \label{subfig:topk_k=1}
    \end{subfigure}
    \hspace{0.1cm}
    \begin{subfigure}[b]{0.187\textwidth}
        \centering
        \includegraphics[width=\textwidth]{prebuiltimages/simplices/EntLoss_k=2_7_margin_1}
        \caption{Cross-entropy:\\ $\ell=\ell_{\textrm{CE}}$.\phantom{$\ell=\ell_{\textrm{Cal.~Hinge}}^K$}}
        \label{subfig:topk_k=1}
    \end{subfigure}
    \hspace{0.1cm}
    \begin{subfigure}[b]{0.187\textwidth}
        \centering
        \includegraphics[width=\textwidth]{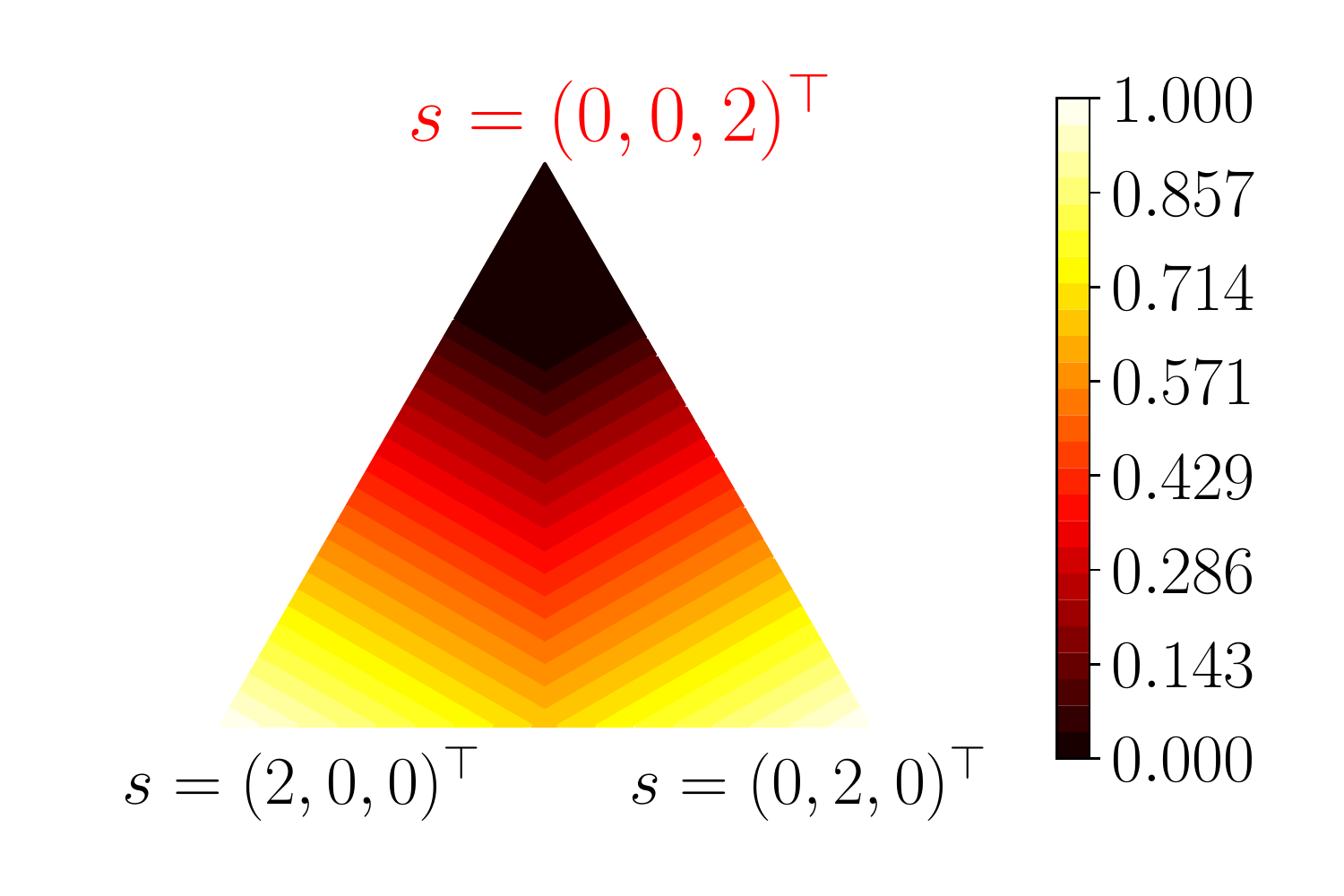}
        \caption{Multi-class hinge:\\ $\ell=\ell_{\textrm{Hinge}}^K$.}
    \end{subfigure}
    \hspace{0.1cm}
    \begin{subfigure}[b]{0.187\textwidth}
        \centering
        \includegraphics[width=\textwidth]{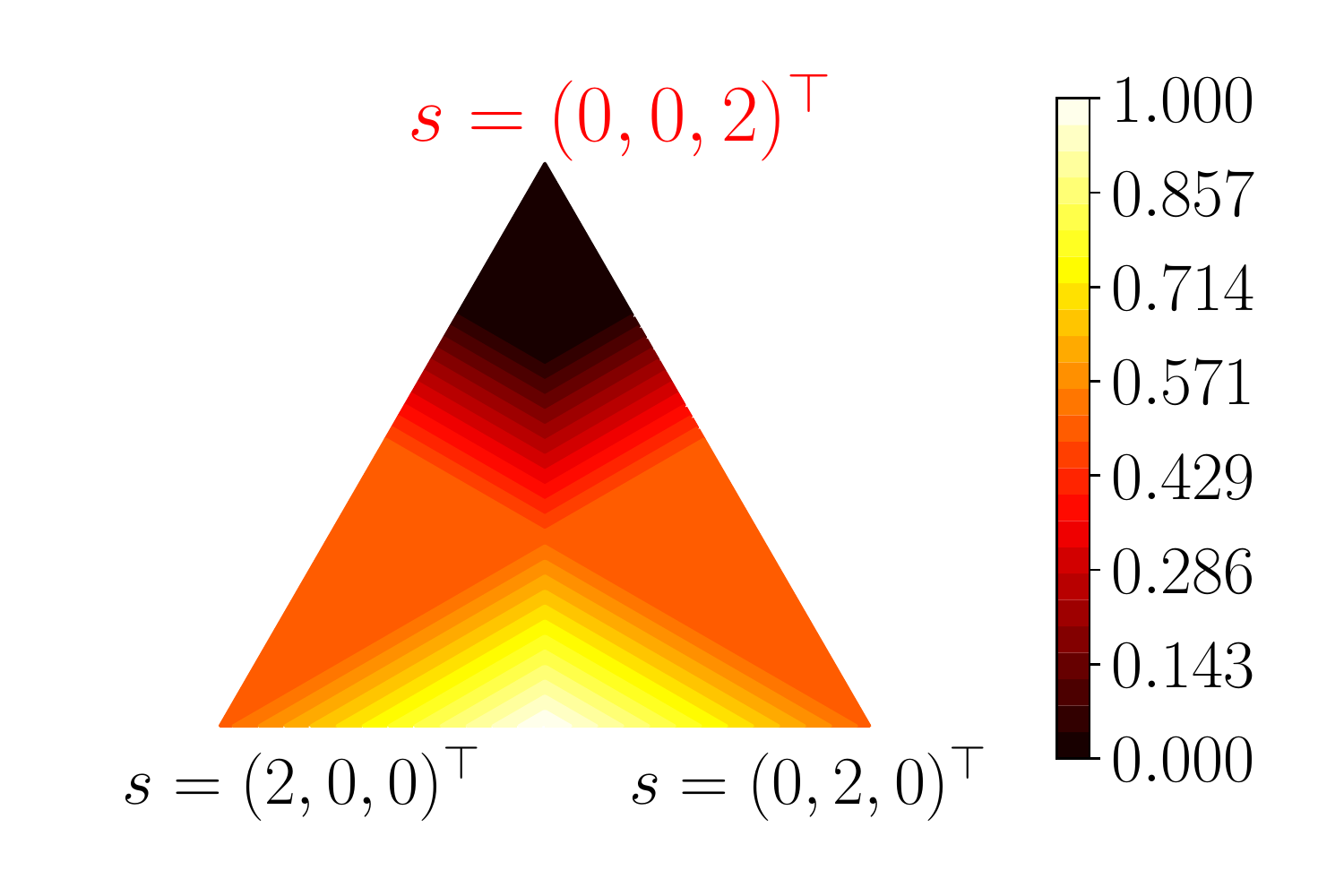}
        \caption{Calibrated hinge:\\$\ell=\ell_{\textrm{Cal.~Hinge}}^K$.}
    \end{subfigure}
    \hspace{0.1cm}
    \begin{subfigure}[b]{0.187\textwidth}
        \centering
        \includegraphics[width=\textwidth]{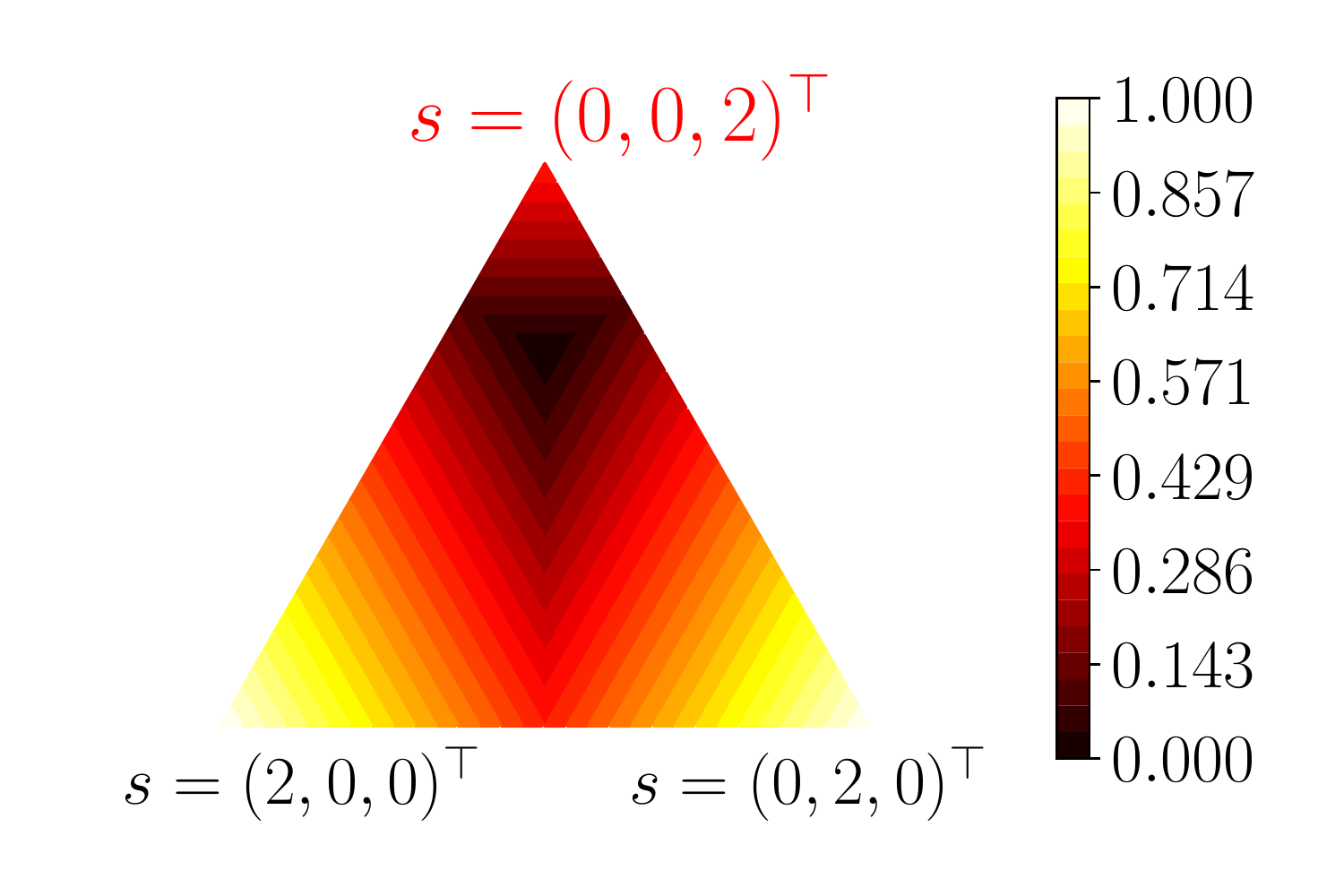}
        \caption{Convexified hinge:\\$\ell=\ell_{\textrm{CVXHinge}}^K$.}
        \label{subfig:convex_k=1_10}
    \end{subfigure}\\
    \begin{subfigure}[b]{0.187\textwidth}
        \centering
        \includegraphics[width=\textwidth]{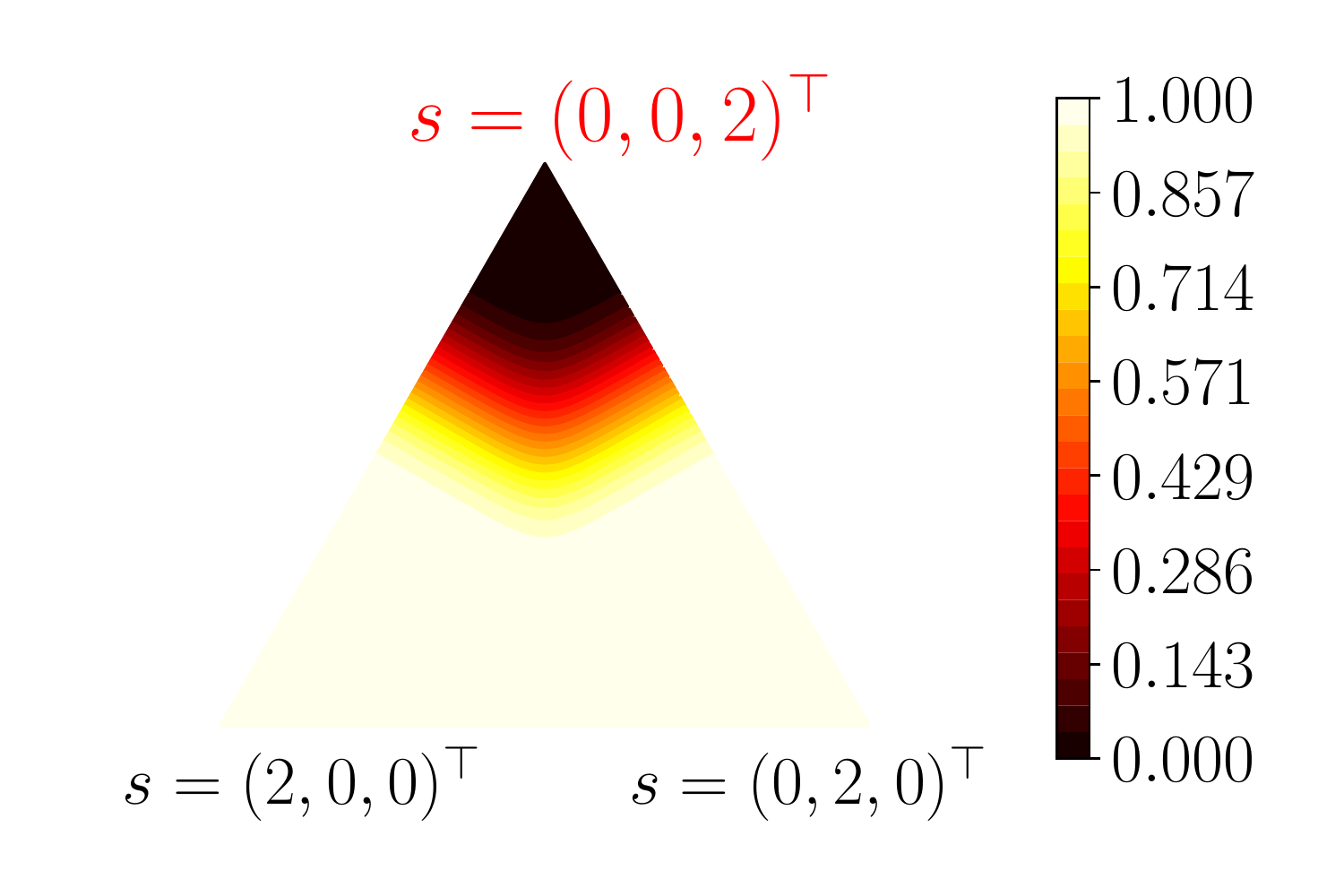}
        \caption{Smoothed hinge\\$\ell_{\text{Smoothed Hinge}}^{K, 0.1}$.}
    \end{subfigure}
    \hspace{0.1cm}
    \begin{subfigure}[b]{0.187\textwidth}
        \centering
        \includegraphics[width=\textwidth]{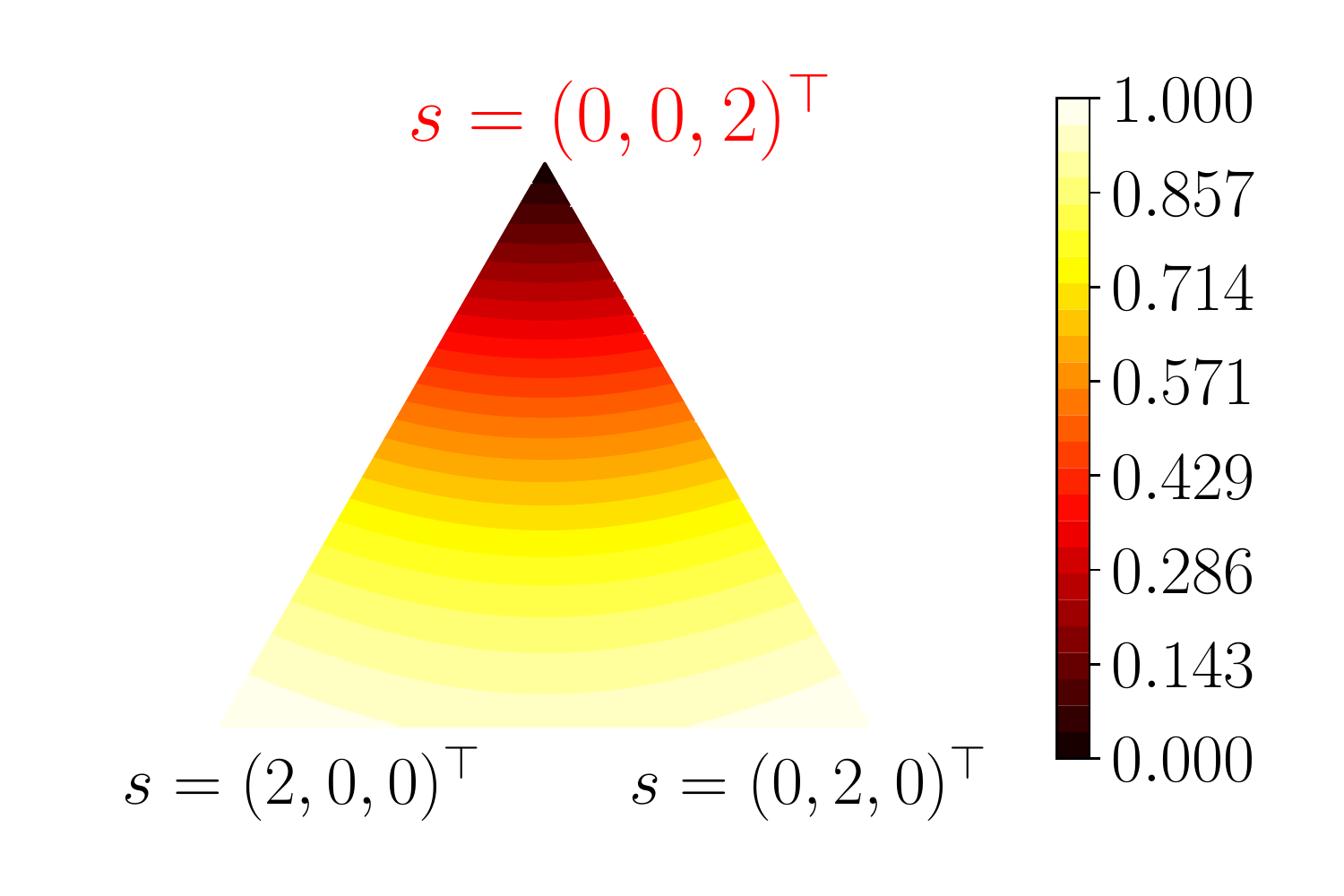}
        \caption{Smoothed hinge\\$\ell_{\text{Smoothed Hinge}}^{K, 1}$.}
    \end{subfigure}
    \hspace{0.1cm}
    \begin{subfigure}[b]{0.187\textwidth}
        \centering
        \includegraphics[width=\textwidth]{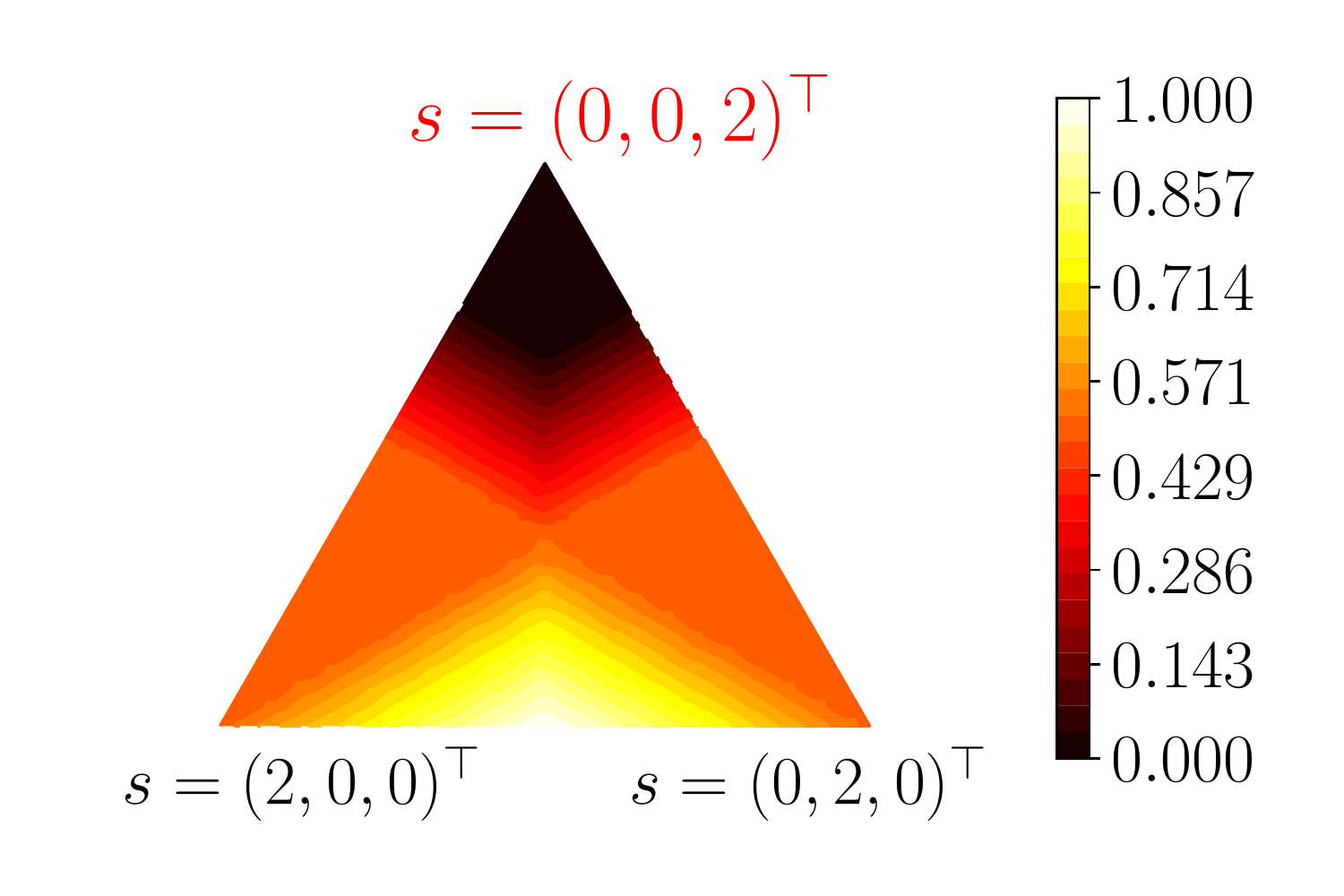}
        \caption{Noised balanced: \phantom{g}\\ $\ell_{\text{Noised bal.}}^{K, 0.3, 30}$.\phantom{$\ell_{\text{Smoothed Hinge}}^{K, 0.4}$}}
        \label{subfig:epsilon_K1_03}
    \end{subfigure}
    \hspace{0.1cm}
    \begin{subfigure}[b]{0.187\textwidth}
        \centering
        \includegraphics[width=\textwidth]{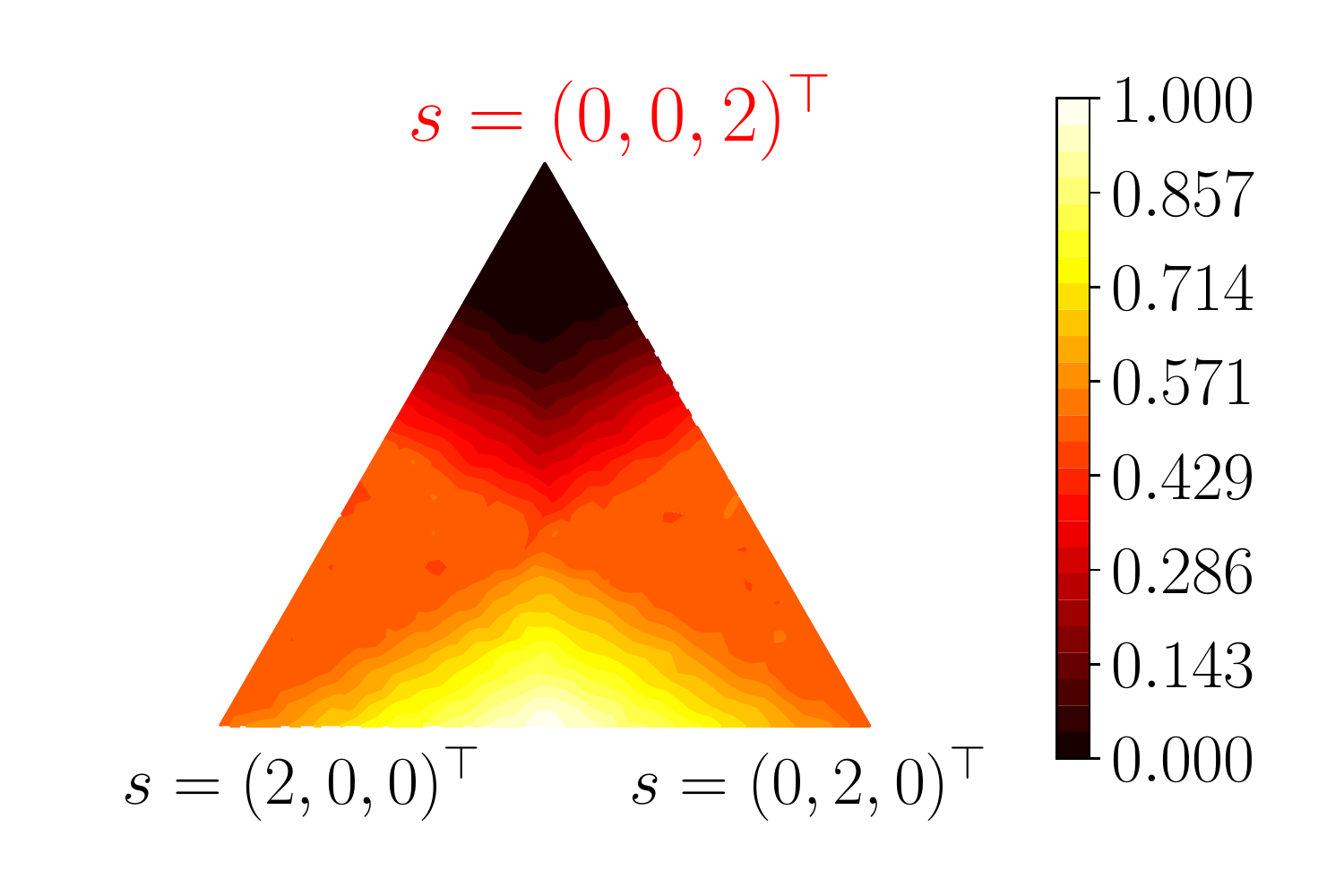}
        \caption{Noised balanced:\phantom{g}\\ $\ell_{\text{Noised bal.}}^{K, 1, 30}$.\phantom{$\ell_{\text{Smoothed Hinge}}^{K, 0.4}$}}
        \label{subfig:epsilon_K1_1}
    \end{subfigure}
    \hspace{0.1cm}
    \begin{subfigure}[b]{0.187\textwidth}
        \centering
        \includegraphics[width=\textwidth]{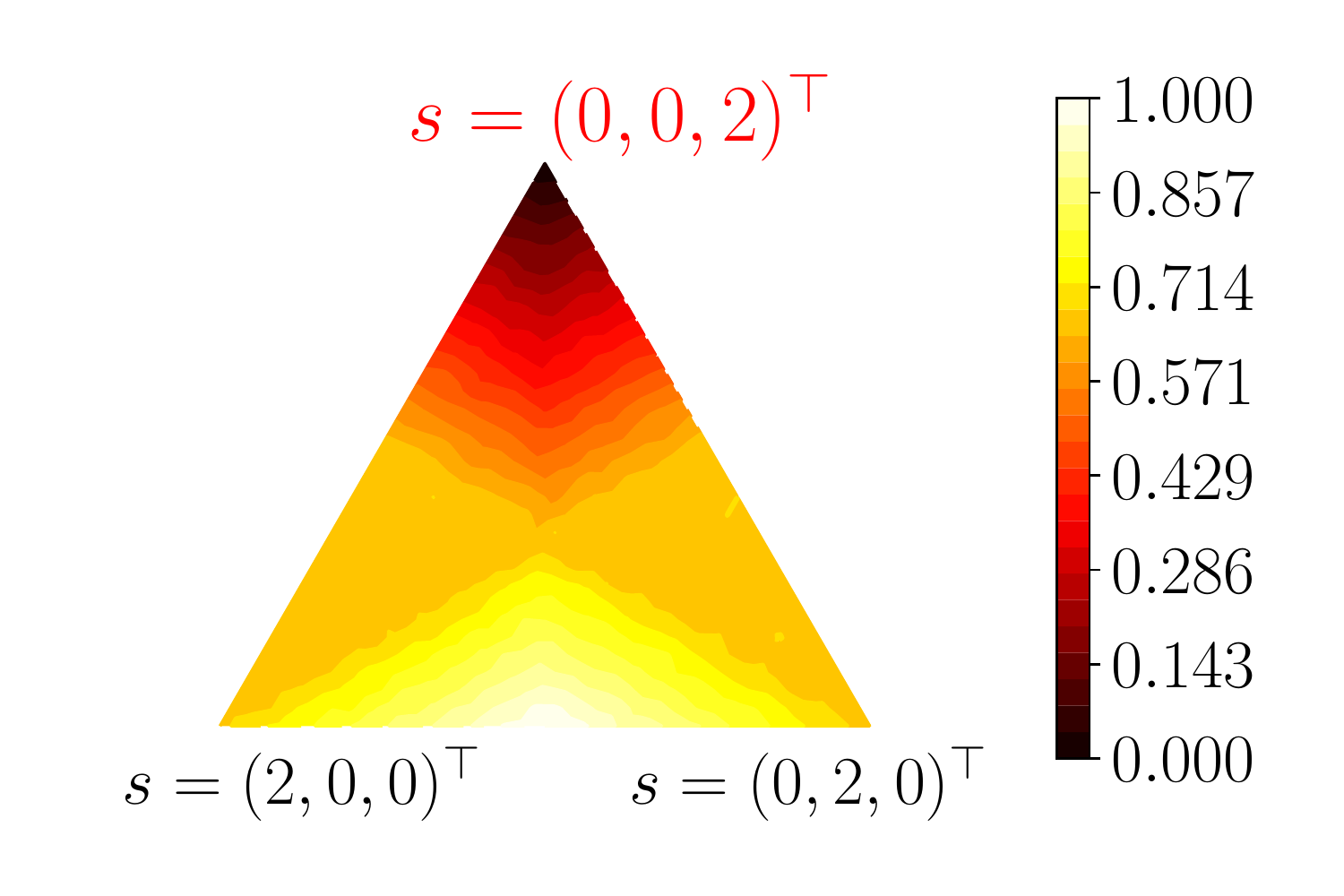}
        \caption{Noised imbalanced:\phantom{g}\\ $\ell_{\text{Noised imbal.}}^{K, 1, 30,5}$.\phantom{$\ell_{\text{Smoothed Hinge}}^{K, 0.4}$}}
        \label{subfig:epsilon_K1_1}
    \end{subfigure}
    \caption{Level sets of the function $\bfs\mapsto \ell(\bfs,y)$ for different losses described in \Cref{tab:losses}, for $L=3$ classes, $K=1$ and a true label $y=2$ (corresponding to the upper corner of the triangles).
        For visualization the loss are rescaled between 0 and 1, and the level sets are restricted to vector $\bfs \in 2 \cdot \Delta_3$.
        The losses have been harmonized to display a margin equal to 1.
        For our proposed loss, we have averaged the level sets over 100 replications to avoid meshing artifacts.}
    \label{fig:simplices_k=1}
\end{figure*}

In \Cref{fig:simplices_k=1,fig:simplices_k=2}, we provide a visualization of the loss landscapes for respectively $K=1$ and $K=2$ with $L=3$ labels.
With $L=3$ labels, we display the visualization as level-sets restricted to a rescaled simplex: $2\dots\Delta_3$.
Moreover, we have min/max rescaled all the losses so that they fully range the interval $[0, 1]$.
Note that as we are mainly interested in minimizing the losses, this post-processing would not modify the learned classifiers.

We provide also two additional figures illustrating the impact on our loss of the two main parameters: $\epsilon$ and $B$.

\captionsetup[subfigure]{justification=justified,singlelinecheck=false}
\begin{figure*}[t]
    \centering
    \begin{subfigure}[b]{0.187\textwidth}
        \centering
        \includegraphics[width=\textwidth]{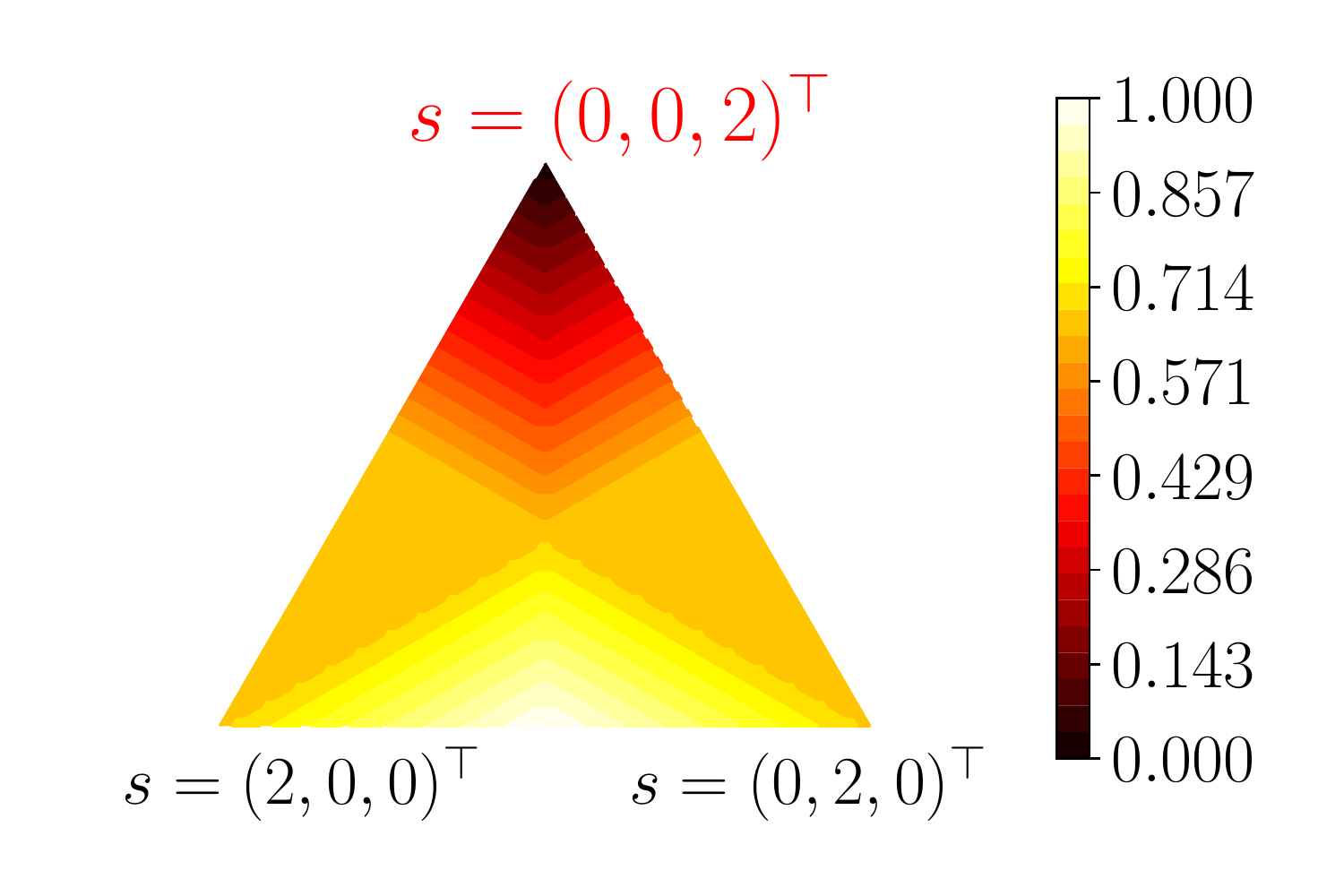}
        \caption{$\epsilon=0.01$}
        \label{subfig:k1B1}
    \end{subfigure}
    \hspace{0.1cm}
    \begin{subfigure}[b]{0.187\textwidth}
        \centering
        \includegraphics[width=\textwidth]{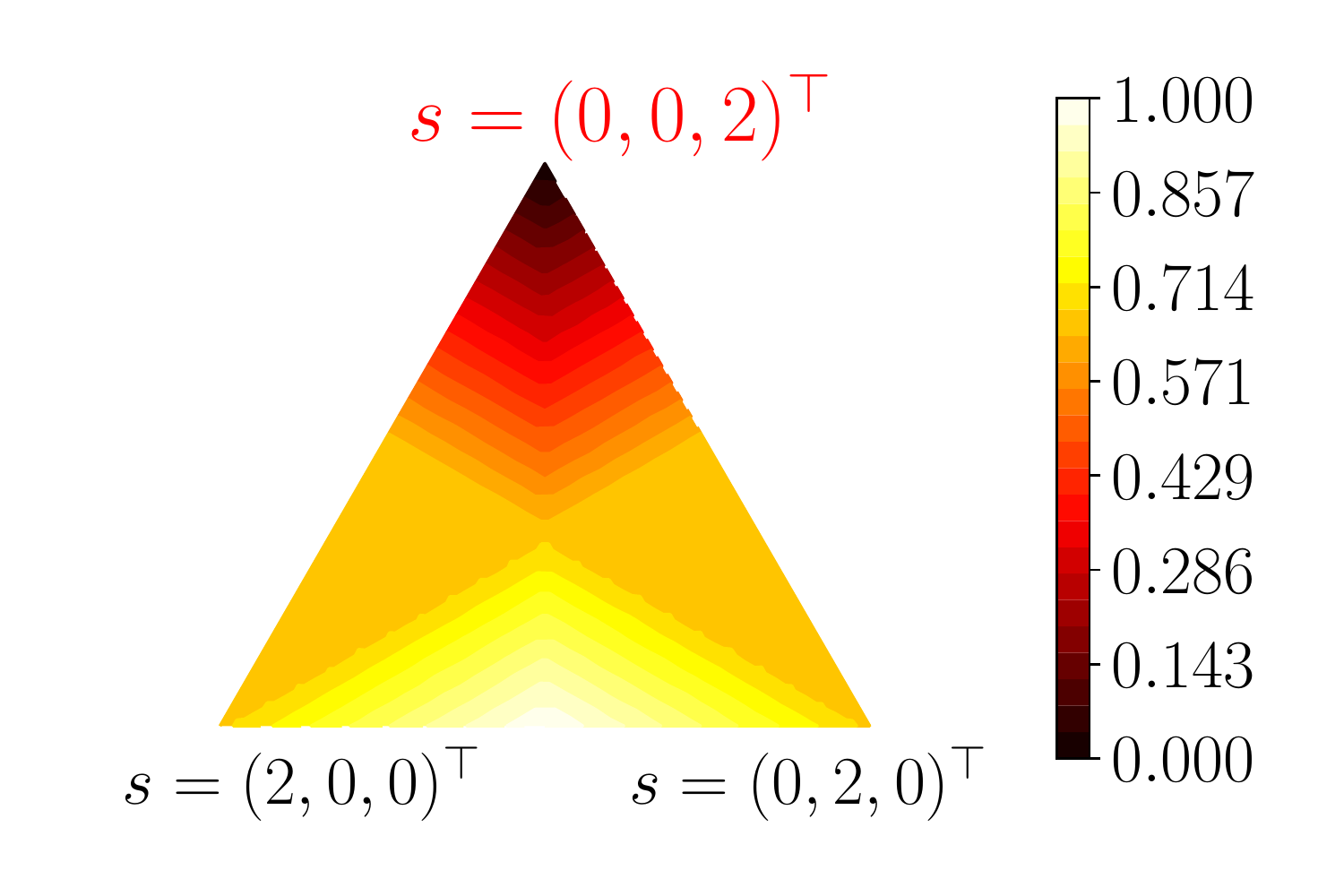}
        \caption{$\epsilon=0.1$}
        \label{subfig:k1B5}
    \end{subfigure}
    \hspace{0.1cm}
    \begin{subfigure}[b]{0.187\textwidth}
        \centering
        \includegraphics[width=\textwidth]{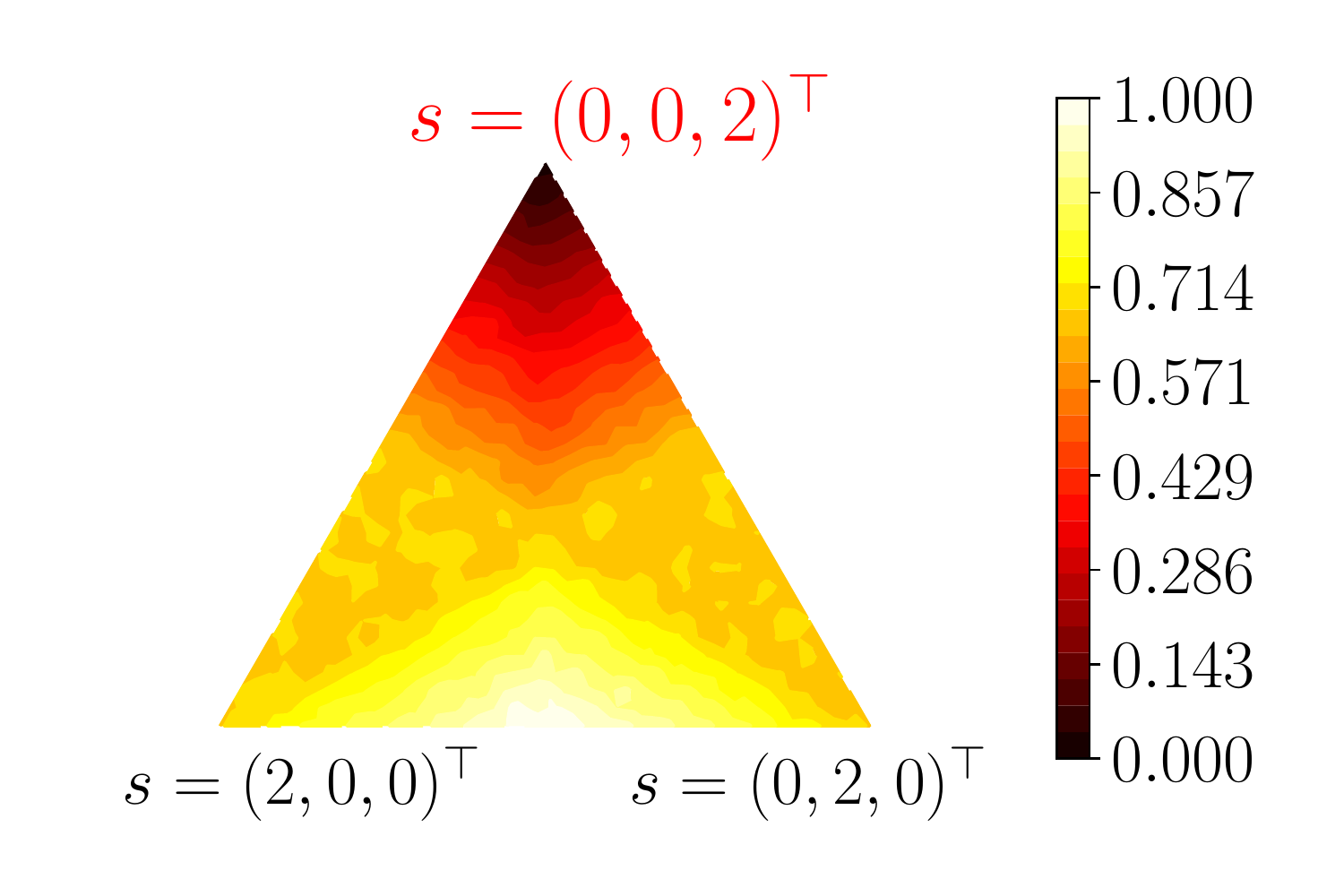}
        \caption{$\epsilon=1$}
        \label{subfig:k110}
    \end{subfigure}
    \hspace{0.1cm}
    \begin{subfigure}[b]{0.187\textwidth}
        \centering
        \includegraphics[width=\textwidth]{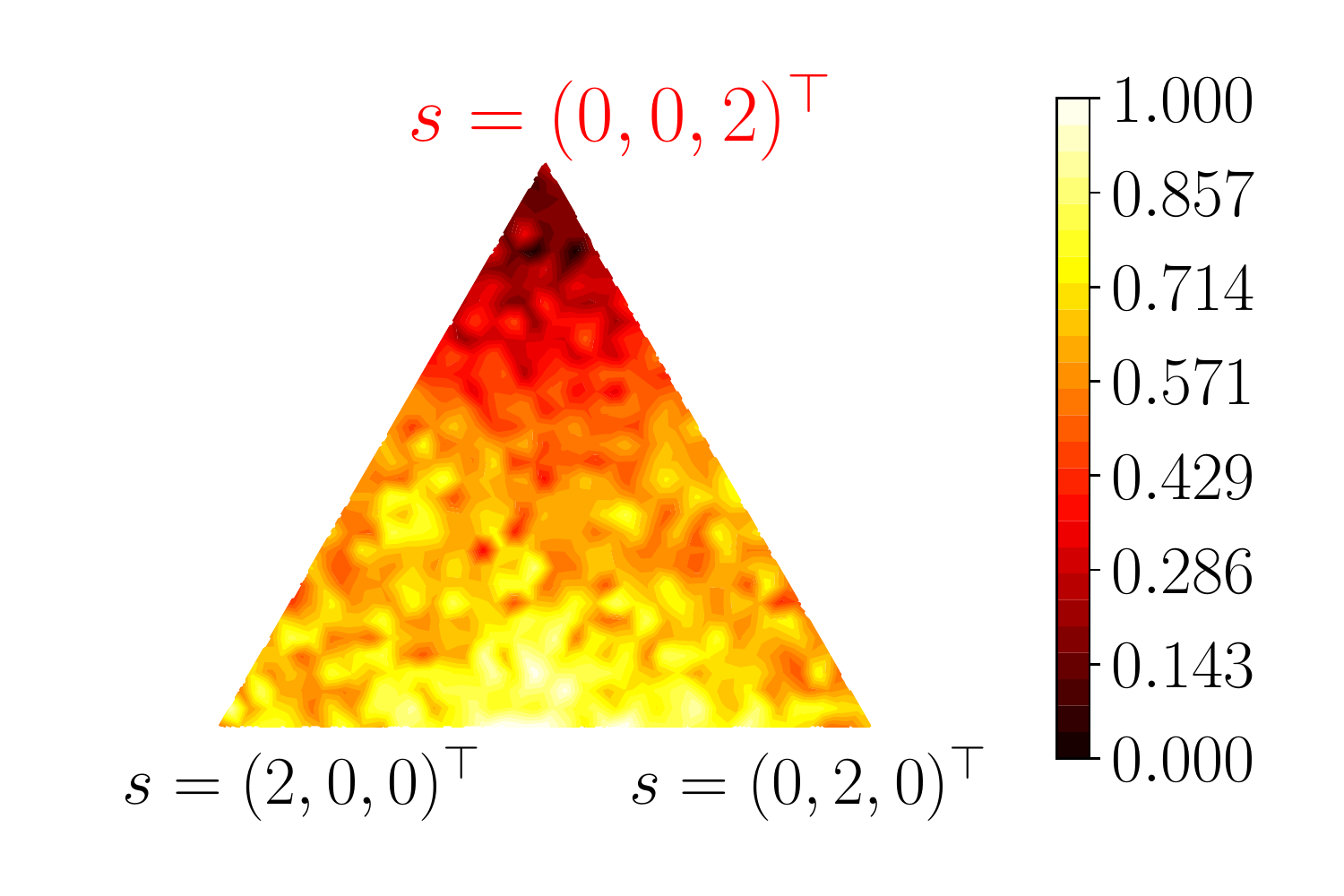}
        \caption{$\epsilon=10$}
        \label{subfig:k150}
    \end{subfigure}
    \hspace{0.1cm}
    \begin{subfigure}[b]{0.187\textwidth}
        \centering
        \includegraphics[width=\textwidth]{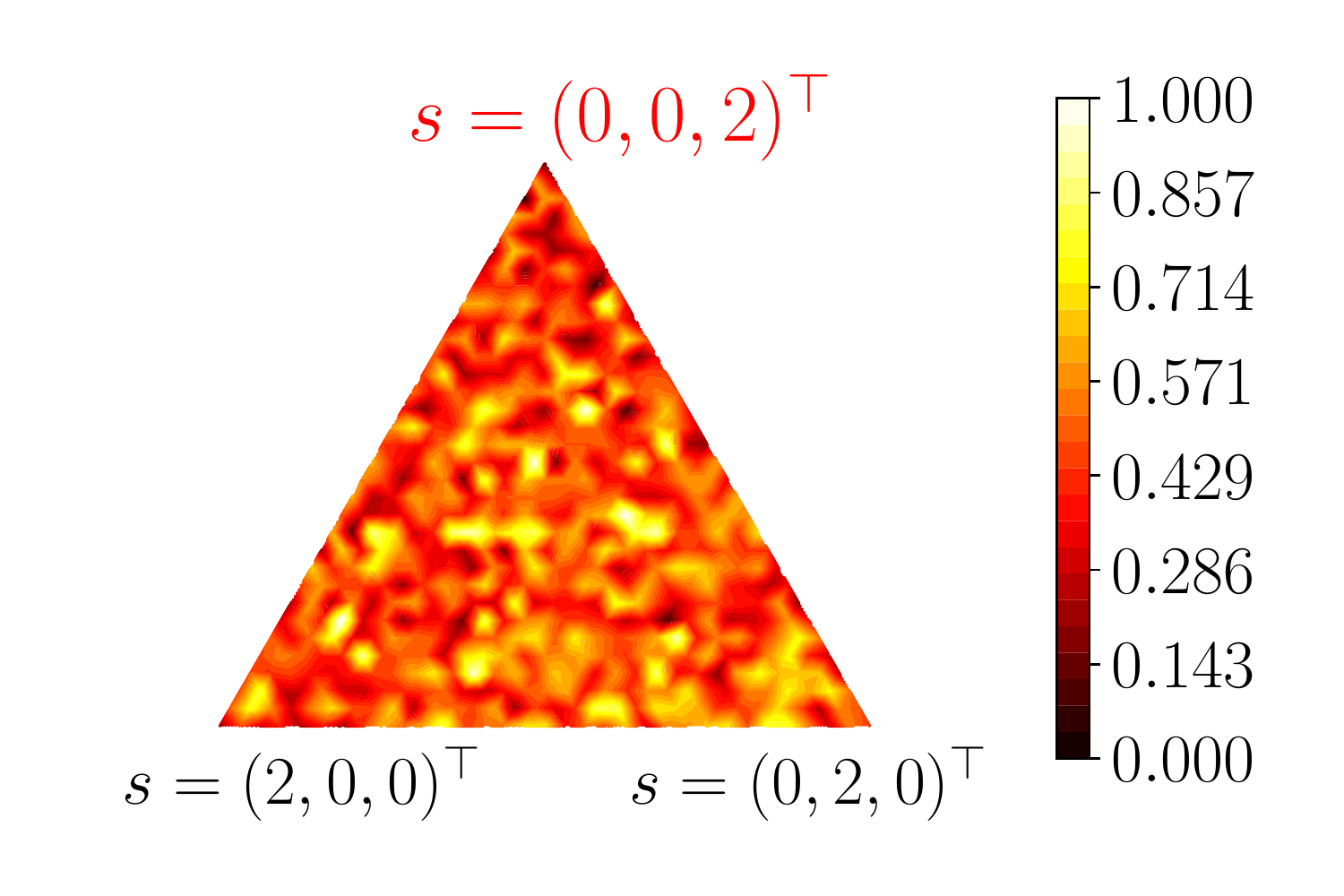}
        \caption{$\epsilon=100$}
        \label{subfig:k1B100}
    \end{subfigure}\\
    \begin{subfigure}[b]{0.187\textwidth}
        \centering
        \includegraphics[width=\textwidth]{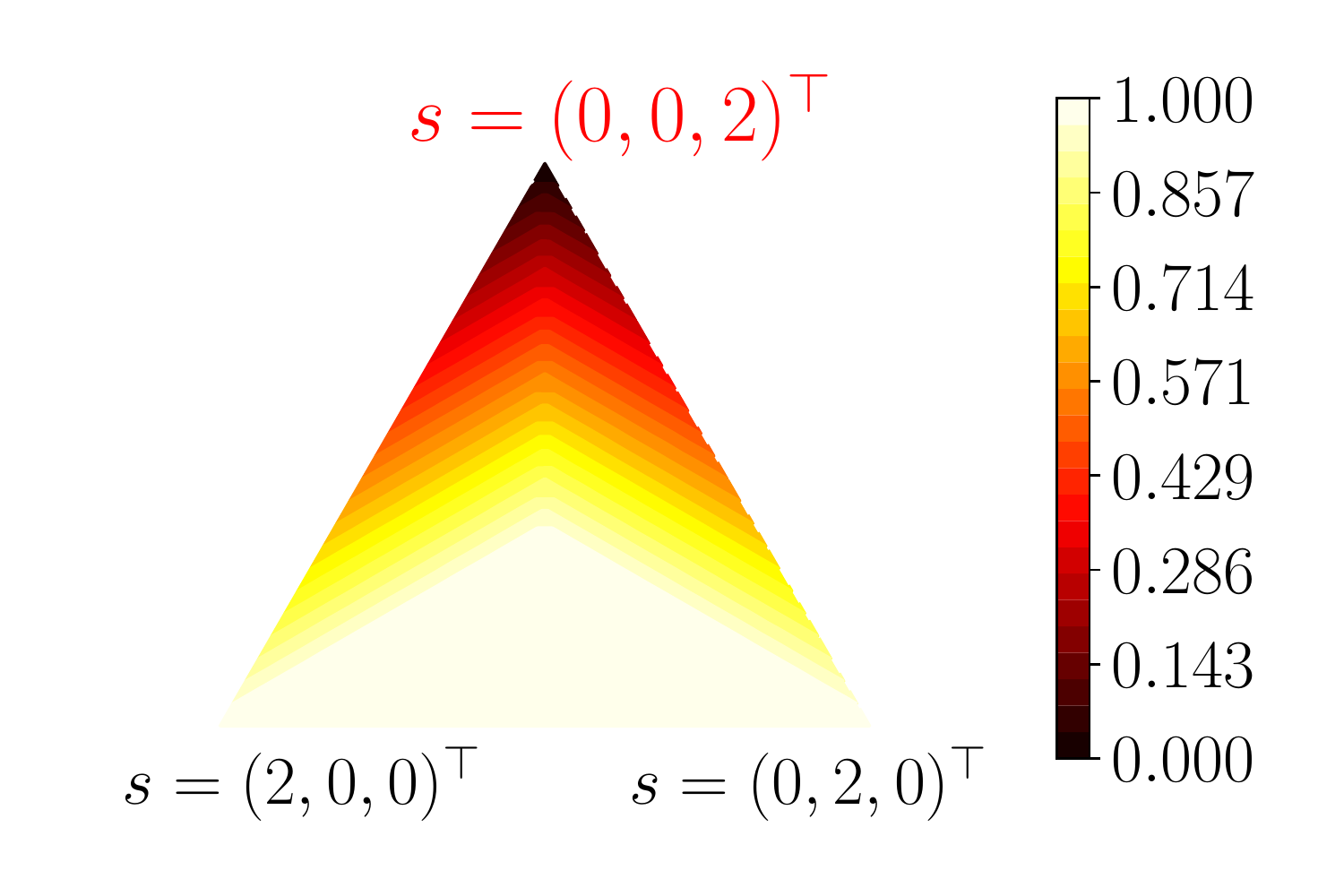}
        \caption{$\epsilon=0.01$}
        \label{subfig:k1B1}
    \end{subfigure}
    \hspace{0.1cm}
    \begin{subfigure}[b]{0.187\textwidth}
        \centering
        \includegraphics[width=\textwidth]{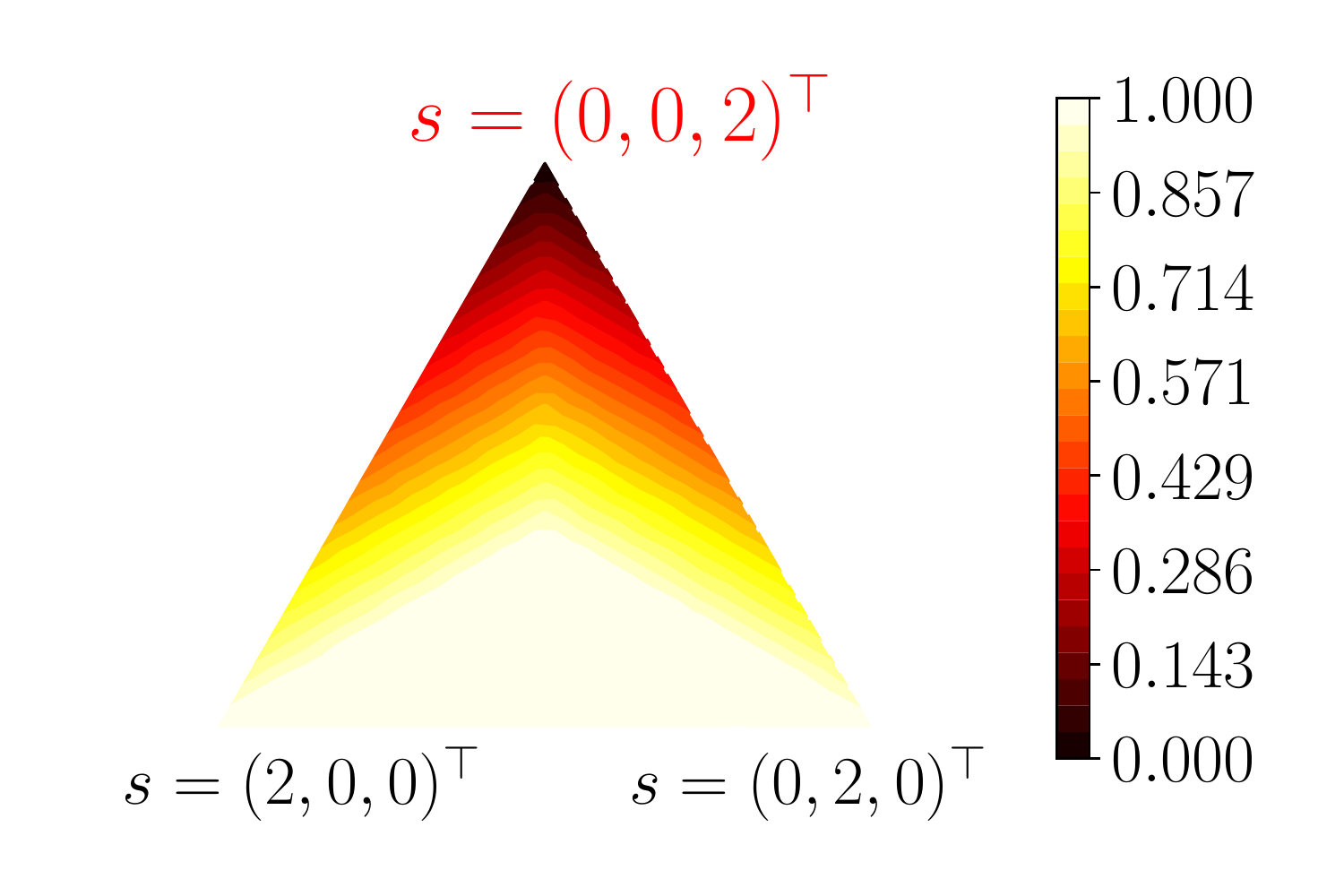}
        \caption{$\epsilon=0.1$}
        \label{subfig:k1B5}
    \end{subfigure}
    \hspace{0.1cm}
    \begin{subfigure}[b]{0.187\textwidth}
        \centering
        \includegraphics[width=\textwidth]{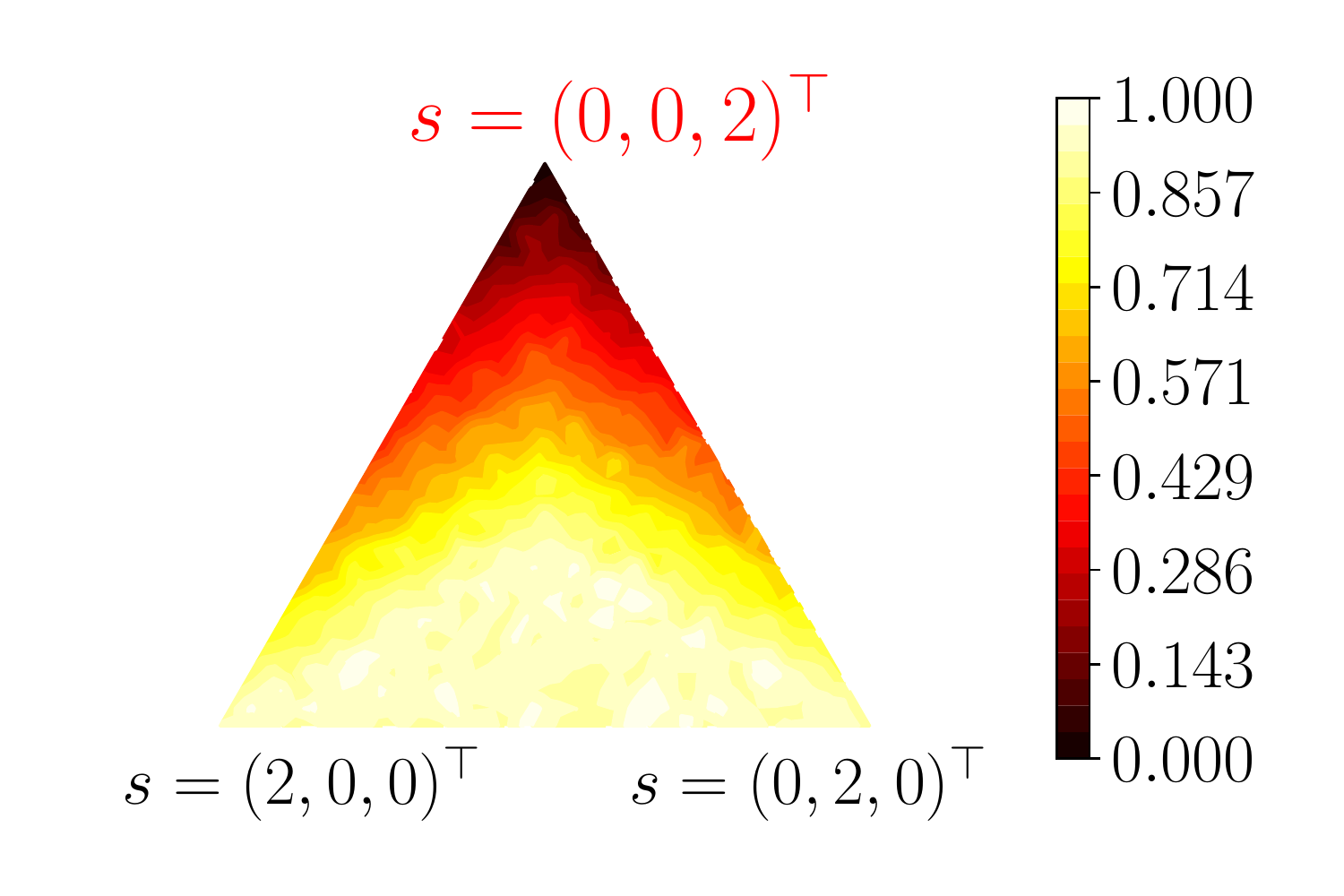}
        \caption{$\epsilon=1$}
        \label{subfig:k110}
    \end{subfigure}
    \hspace{0.1cm}
    \begin{subfigure}[b]{0.187\textwidth}
        \centering
        \includegraphics[width=\textwidth]{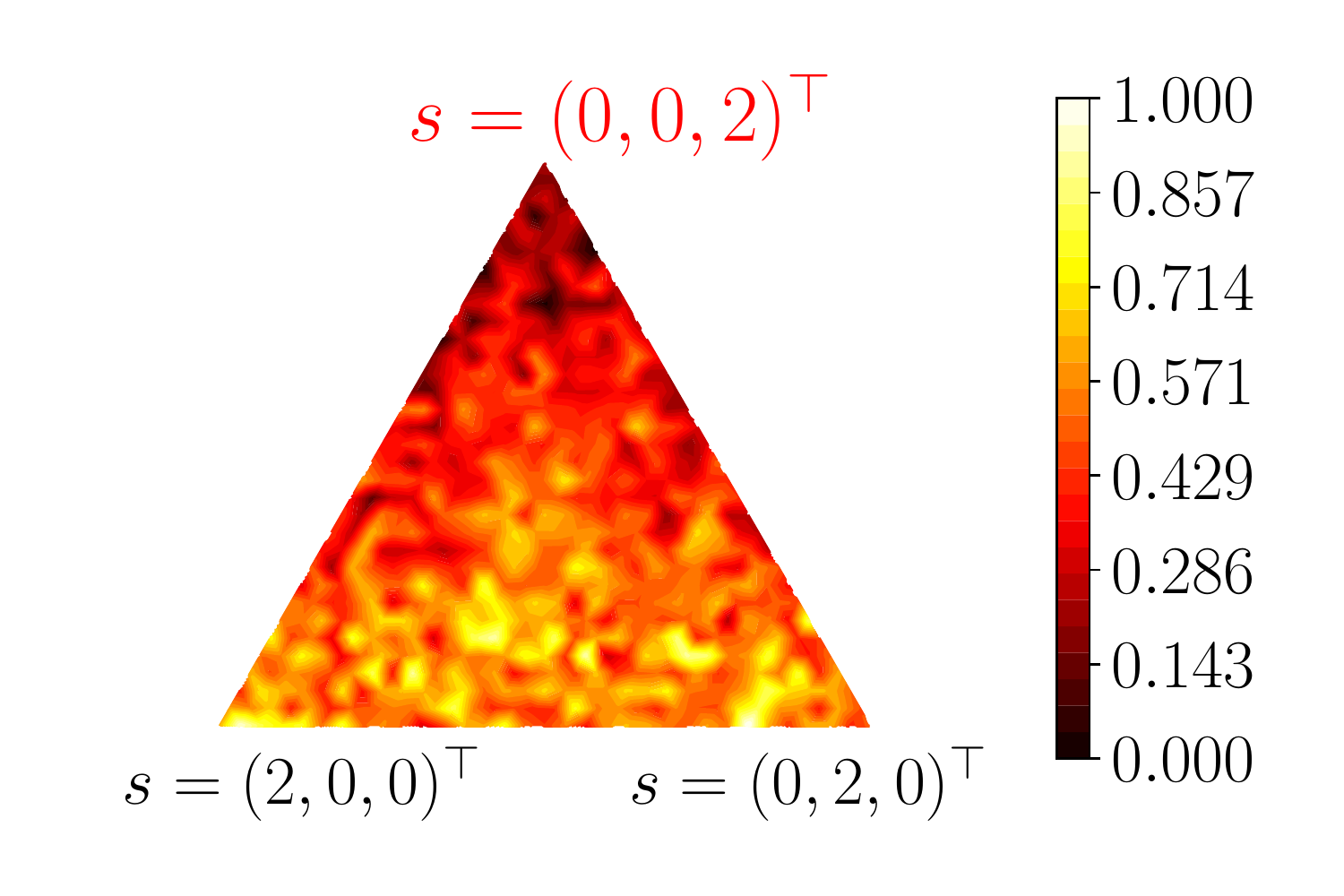}
        \caption{$\epsilon=10$}
        \label{subfig:k150}
    \end{subfigure}
    \hspace{0.1cm}
    \begin{subfigure}[b]{0.187\textwidth}
        \centering
        \includegraphics[width=\textwidth]{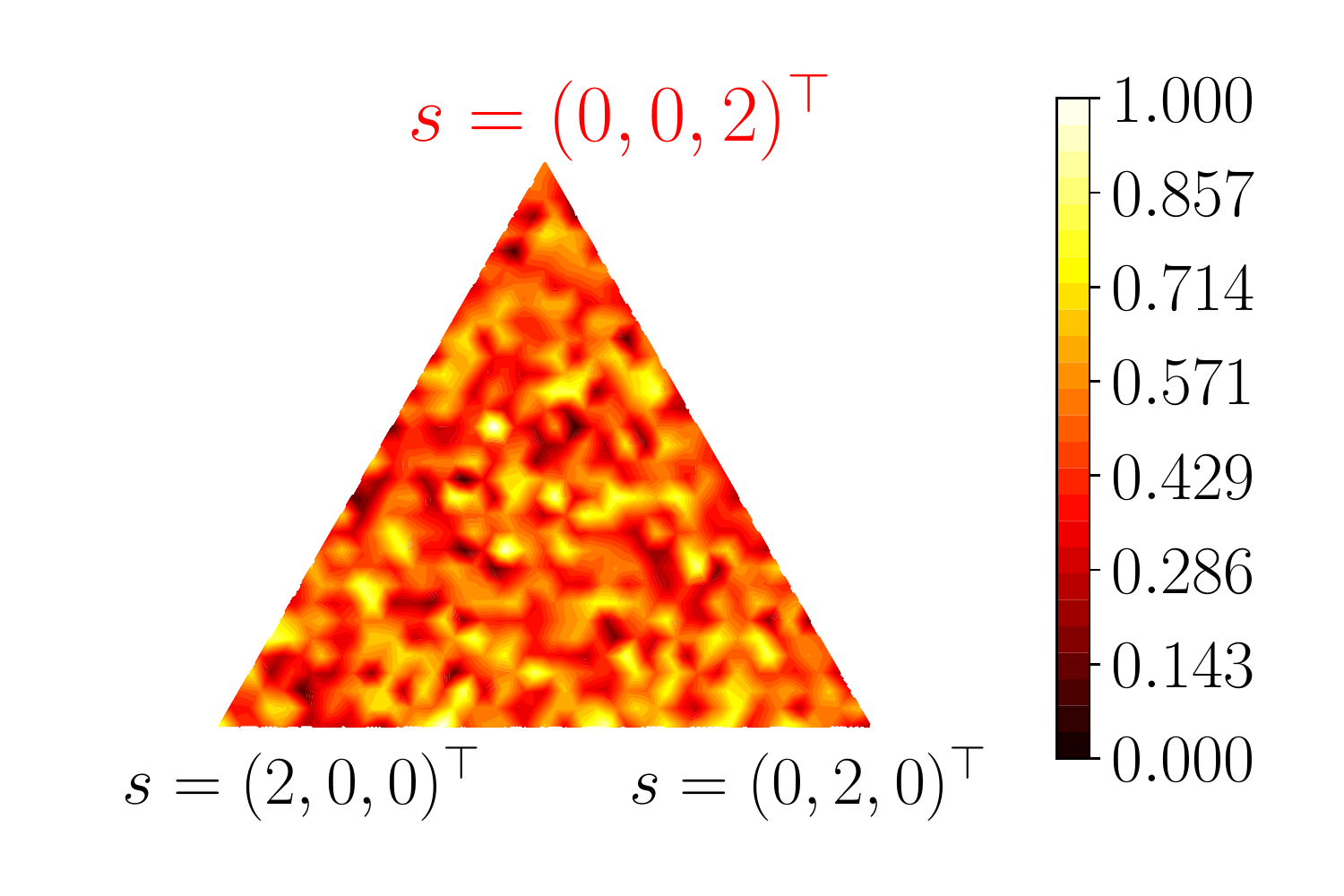}
        \caption{$\epsilon=100$}
        \label{subfig:k1B100}
    \end{subfigure}
    \caption{Impact of the smoothing parameter $\epsilon$ on the loss $\ell_{\text{Noised bal.}}^{K, \epsilon, 50}$ (top part: $K=1$, bottom part: $K=2$)}
    \label{fig:impact_epsilon}
\end{figure*}

\captionsetup[subfigure]{justification=justified,singlelinecheck=false}

\section{Additional experiments}
\label{subsec:Additional experiments}

\begin{table}
    \caption{top-1 accuracy cifar100}
        \centering
        \begin{normalsize}
        \begin{tabular}{lcccr}
            \toprule
            Label noise & $\ell^K_{\textrm{CE}}$ &      $\ell_{\text{Smoothed Hinge}}^{5, 1.0} $ & $\ell_{\text{Noised bal.}}^{5, 0.2, 10}$ \\
            \midrule
            0.0         & $\mathbf{76.6} {\pm} 0.1$              & $69.2 {\pm} 0.1$                           & $68.7 {\pm} 0.3$                                    \\
            0.1         & $71.0 {\pm} 0.2$              & $\mathbf{71.2} {\pm} 0.4$                           & $68.3 {\pm} 0.5$                                    \\
            0.2         & $68.1 {\pm} 0.1$              & $\mathbf{71.3} {\pm} 0.3$                           & $69.4 {\pm} 0.5$                                 \\
            0.3         & $65.5 {\pm} 0.3$              & $\mathbf{70.8} {\pm} 0.6$                           & $69.3 {\pm} 0.3$                                    \\
            0.4         & $61.8 {\pm} 0.4$              & $\mathbf{70.6} {\pm} 0.2$                           & $69.1 {\pm} 0.4$                                    \\
            \bottomrule
        \end{tabular}
        \label{table:cifar100-acc}
    \end{normalsize}
\end{table}

\textbf{CIFAR-100}: \Cref{table:cifar100-acc} reports the top-1 accuracy obtained by the models corresponding to the results of \Cref{table:cifar100-top5-acc}.
Hence, we show here a misspecified case: we optimized our balanced loss for $K=5$, seeking to optimize top-5 accuracy, which is reported in \Cref{table:cifar100-top5-acc}, but report top-1 information.
\Cref{table:cifar100-acc} shows that when there is no label noise, as expected cross entropy gives better top-1 accuracy than our top-5 loss.
When the label noise is high, however, our loss leads to better top-1 accuracy.

\textbf{Pl@ntNet-300K}: \Cref{table:plantnet-300k_appendix} reports the top-$K$ accuracy obtained by the models corresponding to the results of \Cref{table:plantnet-300k}.
The top-$K$ accuracies are much higher than the macro-average top-$K$ accuracies reported in \Cref{table:plantnet-300k} because of the long-tailed distribution of Pl@ntNet-300K.
The models perform well on classes with a lot of examples which leads to high top-$K$ accuracy. However, they struggle on classes with a small number of examples (which is the majority of classes, see \cite{plantnet-300k}).
Thus, for Pl@ntNet-300K top-$K$ accuracy is not very relevant as it mainly reflects the performance of the few classes with a lot of images, ignoring the performance on challenging classes (the one with few labels) \cite{plantnet}.
We report it for completeness and make a few comments:
\begin{table}
    \caption{regular top-$K$ accuracy corresponding to the models in \Cref{table:plantnet-300k}}
    \vskip 0.1in
    \centering
    \begin{normalsize}
        \begin{tabular}{lcccccc}
            \toprule
            K  & $\ell_{\textrm{CE}}$ &          $\ell_{\text{Smoothed Hinge}}^{K, \tau}$ &               $\ell_{\text{Noised bal.}}^{K, \epsilon, B}$           & focal                         & LDAM  &                    $\ell_{\text{Noised imbal.}}^{K, \epsilon, B,m_y}$ \\ \midrule
            1  & $80.1 {\pm} 0.1$                & $79.8 {\pm} 0.0$                                    & $80.8 {\pm} 0.1$                              & $79.8 {\pm} 0.1$              & $79.6 {\pm} 0.1$           & $\mathbf{81.0} {\pm} 0.1$                                              \\
            3  & $93.1 {\pm} 0.0$                & $93.2 {\pm} 0.0$                                    & $\mathbf{93.5} {\pm} 0.1$                              & $93.5 {\pm} 0.0$              & $92.3 {\pm} 0.1$           & $93.5 {\pm} 0.1$                                              \\
            5  & $95.7 {\pm} 0.0$                & $95.0 {\pm} 0.1$                                    & $95.8 {\pm} 0.0$                              & $\mathbf{96.0} {\pm} 0.0$              & $95.2 {\pm} 0.2$           & $95.8 {\pm} 0.1$                                              \\
            10 & $97.5 {\pm} 0.0$                & $95.5 {\pm} 0.0$                                    & $97.5 {\pm} 0.0$                              & $\mathbf{97.7} {\pm} 0.0$              & $97.2 {\pm} 0.1$           & $97.5 {\pm} 0.0$                                              \\ \bottomrule
        \end{tabular}
    \end{normalsize}

    \vskip -0.1in
    \label{table:plantnet-300k_appendix}
\end{table}
First, our balanced noise loss gives better top-$K$ accuracies than the cross entropy or the loss from \citet{berrada} for all $K$.
Then, $\ell_{\text{Noised imbal.}}^{K, \epsilon, B,m_y}$ and $\ell_{\text{Noised bal.}}^{K, \epsilon, B}$ produce the best top-1 accuracy, respectively 81.0 and 80.8.

However, we insist that in such an imbalanced setting the macro-average top-$K$ accuracy reported in \Cref{table:plantnet-300k} is much more informative than regular top-$K$ accuracy.
We see significant differences between the losses in \Cref{table:plantnet-300k} which are hidden in \Cref{table:plantnet-300k_appendix} because of the extreme class imbalance.

\textbf{ImageNet}: We test $\ell_{\text{Noised bal.}}^{K, \epsilon, B}$ on ImageNet.
We follow the same procedure as \citet{berrada}: we train a ResNet-18 with SGD for 120 epochs with a batch size of 120 epochs.
The learning rate is decayed by ten at epoch 30, 60 and 90.
For $\ell_{\text{Smoothed Hinge}}^{K, \tau}$, the smoothing parameter $\tau$ is set to 0.1, the weight decay parameter to 0.000025 and the initial learning rate to 1.0, as in \citet{berrada}.
For $\ell_{\textrm{CE}}$, the weight decay parameter is set to 0.0001 and the initial learning rate to 0.1, following \citet{berrada}.
For $\ell_{\text{Noised bal.}}^{K, \epsilon, B}$, we use the same validation set as in \citet{berrada} to set $\epsilon$ to 0.5, the weight decay parameter to 0.00015 and the initial learning rate to 0.1. $B$ is set to 10.
We optimize all losses for $K=1$ and $K=5$. We perform early stopping based on best top-$K$ accuracy on the validation set and report the results on the test set (the official validation set of ImageNet) in \Cref{table:imagenet-appendix} (3 seeds, 95\% confidence interval).
In the presence of low label noise, with an important number of training examples per class and for a nearly balanced dataset, all three losses give similar results.
This is in contrast with \Cref{table:cifar100-top5-acc} and \Cref{table:plantnet-300k}, where significant differences appear between the different losses in the context of label noise or class imbalance.

\begin{table}
    \caption{ImageNet test top-$K$ accuracy, ResNet-18.}
    \vskip 0.1in
    \centering
    \begin{normalsize}
        \begin{tabular}{lccc}
            \toprule
            K  & $\ell_{\textrm{CE}}$ &          $\ell_{\text{Smoothed Hinge}}^{K, 0.1}$ &               $\ell_{\text{Noised bal.}}^{K, 0.5, 10}$           \\ \midrule
            1  & $72.24 {\pm} 0.15$                & $71.43 {\pm} 0.14$                                     & $\mathbf{72.46} {\pm} 0.15$                                     \\
            5  & $90.60 {\pm} 0.05$                & $\mathbf{90.71} {\pm} 0.06$                                     & $90.52 {\pm} 0.07$                                     \\ \bottomrule
        \end{tabular}
    \end{normalsize}

    \vskip -0.1in
    \label{table:imagenet-appendix}
\end{table}

\section{Hyperparameter tuning}
\label{subsec:hyperparams}

\textbf{Balanced case}: For both experiments on CIFAR-100 and ImageNet, we follow the same learning strategy and use the same hyperparameters for $\ell_{\text{Smoothed Hinge}}^{K, \tau}$ than \citet{berrada}.
For $\ell_{\text{Noised bal.}}^{K, \epsilon, B}$, we refer the reader for the choice of $\epsilon$ and $B$ respectively to \Cref{subsec:noise-parameter} and \Cref{subsec:B}: $\epsilon$ should be set to a sufficiently large value so that learning occurs and $B$ should be set to a small value for computational efficiency.

\textbf{Imbalanced case}: For our experiments on imbalanced datasets, we use the grid $\{0.5, 1.0, 2.0, 5.0\}$ for the parameter $\gamma$ of the focal loss and $\{0.1, 1.0\}$ for the parameter $\tau$ of $\ell_{\text{Smoothed Hinge}}^{K, \tau}$.
For $\ell_{\textrm{LDAM}}^{\max m_y}$ and $\ell_{\text{Noised imbal.}}^{K, \epsilon, B, \max m_y}$, the hyperparameter $\max m_y$ is searched in the grid $\{0.2, 0.3, 0.4, 0.5\}$ and we find in our experiments that the best working values of $\max m_y$ happen to be the same for both losses.
For the scaling constant for the scores, we find that 30 and 50 are good default values for respectively $\ell_{\textrm{LDAM}}^{\max m_y}$ and $\ell_{\text{Noised imbal.}}^{K, \epsilon, B, \max m_y}$.
Finally, for $\ell_{\text{Noised imbal.}}^{K, \epsilon, B, \max m_y}$, $\epsilon$ is searched in the grid $\{0.01, 0.05, 0.1\}$.

\end{document}